\pdfoutput=1
\documentclass[10pt,journal]{IEEEtran}
\ifCLASSOPTIONcompsoc
  \usepackage[nocompress]{cite}
\else
  \usepackage{cite}
\fi
\usepackage{mdframed}
\usepackage{amsmath,amssymb,amsfonts }
\usepackage{bbm}
\usepackage{graphicx}
\usepackage{textcomp}
\usepackage{xcolor}
\usepackage{kotex}

\usepackage{setspace}
\usepackage{slashbox}
\usepackage{amsthm,thmtools,thm-restate}
\usepackage{subfigure}
\usepackage{booktabs}
\usepackage{multicol}
\usepackage{multirow}
\usepackage{verbatim} 
\usepackage[normalem]{ulem}
\usepackage{lettrine} 

\usepackage{cases}
\usepackage{mathrsfs}
\usepackage{rotating}
\usepackage{makecell}
\usepackage{float}
\usepackage{hyperref}
\usepackage{cleveref}

\usepackage{colortbl}
\usepackage{arydshln}
\usepackage{array}

\usepackage{amsmath,amsfonts}
\usepackage{braket}
\usepackage{physics}
\usepackage{adjustbox}  
\usepackage{algorithmicx,algpseudocode}
\usepackage[linesnumbered,ruled]{algorithm2e}

\usepackage{amsmath,amsfonts,amssymb}
\usepackage{tabularx}
\usepackage{tikz}
\SetKw{End}{end}
\SetKwBlock{IfBlock}{if}{end}
\SetKwBlock{ElseBlock}{else}{end}

\definecolor{dullgreen}{rgb}{0.3,0.7,0.3}
\definecolor{lightdullgreen}{rgb}{0.5, 0.7, 0.1}
\definecolor{red}{rgb}{0.9, 0.0, 0.0}
\definecolor{lightorange}{rgb}{1.0, 0.45, 0.0}
\definecolor{pastelblue}{rgb}{0.75, 0.90, 0.95}

\newcommand{\fcol}{\cellcolor{dullgreen!40}}
\newcommand{\scol}{\cellcolor{lightdullgreen!30}}
\newcommand{\tcol}{\cellcolor{yellow!30}}

\newcommand{\ftext}[1]{\colorbox{dullgreen!40}{#1}}
\newcommand{\stext}[1]{\colorbox{lightdullgreen!30}{#1}}
\newcommand{\ttext}[1]{\colorbox{yellow!30}{#1}}

\newcommand{\worsecol}{\cellcolor{red!30}}
\newcommand{\badcol}{\cellcolor{lightorange!30}}

\newcommand{\wtext}[1]{\colorbox{red!30}{#1}}
\newcommand{\wstext}[1]{\colorbox{lightorange!30}{#1}}
\usepackage{bbold}

\usepackage{color}

\usepackage{tabulary}
\usepackage{amssymb} %
\usepackage{booktabs} %

\usepackage{bm}

\usepackage{url}
\usepackage{capt-of}
\hypersetup{
    colorlinks=true,
    linkcolor=blue,      %
    citecolor=blue,      %
    urlcolor=blue        %
}

\makeatletter
\def\@IEEEpubidpullup{4.2\baselineskip}
\makeatother
\IEEEpubid{\parbox[b]{\textwidth}{\footnotesize\copyright~2026 IEEE. Personal use of this material is permitted. Permission from IEEE must be obtained for all other uses, in any current or future media, including reprinting/republishing this material for advertising or promotional purposes, creating new collective works, for resale or redistribution to servers or lists, or reuse of any copyrighted component of this work in other works. Accepted for publication in IEEE Transactions on Intelligent Transportation Systems.}}

\begin{document}

\title{ALIVE-LIO: Degeneracy-Aware Learning of Inertial Velocity for Enhancing ESKF-Based  LiDAR-Inertial Odometry}

\author{
    Seongjun Kim\textsuperscript{1}, Daehan Lee\textsuperscript{2}, Junwoo Hong\textsuperscript{2}, Sanghyun Park\textsuperscript{2}, Hyunyoung Jo\textsuperscript{2}, and Soohee Han\textsuperscript{3*}, \textit{Senior Member, IEEE}
        \IEEEcompsocitemizethanks{
        \IEEEcompsocthanksitem 
        {\textsuperscript{1}S. Kim is with the Department of Electrical Engineering, POSTECH, Pohang 37673, Republic of Korea (e-mails: seongjun4527@postech.ac.kr).}
        \IEEEcompsocthanksitem 
        {\textsuperscript{2}D. Lee, J. Hong, S. Park, and H. Jo are with the Department of Convergence IT Engineering, POSTECH, Pohang 37673, Republic of Korea (e-mails: daehanlee@postech.ac.kr, junwooh@postech.ac.kr, pash0302@postech.ac.kr and hyunyoungjo@postech.ac.kr, respectively).}
        \IEEEcompsocthanksitem 
        {\textsuperscript{3}S. Han is with the Department of Electrical Engineering and Convergence IT Engineering, POSTECH, Pohang 37673, Republic of Korea (e-mails: soohee.han@postech.ac.kr).}

}%
}

\IEEEtitleabstractindextext{%
\begin{abstract}
Odometry estimation using light detection and ranging (LiDAR) and an inertial measurement unit (IMU), known as LiDAR–inertial odometry (LIO), often suffers from performance degradation in degenerate environments, such as long corridors or single-wall scenarios with narrow field-of-view LiDAR.
To address this limitation, we propose ALIVE-LIO, a degeneracy-aware LiDAR–inertial odometry framework that explicitly enhances state estimation in degenerate directions.
The key contribution of ALIVE-LIO is the strategic integration of a deep neural network into a classical error-state Kalman filter (ESKF) to compensate for the loss of LiDAR observability.
Specifically, ALIVE-LIO employs a neural network to predict the body-frame velocity and selectively fuses this prediction into the ESKF only when degeneracy is detected, providing effective state updates along degenerate directions.
This design enables ALIVE-LIO to utilize the probabilistic structure and consistency of the ESKF while benefiting from learning-based motion estimation.
The proposed method was evaluated on publicly available datasets exhibiting degeneracy, as well as on our own collected data.
Experimental results demonstrate that ALIVE-LIO substantially reduces pose drift in degenerate environments, yielding the top-ranked result in 22 out of 32 sequences. %
The implementation of ALIVE-LIO is publicly available\footnote{\url{https://github.com/cocel-postech/ALIVE-LIO}}.
\end{abstract}

}

\maketitle

\IEEEdisplaynontitleabstractindextext

\IEEEpeerreviewmaketitle

\section{Introduction}\label{sec:1_intro}

Reliable localization and mapping are fundamental to autonomous robot navigation.
To achieve these capabilities, systems have integrated multiple sensors, including cameras, light detection and ranging (LiDAR), inertial measurement units (IMUs), and global positioning systems (GPSs).
In particular, LiDAR sensors play a crucial role in accurately capturing the three-dimensional (3D) structure of the environment, thereby enabling reliable localization~\cite{intro_lidar_feat, intro_survey_lo}.
Furthermore, integration of LiDAR with an IMU enables robust pose estimation even under highly dynamic motion.
However, in environments with limited geometric features, such as long corridors with one side open, LiDAR measurements may be insufficient for reliable perception, leading to a condition of LiDAR degeneracy where localization may fail even with the aid of IMUs~\cite{intro_survey_lo}.

\IEEEpubidadjcol
To mitigate the impact of insufficient LiDAR measurements, numerous approaches have been proposed.
Some methods focus on fully leveraging LiDAR data~\cite{genzicp, coin_lio}; however, these approaches may yield limited performance when LiDAR measurements along certain directions are sparse.
Alternatively, LiDAR can be fused with other sensors, such as cameras or wheel encoders.
Many studies have incorporated cameras into LiDAR–visual–inertial odometry (LVIO) systems to alleviate the aforementioned limitation by minimizing
photometric errors~\cite{r3live, fast_livo1, fast_livo2, switch_slam, hanbiao_tits_lvio}.
However, LVIO systems remain challenged in environments that are both geometrically and photometrically sparse, which may result in odometry estimation failures.
As one type of auxiliary sensor, wheel encoders can also provide velocity information for odometry, but they cannot be applied to drones, handheld devices, and humanoid or quadruped robots~\cite{taku_wheel_lio}.
Algorithmically, some methods place substantial weight on pose priors to mitigate degeneracy~\cite{x_icp, tucan_field_robotics, relead, zongbo_tits}.
These methods detect degeneracy and perform constrained optimization along the degenerate directions, leveraging pose priors provided by sensors such as IMUs and joint encoders.
However, reliance on IMU-only systems can be susceptible to bias and noise accumulation, which may hinder the maintenance of accurate pose priors over extended periods.

Recent advances in deep learning have begun to address the degeneracy issue with data-driven approaches.
ININ-LIO~\cite{inin_lio} mitigates degeneracy by leveraging deep learning to predict relative positions from bias- and gravity-compensated IMU measurements.
Specifically, this method directly incorporates the deep-learning-estimated degenerate directions into the state update by fully trusting the model outputs, which can lead to performance degradation when the deep-learning outputs are unreliable.
Liao~et al.~\cite{zongbo_tits} proposed a weighting-based update instead of using them directly.
Under degenerate conditions, this method assigns greater weight to the model-based IMU motion prediction according to a heuristic degeneracy ratio, rather than adopting a dynamically principled approach such as Kalman filtering.
In other words, the deep-learning outputs are reflected in the state update by considering only the degeneracy information (as illustrated in Fig.~\ref{fig:state_update}, top), without explicitly accounting for potential correlations with other state variables, such as bias and gravity, which can result in performance degradation in scenarios with persistent degeneracy.
In this regard, developing LiDAR-inertial odometry (LIO) that more comprehensively integrates deep-learning outputs while accounting for their correlations with other state variables is highly valuable.

To more tightly fuse data-driven and model-based LIO techniques, we propose an error-state Kalman filter (ESKF)-based approach that incorporates inertial motion predicted by deep learning from IMU data.
Throughout this paper, we refer to the proposed scheme as ALIVE-LIO (degeneracy-Aware Learning of Inertial Velocity for enhancing ESKF-based LiDAR-Inertial Odometry). 
By restricting state updates to the degenerate directions, ALIVE-LIO maintains accurate estimates along the non-degenerate directions, similarly to ININ-LIO~\cite{inin_lio}.
Specifically, before incorporating learning-based velocity estimates, we first apply constrained optimization within the conventional LIO step to prevent incorrect state updates and ensure reliable corrections along degenerate directions~\cite{x_icp, lp_icp,relead, tucan_field_robotics}.
When degeneracy is detected, learning-based velocity estimates are appropriately weighted and systematically fused into the ESKF, enabling consistent updates of all state components while explicitly accounting for their cross-correlations.
In this sense, ALIVE-LIO can be regarded as a strategic integration of a deep neural network into a classical ESKF, designed to
compensate for the loss of LiDAR observability. 
This design allows ALIVE-LIO to fully exploit the probabilistic structure and consistency of the ESKF, while benefiting from learning-based motion estimation.

To improve generalization of deep learning, the network outputs are represented in the body frame, following motivations from recent inertial odometry (IO) studies~\cite{airio, tartan_imu}.
Additionally, by excluding biases and gravity estimates from the LIO system and representing orientation in the body frame, we further enhance generalization performance.
For a real-world demonstration, the proposed ALIVE-LIO was integrated with the widely used PV-LIO~\cite{pv_lio} via ONNX~\cite{onnx} in an online manner, illustrating effective mitigation of prolonged degeneracy.
Notably, our approach offers an effective alternative in environments with limited geometric or photometric features, where obtaining reliable external cues is challenging.

In summary, our approach makes the following four key claims:
(i)~Degeneracy is effectively mitigated across various platforms operating in degeneracy-prone environments.
(ii)~The proposed ESKF-based integration of learning-based velocity estimates demonstrates greater robustness than existing heuristic methods that loosely couple data-driven and model-based schemes without explicitly modeling system dynamics.
(iii)~In environments that are simultaneously geometrically and photometrically sparse, where the visual measurements that LVIO systems rely on also degrade, ALIVE-LIO maintains reliable odometry, suggesting its potential as an alternative under such conditions.
(iv)~By excluding the LIO-estimated bias and gravity from the neural network inputs, we improve generalization performance.
The following sections, along with our experimental evaluation, support these claims.

The remainder of this study is organized as follows.
Section~\ref{sec:2_preliminaries} reviews related work, including degeneracy detection in LiDAR-based odometry, degeneracy-mitigation strategies, and deep-learning–based inertial odometry. 
Section~\ref{sec:3_algorithm_design} presents the ALIVE-LIO pipeline and its key components.
Section~\ref{sec:4_evaluation} describes the experimental setup and discusses results that support our key claims.
Finally, Section~\ref{sec:5_conclusion} concludes the paper.

\section{Related Work}\label{sec:2_preliminaries}
\subsection{Degeneracy Detection}\label{subsec:21_degeneracy_detection}
LiDAR odometry typically employs the iterative closest point (ICP) algorithm~\cite{ptpl_icp}.
As noted by Tuna~et al.~\cite{x_icp} and Zhang~et al.~\cite{zhang_icra}, LiDAR degeneracy often occurs because of a lack of sufficient constraints in specific directions during the ICP optimization process.
For this reason, LION~\cite{lion} detects degeneracy by performing singular value decomposition (SVD) on the Hessian matrix of the ICP objective, where the condition number is computed by comparing the maximum and minimum eigenvalues.
Following a similar SVD-based strategy, MM-LINS~\cite{mm_lins} detects degeneracy using a state covariance matrix obtained as an output of the LIO system rather than the ICP Hessian, by computing its SVD and checking whether the maximum eigenvalue exceeds a predefined threshold over a specified duration.
Although these approaches are simple to implement, they have the limitation of considering degeneracy along only a single axis.
In other words, they cannot detect degeneracy occurring simultaneously along multiple directions.
In contrast, X-ICP~\cite{x_icp} detects degeneracy along multiple axes by analyzing degeneracy independently along each axis.
Specifically, this method projects the normalized Jacobians of points onto the eigenspace, filters them according to magnitude, and then evaluates strong and weak localizability to classify degeneracy along each axis as ``NONE,'' ``PARTIAL,'' or ``FULL.''
Numerous researchers have extended this method to perform degeneracy detection~\cite{inin_lio, lp_icp, relead, tucan_field_robotics}.
We also adopted the same degeneracy detection framework in the present study.

\subsection{Degeneracy Mitigation} \label{subsec:22_degeneracy_mitigation}
To mitigate degeneracy, numerous approaches have been proposed, which we categorize into four main types.
First, additional LiDAR features can be measured.
GenZ-ICP~\cite{genzicp} addresses the challenge that plane estimation becomes difficult in directions where points are sparse. 
By combining point-to-point and point-to-plane error metrics, it aims to resolve this issue and demonstrates robust performance, particularly in long-corridor environments.
Additionally, COIN-LIO~\cite{coin_lio} addresses the same by leveraging LiDAR intensity measurements to minimize photometric errors.

Second, additional sensors, such as cameras, can be used.
Most LVIO methods use camera, LiDAR, and IMU to estimate the pose of a single frame~\cite{r3live, fast_livo1, fast_livo2}.
In degenerate cases, Lee~et al.~\cite{switch_slam} suggested estimating the pose using only visual and inertial information while excluding LiDAR measurements.

The third approach relies on pose priors along degenerate directions after detecting degeneracy.
X-ICP~\cite{x_icp} performs constrained optimization that imposes hard constraints along degenerate directions, placing greater reliance on pose priors from IMUs and wheel encoders.
Additionally, Tuna~et al.~\cite{tucan_field_robotics} applied various techniques, including inequality constraints and regularization.
Furthermore, to directly apply constrained optimization to LIO systems, RELEAD~\cite{relead} enforces hard constraints within an iterative error-state Kalman filter (IESKF).
However, these methods are effective only for short-term degeneracy, because the IMU motion prediction becomes increasingly inaccurate during prolonged degeneracy.

\begin{figure}[t!]
    \centering
    \subfigure[Liao \textit{et al.}~\cite{zongbo_tits}]{\includegraphics[width=1.0\linewidth]{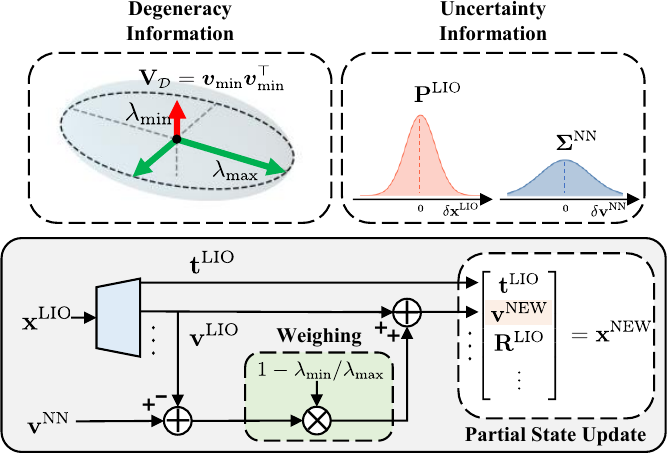}} \\[-0.3mm]
    \subfigure[ININ-LIO~\cite{inin_lio}]{\includegraphics[width=1.0\linewidth]{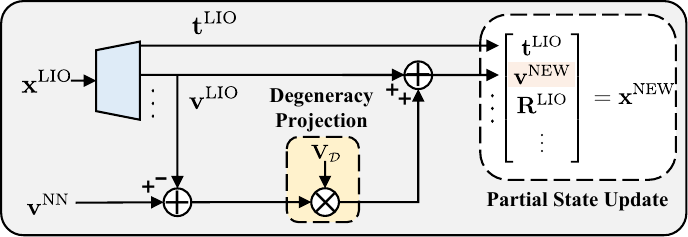}} \\[-0.3mm]
    \subfigure[Ours]{\includegraphics[width=1.0\linewidth]{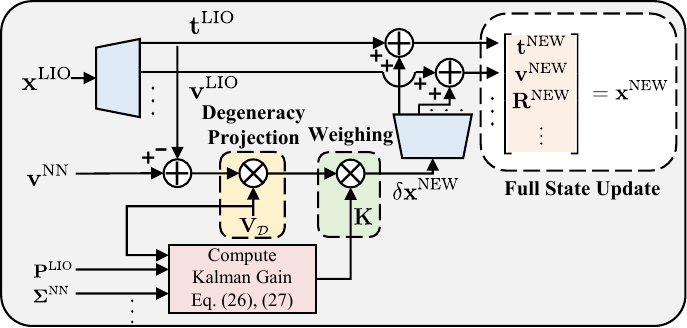}}
    
    \caption{Comparison of state update methods.
    The state $\mathbf{x}$ and its components, such as translation $\mathbf{t}$, velocity $\mathbf{v}$ and rotation $\mathbf{R}$, are denoted as $(\cdot)^\text{LIO}$, $(\cdot)^\text{NN}$, and $(\cdot)^\text{NEW}$ for the LIO state, the neural network state, and the newly updated state, respectively.
    (a) Degeneracy ratio-based weighted update, %
    (b) Projection onto the degenerate space, %
    (c) Our ESKF-based update.
    Note that, in contrast to (a) and (b), our approach performs a full state update by leveraging the system uncertainty $\mathbf{P}^\text{LIO}$ and the data-driven uncertainty $\bm{\Sigma}^\text{NN}$, rather than updating only the velocity.
    }
    \label{fig:state_update}
    \vspace{-6mm}
\end{figure}

As a fourth approach, ININ-LIO~\cite{inin_lio} applied learning-based IO specifically along degenerate directions.
However, this approach directly relies on the deep-learning outputs, which may occasionally impede stable state updates.
In contrast, Liao~et al.~\cite{zongbo_tits} adopted a heuristic weighting strategy rather than directly applying the estimates.
However, these two methods consider IMU motion estimation without explicitly accounting for correlations among state variables, which can result in odometry errors during long-term degeneracy.
In particular, because ININ-LIO~\cite{inin_lio} feeds bias- and gravity-compensated IMU measurements into a deep-learning model, insufficient modeling of the correlations among state variables can lead to inaccurate compensation during prolonged degeneracy, thereby undermining the reliability of the learning-based predictions.
Motivated by these findings, we propose an ESKF-based state update that tightly fuses data-driven and model-based information.
The corresponding Kalman gain is computed from the combined uncertainties of the deep-learning outputs and the LIO system, enabling consistent and robust state estimation.
In addition, degeneracy information is incorporated by projection along the degenerate directions, similar to ININ-LIO~\cite{inin_lio}.
To provide a clear schematic illustration of our approach, we highlight the differences from the aforementioned approaches~\cite{zongbo_tits, inin_lio} in Fig.~\ref{fig:state_update}.

\subsection{Deep Inertial Odometry} \label{subsec:23_learning_io}
Deep-learning-based IO can be categorized into three levels: sensor, algorithm, and application~\cite{io_survey}.
Because the application level focuses on platform-specific designs, it is outside the scope of this study.
At the sensor level, performance is improved through the calibration of measurement errors and noise characteristics.
Ori-Net~\cite{ori_net} generates calibrated gyroscope signals from raw inputs, whereas AirIMU~\cite{airimu} predicts bias and noise from raw gyroscope measurements and linear accelerations, and subsequently corrects the raw signals.
At the algorithm level, inertial positioning is performed.
TLIO~\cite{tlio} learns 3D location displacements and covariances from gravity-aligned inertial data and fuses them with an extended Kalman filter.
Similarly, ININ-LIO~\cite{inin_lio} utilizes the same outputs while leveraging a transformer-based neural network architecture to improve performance.
Furthermore, several studies focus on learning inertial positioning through velocity outputs instead of relative positions~\cite{ridi, dive, airio, jin_tits, tartan_imu}.
Notably, AirIO~\cite{airio} enhances generalization by representing outputs in the body frame and additionally encodes orientation to supply extra information to the neural network.
Furthermore, by excluding biases and gravity, neural networks are prevented from learning unnecessary elements, which improves their generalization performance.
For this reason, some existing methods exclude bias and gravity~\cite{tartan_imu, inin_lio}.
TartanIMU~\cite{tartan_imu} assumes constant biases and gravity.
ININ-LIO~\cite{inin_lio} also omits the biases estimated from the LIO system, but assumes gravity to be constant.
However, the assumption of constant gravity leads to discrepancies with LIO systems that estimate gravity as part of the state~\cite{fast_lio1, fast_lio2, pv_lio}.
Therefore, unlike these methods~\cite{tartan_imu, inin_lio}, we also omit gravity estimated from LIO, reducing such discrepancies and improving the generalization performance of the neural network.

\begin{figure*}[t!]
    \centering
    \includegraphics[width=0.95\linewidth]{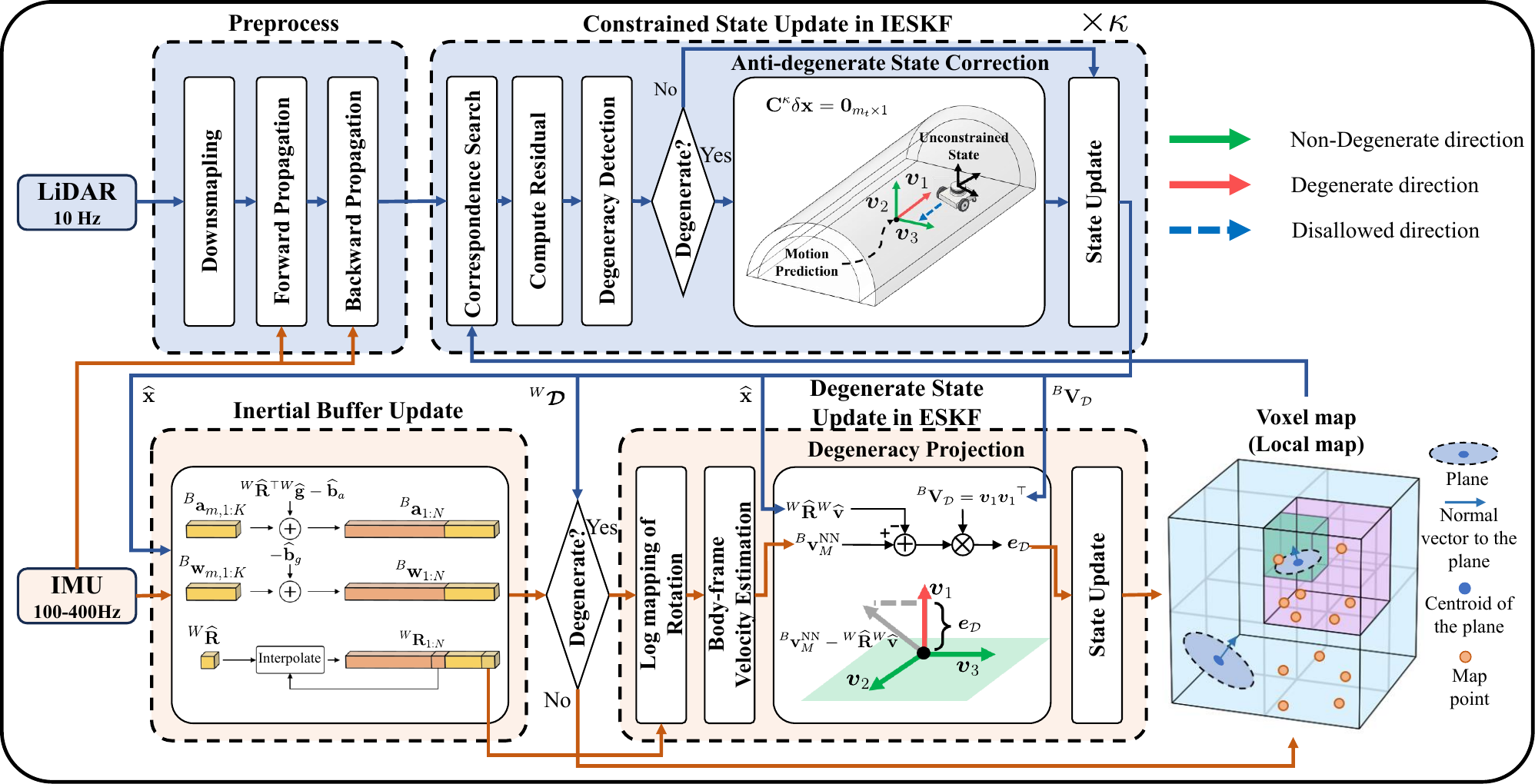}\\

    \caption{Pipeline of ALIVE-LIO.}
    \label{fig:pipeline}
    \vspace{-4mm}
\end{figure*}

\section{Degeneracy-Aware Learning of Inertial Velocity for ESKF-Based LiDAR-Inertial Odometry}\label{sec:3_algorithm_design}
\subsection{System Overview} \label{subsec:31_system_overview}
The pipeline of our method is based on PV-LIO~\cite{pv_lio}, as shown in Fig.~\ref{fig:pipeline}.
Our system consists of two main modules arranged in a sequential pipeline.
The first module detects degeneracy and suppresses its effects within the IESKF, producing an odometry estimate along the non-degenerate directions (upper part of Fig.~\ref{fig:pipeline}).
From this output, the second module then employs deep learning to correct the state along the degenerate directions (lower part of Fig.~\ref{fig:pipeline}).
In the first module, we begin by performing motion prediction by forward propagation using the IMU.
Subsequently, backward propagation is performed for each LiDAR point according to its timestamp, transforming the points into the world frame.
\begin{table}[t!]
\caption{Notations used in this paper.}
\centering
\small
\begin{tabular}{c|c}
\toprule[1pt]
\textbf{Notation} & \textbf{Description} \\
\midrule
$^W(\cdot)\quad $                 & A vector $(\cdot)$ in the world frame \\
$^B(\cdot)\ \ $                   & A vector $(\cdot)$ in the body frame \\
$\ \ (\cdot)_m$                   & Measured raw sensor data \\
$\quad(\cdot)^\text{NN}$          & Output of the neural network \\
$\ \widehat{\mathbf{x}}^\kappa$   & $\kappa$-th update of $\mathbf{x}$ in IESKF \\ 
$\delta \mathbf{x}$               & Error state in IESKF \\
$\overline{\mathbf{x}}$           & State of $\mathbf{x}$ resulting from update \\ 
\bottomrule[1pt]
\end{tabular}
\label{tab:notation}
\vspace{-4mm}
\end{table}
In the IESKF, the state is iteratively updated up to $\kappa$ times or until convergence by minimizing the point-to-plane error through matching each point with the corresponding plane on the local map.
To mitigate degeneracy, we detect it at each iteration, and whenever degeneracy occurs, a constrained optimization is performed in the world frame.
Once the first module has completed its processing, the accelerometer and gyroscope measurements are compensated for gravity and biases, and the estimated orientation ${}^W\widehat{\mathbf{R}}$ is interpolated to the IMU timestamps, during which $K$ IMU measurements are collected per LiDAR scan. These results are stored in the inertial buffer $\mathcal{B}_I$.
When degeneracy occurs, the velocity in the body frame is predicted using the $N$ data points stored in $\mathcal{B}_I$.
Specifically, the stored orientations ${}^W\widehat{\mathbf{R}}_{1:N}$ are expressed in the body frame based on the most recently received value ${}^W \widehat{\mathbf{R}}_N$, and are represented as its Lie algebra $\mathfrak{so}(3)$, before being fed into the neural network trained for body-frame velocity estimation.
The residual between the velocities estimated by the neural network and those estimated by the IESKF is projected onto the degenerate directions in the body frame, which are obtained at the last step of the IESKF and then rotated into the body frame for use.
The ESKF then updates the state using the projected residual, thereby mitigating degeneracy.
Once odometry estimation is complete, the downsampled LiDAR points are transformed into the world frame and stored in the local map.
The state $\mathbf{x}$ is defined as
\begin{equation*}
    \mathcal{M} \triangleq \mathrm{SO}(3) \times \mathbb{R}^{15},\quad \mathrm{dim}(\mathcal{M})=18,
\end{equation*}
\begin{equation*}
    \mathbf{x} = 
    \begin{bmatrix}
        {}^W\mathbf{R}^\top & {}^W\mathbf{t}^\top & {}^W\mathbf{v}^\top & \mathbf{b}_{g}^\top & \mathbf{b}_a^\top & {}^W\mathbf{g}^\top
    \end{bmatrix}^\top \in \mathcal{M},
\end{equation*}
where  ${}^W\mathbf{R}$, ${}^W\mathbf{t}$, ${}^W\mathbf{v}$, and ${}^W\mathbf{g}$ are the rotation, translation, velocity, and gravity in the world frame, and $\mathbf{b}_g$ and $\mathbf{b}_a$ denote the gyroscope and accelerometer biases, respectively.
The coordinates and notations are summarized in Table~\ref{tab:notation}.

\subsection{Deep Learning-Based Velocity Estimation} \label{subsec:32_neural_network}
 
\begin{figure}[t!]
    \centering
    \vspace{-2.7mm}
    \subfigure[]{\includegraphics[width=0.492\linewidth]{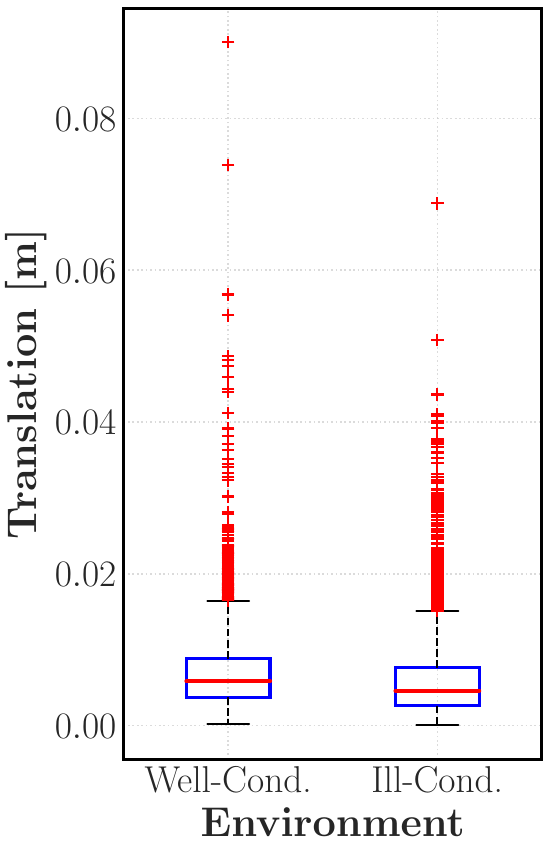}}
    \subfigure[]{\includegraphics[width=0.492\linewidth]{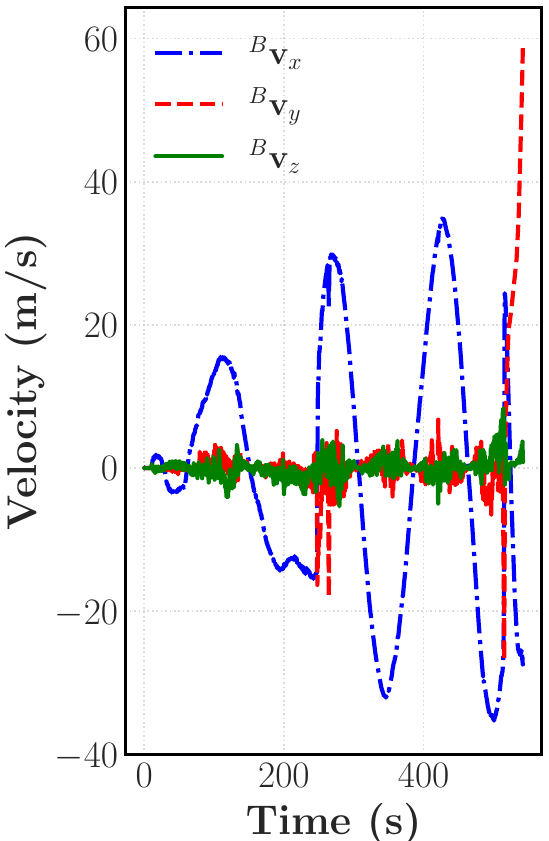}}
    
    \caption{Comparison of motion updates between the well-conditioned \texttt{TunnelingTunnel\_1} and the ill-conditioned (degenerate) \texttt{ShieldTunnel\_1} sequences of the GEODE dataset~\cite{geode}. (a) Box plot of the LiDAR-only translational change in well- and ill-conditioned (degenerate) scenarios. (b) Body-frame velocity in the degenerate scenario.}
    \label{fig:degeneracy_illustrate_figure}
    \vspace{-6mm}
\end{figure}

\subsubsection{Rationale for Velocity-Based Inertial Estimation}
Intuitively, LiDAR-based odometry may diverge in degenerate scenarios because the corresponding least-squares problem of ICP becomes approximately singular, resulting in large relative translation updates along poorly constrained directions.
To investigate this phenomenon, we conducted experiments using PV-LIO~\cite{pv_lio} in the long corridor scenarios of the GEODE dataset~\cite{geode}, distinguishing between the well-conditioned \texttt{TunnelingTunnel\_1} and the ill-conditioned (degenerate) \texttt{ShieldTunnel\_1} cases.
Specifically, the LiDAR-only translational changes are calculated as the difference between the LIO-updated pose ${}^W\mathbf{T}^\text{LIO} \in \mathrm{SE}(3)$ and the pose ${}^W\mathbf{T}^\text{IMU}\in \mathrm{SE}(3)$ predicted from IMU in translation motion, as expressed in $\|(({}^W\mathbf{T}^\text{IMU})^{-1}\cdot{}^W\mathbf{T}^\text{LIO})_\text{trans}\|_2$.
As shown in Fig.~\ref{fig:degeneracy_illustrate_figure}(a), the actual LiDAR-only translation changes show no substantial difference between the two sequences.
This result suggests that, even under degeneracy, numerical instability is not the main source of divergence.
The reason for this effect is that Kalman filter–based LIO algorithms incorporate system uncertainty, which serves a role similar to that of a regularization term~\cite{marios_kalman_eccomas}.
Furthermore, the small magnitude of the LiDAR-only translational change suggests that IMU motion predictions drive most pose updates in LIO.
This result can be explained by the following formulas~\cite{christian_tro}:
\vspace{-2mm}
\begin{align}
\mathbf{b}_{a,{i+1}} & = \mathbf{b}_{a,i},  \label{eq1} \\
    {}^W\mathbf{v}_{i+1} &= {}^W\mathbf{v}_i  + {}^W\mathbf{g}\Delta t + {}^W\mathbf{R}_i({}^B\mathbf{a}_{m,i} - \mathbf{b}_{a,i}), \label{eq2}\\
    {}^W\mathbf{t}_{i+1} &= {}^W \mathbf{t}_i + \underbrace{{}^W\mathbf{v}_i\Delta t + \frac{1}{2} {}^W\mathbf{g}\Delta t^2}_{\Delta {}^{W}\mathbf{t}^{i+1}_i}. \label{eq3}
\end{align}
As shown in (\ref{eq1})--(\ref{eq3}), the temporal translation $\Delta{}^W\mathbf{t}^{i+1}_{i}$ is substantially influenced by a velocity, highlighting the critical importance of accurate velocity estimation.
As shown in Fig.~\ref{fig:degeneracy_illustrate_figure}(b), the body-frame velocity ${}^B \mathbf{v}$ is observed to be erroneous in the degenerate scenario.
Although this scenario involves forward motion only, ${}^B \mathbf{v}_x$ occasionally becomes negative, indicating a backward motion.
Furthermore, the velocity magnitude appears substantially larger than expected, suggesting that error accumulation prevents the IMU from accurately estimating the state alone.
Although accurately estimating the bias of an IMU could mitigate this issue, obtaining calibrated bias data for constructing training datasets is challenging~\cite{io_survey}.
Therefore, we set velocities as the target output, with the expectation that updating them can help stabilize the LIO system.

\begin{figure}[t!]
  \centering
  \vspace{-0.8mm}
  \includegraphics[width=0.95\linewidth]{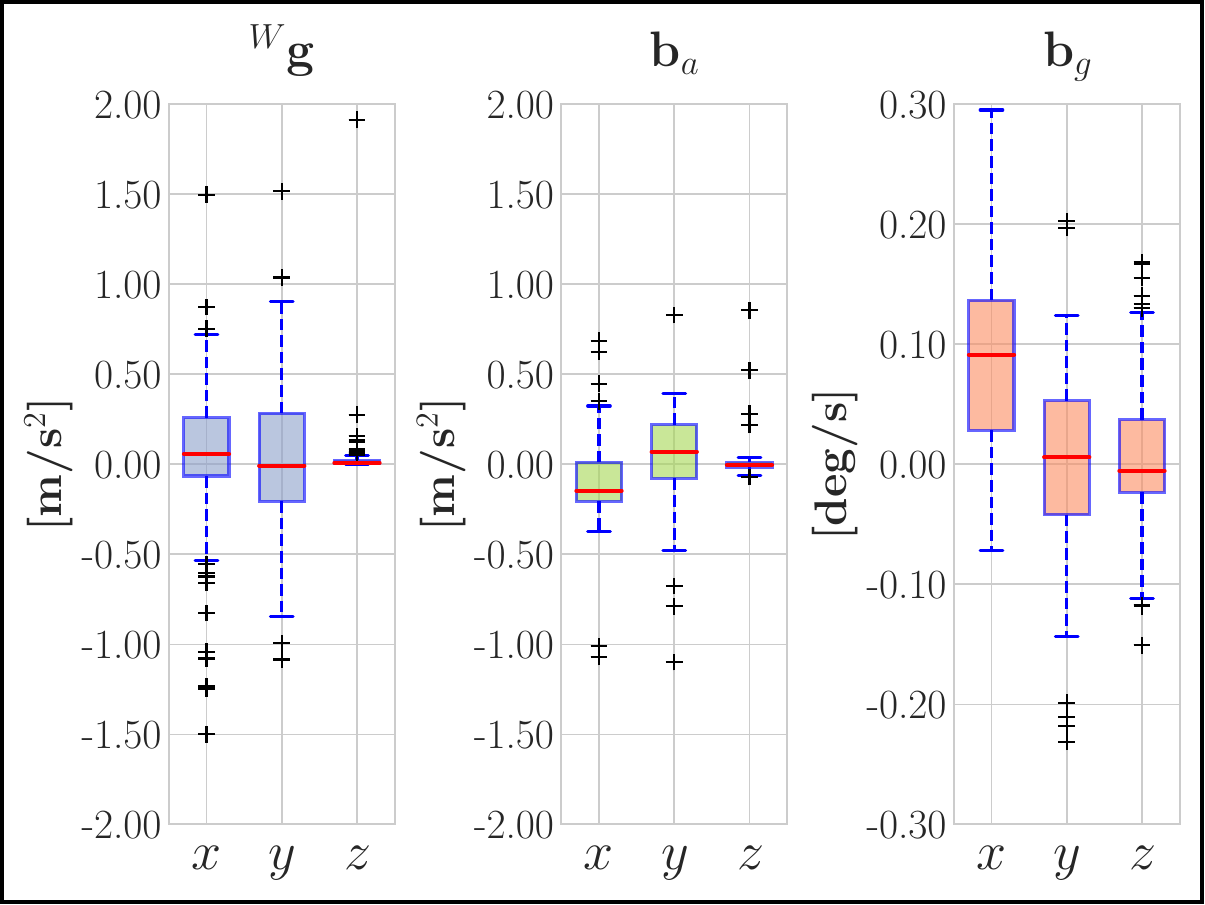}
  \caption{Distribution of gravity ${}^W\mathbf{g}$, acceleration bias $\mathbf{b}_a$ and gyro bias $\mathbf{b}_g$ estimated by the LIO system.
  The $z$-component of gravity is shown after removing the well-known constant magnitude.}
  \label{fig:imu_bias_gravity_distribution}
  \vspace{-4mm}
\end{figure}
\subsubsection{Input Design Using LIO Outputs}
Most learning-based IO methods use ${}^W\mathbf{R}$ as ground truth (GT)~\cite{tlio, airio, imo}.
In addition, to remove ${}^W\mathbf{g}$ in the body frame, it is expressed as ${}^W\mathbf{R}^\top {}^W\mathbf{g}$, where ${}^W\mathbf{g}$ is set to the well-known constant 9.81\,m/s$^2$~\cite{inin_lio, tartan_imu}.
However, because many LIO systems do not know the initial roll and pitch, the rotation of the starting frame is set to the identity~\cite{fast_lio1, fast_lio2, pv_lio}.
${}^W\mathbf{g}$ is initialized by remaining stationary for a few seconds and taking the mean value, which is then used as part of the state for subsequent updates.
Moreover, IMUs can introduce biases, and if these are incorporated, the neural network would need to learn the bias across various scenarios, which inevitably degrades generalization performance.
As shown in Fig.~\ref{fig:imu_bias_gravity_distribution}, for the 50 training and evaluation sequences listed in Table~\ref{tab:dataset} of the Supplementary Material (hereafter, the prefix ``S'' denotes the Supplementary Material), the biases and gravity estimated by the LIO system at stationary states (start and end frames) are distributed quite variably.
Therefore, to improve generalization, we exclude the gravity and biases estimated by the LIO system during neural network training.
Additionally, AirIO~\cite{airio} represents the orientation in the world frame, which introduces a mismatch with other inputs expressed in the body frame.
Thus, we represent the body-frame orientations relative to the last frame of $\mathcal{B}_I$.
Our neural network takes inputs that are preprocessed from the measurements stored in $\mathcal{B}_I$ as follows:
\begin{align}
	{}^B\mathbf{a}_i &= {}^B{\mathbf{a}}_{m, i} - \widehat{\mathbf{b}}_{a,i} + {}^W\widehat{\mathbf{R}}_i^\top {}^W\widehat{\mathbf{g}}_i, \label{eq4}\\
	{}^B\mathbf{w}_i &= {}^B{\mathbf{w}}_{m, i} - \widehat{\mathbf{b}}_{g,i}, \label{eq5}\\
	{}^B\boldsymbol{\xi}_i &= \mathrm{Log}({}^W\widehat{\mathbf{R}}_N^\top {}^W\widehat{\mathbf{R}}_i),    \label{eq6}
\end{align}
where $i$ indexes each sequential data point ($i = 1, 2, \dots, N$), and $\mathrm{Log}: \mathrm{SO(3)} \rightarrow \mathfrak{so}(3)$ maps a rotation matrix to its corresponding Lie algebra representation.
Additionally, because the LIO system updates at 10\,Hz and the IMU operates at a higher frequency, ${}^W\widehat{\mathbf{R}}$ is interpolated to align with the IMU measurement.
Gravity and the biases are excluded from interpolation because they remain stable over short intervals.

\subsubsection{Learning Framework for Velocity-Based Inertial Estimation}
We adopted the neural network architecture from AirIO~\cite{airio}, which leverages bi-GRUs to learn temporal dependencies in the time-series data.
The network is designed to output the body-frame velocity and its covariance, using the following loss function:
\begin{equation}    \label{eq7}
    ({}^B\mathbf{v}^\text{NN}_{1:M}, \boldsymbol{\sigma}^\text{NN}_{1:M}) = f_\theta({}^B\mathbf{a}_{1:N}, {}^B\mathbf{w}_{1:N}, {}^B\boldsymbol{\xi}_{1:N}),
\end{equation}
\begin{align}
    \mathcal{L} &= \frac{1}{M}\sum_{j=1}^{M} (\mathcal{L}_{\text{Huber},j} + \lambda_C\mathcal{L}_{C,j}),   \label{eq8}\\
    \mathcal{L}_{\text{Huber},j} &= \begin{cases}
        \frac{1}{2}\|\bm{e}_{v, j}\|_2^2 & \text{if }\|\bm{e}_{v,j}\|_2 < \delta \\
        \delta (\|\bm{e}_{v,j}\|_2-\frac{1}{2}\delta) & \text{otherwise}
    \end{cases}, \label{eq9}\\
    \mathcal{L}_{C, j} &=  {\bm{e}}_{v,j}^{\top} {(\boldsymbol{\Sigma}^{\text{NN}}_{v,j}})^{-1} {\bm{e}}_{v,j} + \ln{(\det\boldsymbol{\Sigma}^{\text{NN}}_{v, j})}, \label{eq10}
\end{align}
where $M$ is the number of sequential outputs, $\lambda_C$ is a scaling factor, the $j$-th velocity error is defined as ${\bm{e}}_{v,j}={}^B\mathbf{v}^\text{GT}_{j}-{}^B\mathbf{v}^\text{NN}_{j}$, and the corresponding covariance is given by $\boldsymbol{\Sigma}^{\text{NN}}_{v,j} = \operatorname{diag}(\boldsymbol{(\sigma}^\text{NN}_j)^2)$.
The $M$ outputs are predicted at equal intervals over the total time spanned by the $N$ frames, such that the timestamp of the last output frame coincides with that of the last input frame.
Furthermore, the intervals between GT frames are relatively long across datasets; for instance, GEODE and UrbanNav provide GT at 1\,s intervals~\cite{geode, urban_nav_hk, urban_nav_hsu}.
Even with interpolation between these intervals to align the GT with the IMU timestamps, accurately capturing the motion is challenging.
Therefore, we trained the network using the LIO's outputs as GT labels (pseudo-GT) in well-conditioned environments, which provide measurements at 0.1\,s intervals (see Section~\ref{appendix:pvlio_reliability} for the reliability of these pseudo-GT labels).
By representing all inputs and outputs in the body frame, the influence of accumulated drift errors was reduced.

\subsection{Degeneracy Analysis} \label{subsec:33_body_degeneracy}
As in previous studies~\cite{x_icp, relead}, we incorporate the world-frame degeneracy into the IESKF.
However, during the state update of the ESKF, we consider only the degenerate directions expressed in the body frame.
Moreover, as discussed in Section~\ref{subsec:32_neural_network}, inaccuracies in velocity estimation are closely related to degeneracy in the translational component; therefore, the detection is performed exclusively on the translational part.
\subsubsection{Degeneracy Detection in the World Frame}
Here, we follow the X-ICP approach~\cite{x_icp}.
The Jacobian ${}^W\mathbf{J}$ for the translation part in ICP represents the force along the corresponding direction and can be formulated as
\begin{align}
    {}^W\mathbf{J} = \begin{bmatrix}
        {}^W\mathbf{n}_1 & ... & {}^W\mathbf{n}_p & .... & {}^W\mathbf{n}_{N_\text{pts}}
    \end{bmatrix}^\top \in \mathbb{R}^{N_\text{pts} \times 3},  \label{eq11}
\end{align}
where ${}^W \mathbf{n}_{p}$ denotes the normal vector, and $N_\text{pts}$ is the number of points in the scan matched with planes in the map, with $p = 1,2,\dots,N_\text{pts}$.
The corresponding Hessian is given by ${}^W\mathbf{H} = {}^W\mathbf{J}^\top {}^W\mathbf{J}$.
Because ${}^W\mathbf{H}$ is a real symmetric matrix, performing SVD on it yields the following decomposition:
\begin{align}
{}^W\mathbf{H} &= {}^W\mathbf{V} ({}^W\boldsymbol{\Lambda}) {}^W\mathbf{V}^\top, \label{eq12}\\
{}^W\boldsymbol{\Lambda} &= {}^W\mathbf{V}^\top ({}^W\mathbf{H}) {}^W\mathbf{V} \nonumber \\
&= {}^W\mathbf{V}^\top ({}^W\mathbf{J}^\top {}^W\mathbf{J}) {}^W\mathbf{V}  \\
&= ({}^W\mathbf{J} {}^W\mathbf{V})^\top ({}^W\mathbf{J} {}^W\mathbf{V})\nonumber,   \label{eq13}
\end{align}
where ${}^W\boldsymbol{\Lambda} = \operatorname{diag}(\begin{bmatrix}
    {}^W\lambda_1 & {}^W\lambda_2 & {}^W\lambda_3
\end{bmatrix})$
and ${}^W\mathbf{V} = \begin{bmatrix}
    {}^W\bm{v}_1 & {}^W\bm{v}_2 & {}^W\bm{v}_3
\end{bmatrix}$ represent the eigenvalues and eigenvectors, respectively.
The eigenvalues can be expressed as ${}^W \lambda_q = \sum_{p=1}^{N_\text{pts}} \left({}^W \mathbf{n}_{p}^\top \, {}^W \bm{v}_q \right)^2$ for $q \in \{1,2,3\}$, indicating that each point contributes to the $q$-th eigenvalue by projecting into the eigenspace and summing the squared projections~\cite{lp_icp}.
In X-ICP~\cite{x_icp}, instead of using this method directly, the contribution of ${}^W\mathbf{n}_{p}$ to the translation component is emphasized to represent strong and weak localizability.
Instead of squaring ${}^W\mathbf{J} {}^W\mathbf{V}$, ${}^W\mathbf{n}_{p}$ is projected into the eigenspace, and the magnitude of each element is used to represent its contribution to localizability.
To begin, we define
\begin{equation}    \label{eq14}
    {}^W\bm{L} = ({}^W \mathbf{J} {}^W \mathbf{V})^{|\cdot|} \in \mathbb{R}^{N_\text{pts} \times 3}, \\
\end{equation}
where $(\cdot)^{|\cdot|}$ denotes the element-wise absolute operation.
${}^W \bm{L}(p, q)$ means the ($p$, $q$)-th entry of ${}^W \bm{L}$ in (\ref{eq14}).
Subsequently, two localizability vectors parametrized with $\theta_c$ are obtained through a filtering and pointwise summation process as follows:
\begin{equation}    \label{eq15}
    {}^W\bm{L'}_c(q) = \sum_{p=1}^{N_\text{pts}} \mathbb{1}({}^W\bm{L}(p,q) > \cos(\theta_c)) \cdot {}^W\bm{L}(p,q),
\end{equation}
where $\mathbb{1}(\cdot)$ is the indicator function, which equals $1$ if the argument is true and $0$ otherwise, and $c \in \{w, s\}$, with $w$ and $s$ denoting ``weak'' and ``strong'', respectively.
The user-defined thresholds $\cos{(\theta_w)}$ and $\cos{(\theta_s)}$ are applied to obtain strong and weak localizability vectors by performing degeneracy categorization.
Although the original method~\cite{x_icp} classifies degeneracy as ``FULL,'' ``PARTIAL,'' or ``NONE,'' we adopt a binary classification between well-conditioned and degenerate cases, similar to~\cite{zhang_icra}.
The classification is defined as
\begin{align} \label{eq16}
{}^W\mathcal{D}(q) =
\begin{cases}
\text{“Degenerate”} &
\text{if } {}^W\bm{L}'_w(q) < \tau_w 
\mathrel{\wedge}
{}^W\bm{L}'_s(q) < \tau_s, \\
\text{“Well-Cond.”} & \text{otherwise},
\end{cases} \nonumber
\end{align}
where $\tau_w$ and $\tau_s$ are user-defined thresholds, and $\mathrel{\wedge}$ denotes the logical AND operator.
Under this approach, the temporal frame is considered degenerate if any $q$ exists such that ${}^W\boldsymbol{\mathcal{D}}(q) == \text{``Degenerate"}$.
Moreover, in this case, ${}^W\bm{v}_q$ represents the degenerate direction, which the IESKF subsequently uses to perform constrained optimization along the degenerate direction (see Section~\ref{subsec:34_degeneracy_optim} and (\ref{eq22})--(\ref{eq25})).
To explicitly denote the collection of all degenerate components, we define the degenerate index set as
\begin{equation}    \label{eq17}
    \mathbf{Q}_\mathcal{D} = \bigl\{ q \,\big|\, {}^W\boldsymbol{\mathcal{D}}(q) == \text{``Degenerate"}\}.
\end{equation}

\subsubsection{Degeneracy Formulation in the Body Frame}
By rotating ${}^W \mathbf{n}_p$ in the body frame and constructing the corresponding Jacobian ${}^B \mathbf{J}$, degeneracy can be detected in the body frame.
${}^B \mathbf{J}$ is given as
\begin{align}   \label{eq18}
    {}^B\mathbf{J} &= \begin{bmatrix}
        {}^B\mathbf{n}_1 & ... & {}^B\mathbf{n}_{N_\text{pts}}
    \end{bmatrix}^\top \nonumber\\
    & = \begin{bmatrix}
        {}^W\mathbf{R}^\top {}^W\mathbf{n}_1 & ... & {}^W\mathbf{R}^\top {}^W\mathbf{n}_{N_\text{pts}}    \end{bmatrix}^\top\\
    & = {}^W\mathbf{J} {}^W\mathbf{R}  \nonumber,
\end{align}
where $({}^W\mathbf{R}^\top{}^W\mathbf{n}_p)^\top = {}^W\mathbf{n}_p^\top {}^W\mathbf{R}$ and the Hessian ${}^B\mathbf{H}$ in the body frame can be expressed as
\begin{align}   \label{eq19}
    {}^B \mathbf{H} &= {}^B \mathbf{J}^\top {}^B \mathbf{J} \nonumber\\
    &=  ({}^W \mathbf{R}^\top {}^W \mathbf{J}^\top)( {}^W \mathbf{J} {}^W \mathbf{R}) \\
    &= {}^W \mathbf{R}^\top {}^W\mathbf{H} {}^W \mathbf{R}. \nonumber
\end{align}
The eigenvectors in the body frame can be obtained through a similar transformation relating ${}^W \mathbf{H}$ and ${}^B \mathbf{H}$ via ${}^W\mathbf{R}$:
\begin{align}   \label{eq20}
    {}^B\bm{v}_q &= {}^W \mathbf{R}^\top {}^W\bm{v}_q,\quad {}^B\lambda_q={}^W\lambda_q.
\end{align}
This means that the degenerate directions in the world and body frames are related by a simple rotation. 
Furthermore, because ${}^B\lambda_q = {}^W\lambda_q$, the localizability contributions remain unchanged. 
This relation can be expressed mathematically as
\begin{equation}
\begin{aligned} \label{eq21}
    {}^B\mathbf{J} {}^B \mathbf{V} &= ({}^W\mathbf{J} {}^W\mathbf{R}) ({}^W\mathbf{R}^\top{}^W \mathbf{V})\\
    & = {}^W\mathbf{J} \underbrace{{}^W\mathbf{R} {}^W\mathbf{R}^\top{}}_{\mathbf{I}_3} {}^W \mathbf{V} = {}^W\mathbf{J} {}^W \mathbf{V},
\end{aligned}
\end{equation}
demonstrating that the magnitude of the influence is the same in both frames.
Accordingly, the degenerate index set $\mathbf{Q}_\mathcal{D}$ is frame-invariant, i.e., it takes the same form in both the world and body frames.
In this respect, ${}^B \mathbf{V}$ and $\mathbf{Q}_\mathcal{D}$ are employed to project the residual between the velocities estimated by the deep learning and the LIO system onto the degenerate directions expressed in the body frame (see Section~\ref{subsec:34_degeneracy_optim} and (\ref{eq26})--(\ref{eq30})).

\subsection{Degeneracy-Aware Full State Estimation} \label{subsec:34_degeneracy_optim}
As mentioned in Section~\ref{subsec:31_system_overview}, our system mitigates the effects of degeneracy through a structured sequence of operations.
First, at each iteration of the IESKF, degeneracy is detected as described in Section~\ref{subsec:33_body_degeneracy}.
Then, when degeneracy occurs, constrained optimization is performed within the IESKF along the degenerate directions to maintain stability.
Finally, after the IESKF terminates, a state update using deep-learning-based velocity estimates is applied, ensuring robustness under degenerate conditions.
\subsubsection{Anti-Degeneration in LiDAR Inertial Odometry}
We detect degeneracy at each iteration to perform constrained optimization within the IESKF.
The reason for performing constrained optimization is that LiDAR provides insufficient information along degenerate directions~\cite{x_icp}.
Without applying such constrained optimization, repeatedly updating the state along the degenerate directions would result in unacceptably large cumulative errors.

Updating only along non-degenerate directions corresponds to enforcing the following equality constraints:
\begin{equation}    \label{eq22}
    \underbrace{
    \begin{bmatrix}
        {}^W\bm{v}_{q}^{\kappa \top} \\
        \vdots
    \end{bmatrix}}_{\mathbf{C}^\kappa \in \mathbb{R}^{m_\mathrm{t} \times 3}, \; q \in \mathbf{Q}_\mathcal{D}}
    \delta{{}^W\mathbf{t}}
    = \mathbf{0}_{m_\mathrm{t}\times 1},    
\end{equation}
where $\delta{{}^W\mathbf{t}}$ denotes the translational part of $\delta{\mathbf{x}}$, and $m_\mathrm{t}$ denotes the number of degenerate directions in the translational space.
Then, by applying Lagrange multipliers, constrained optimization can be formulated efficiently.
The resulting state update is performed as follows:
\begin{equation}    \label{eq24}
    \mathbf{G}^\kappa = \mathbf{\widehat{P}}_\mathrm{tt} \mathbf{C^{\kappa \top}}(\mathbf{C}^\kappa\mathbf{\widehat{P}}_\mathrm{tt} \mathbf{C^{\kappa \top}})^{-1} \in \mathbb{R}^{3 \times m_\mathrm{t}},
\end{equation}
\begin{equation}    \label{eq25}
    \begin{aligned}
        \delta{\mathbf{t}}^{\kappa+1}_{\kappa} &\leftarrow{} \delta{\mathbf{t}}^{\kappa+1}_{\kappa} \boxminus (\mathbf{G}^\kappa \mathbf{C}^\kappa \mathbf{\delta{t}}^{\kappa+1}_{\kappa}), \\
        \widehat{\mathbf{x}}^{\kappa+1} & = {} \widehat{\mathbf{x}}^{\kappa} \boxplus \delta{\mathbf{x}}^{\kappa+1}_{\kappa} , 
    \end{aligned}
\end{equation}
where $\widehat{\mathbf{P}}_\mathrm{tt}$ is the translational block of the state covariance matrix $\widehat{\mathbf{P}}$, and $\boxplus$/$\boxminus$ are the state update operators defined in~\cite{fast_lio1}.
The error state at the \(\kappa\)-th iteration is defined as $\delta \mathbf{x}^{\kappa+1}_{\kappa} = \widehat{\mathbf{x}}^{\kappa+1} \boxminus \widehat{\mathbf{x}}^{\kappa}$, with $\delta \mathbf{t}^{\kappa+1}_{\kappa}$ denoting its translational part, and $\leftarrow$ indicating an in-place update.
Although the constrained optimization suppresses erroneous translational updates caused by insufficient LiDAR information, it does not fully resolve velocity estimation errors along the degenerate directions.
For this reason, we employ a deep-learning-based velocity estimation method to update states along the degenerate directions, thereby avoiding reliance on corrupted LiDAR information. 

\subsubsection{Degeneracy-Aware State Correction with Learning-Based Velocity Estimation}
After all iterations of the IESKF are completed, the state $\widehat{\mathbf{x}}^{\kappa+1}$ is obtained, and ${}^W\widehat{\mathbf{R}}^{\kappa+1}$ is added to $\mathcal{B}_I$, followed by interpolation according to the IMU timestamps.
For simplicity of notation, we represent $\widehat{\mathbf{x}}^{\kappa+1}$ as $\widehat{\mathbf{x}}$ in this section.
Subsequently, using the accumulated $N$ frames, the neural network predicts the velocities, whose $M$-th output is then incorporated into the ESKF to update the state.
Although a time difference corresponding to the LiDAR scan interval exists, it is limited to 2--9\,ms because the IMU operates at 100--400\,Hz and the difference is thus considered negligible.
We define the measurement model along degenerate directions as follows:
\begin{align}   
    {}^B\mathbf{V}_\mathcal{D} &= \sum_{q \in \boldsymbol{\mathbf{Q}}_\mathcal{D}} {}^B\bm{v}_{q} {}^B\bm{v}_{q}^\top, \label{eq26}\\
    \bm{e}_\mathcal{D} &=  {}^B\mathbf{V}_\mathcal{D}({}^B\mathbf{v}^\text{NN}_M - {}^W\mathbf{\widehat{R}}^{\top} {}^W\mathbf{\widehat{v}}), \label{eq27}
\end{align}
where ${}^B\bm{v}_q$ is obtained by rotating ${}^W\bm{v}_q$, obtained in the final iteration of the IESKF step, in the body frame, and ${}^B\mathbf{V}_\mathcal{D}$ denotes the projection matrix onto the body-frame degenerate space.
Consequently, the error state can be defined in the degenerate subspace.
Moreover, the Jacobian matrix $\mathbf{J}_\mathcal{D}$ is computed as
\begin{equation*}
    \mathbf{J}_\mathcal{D} = {}^B\mathbf{V}_\mathcal{D}\begin{bmatrix}
        {}^W\widehat{\mathbf{R}}^\top[{}^W\widehat{\mathbf{v}}]_\times & \mathbf{0}_{3\times 3} & {}^W\widehat{\mathbf{R}}^\top & \mathbf{0}_{3 \times 9}
    \end{bmatrix} \in \mathbb{R}^{3 \times 18},
\end{equation*} 
where $[\cdot]_\times$ denotes the skew-symmetric operator.

\begin{algorithm}[t]
\small
    \caption{Degenerate-Aware State Estimation}
	\label{alg:degeneracy_aware_optim}
    \SetAlgoLined
    \SetKwInOut{Input}{input}
    \SetKwInOut{Output}{output}
    \DontPrintSemicolon
	\Input{Propagated state $\widehat{\mathbf{x}}$;
    propagated covariance $\widehat{\mathbf{P}}$;
    voxelized scan $\mathcal{V}$;
    local voxel map $\mathbf{M}$;
    convergence threshold $\epsilon$;
    inertial buffer $\mathcal{B}_I$;
    newly measured IMU ${}^B\mathbf{a}_{m, 1:K}, {}^B\mathbf{w}_{m, 1:K}$.}
	\Output{Updated state $\overline{\mathbf{x}}$;
    updated covariance $\overline{\mathbf{P}}$.}
    
    \tcp*{\textcolor{blue}{Constrained State Update in IESKF.
    (Upper module in Fig.~\ref{fig:pipeline})}}
    
    $\widehat{\mathbf{x}}^{0} \gets \widehat{\mathbf{x}}$; \quad $\kappa \gets -1$;
    
    \Repeat{$\| \delta \mathbf{x}^{\kappa+1}_{\kappa}\|_{2} < \epsilon$}{
		$\kappa \gets \kappa + 1$;

		$\{{}^W\mathbf{z}_{p}^\kappa,\, {}^W\mathbf{J}_{p}^\kappa,\, \mathbf{R}_{p}^\kappa\}_{p=1}^{N_\text{pts}} 
		\gets \texttt{ComputePointToPlane}(\mathcal{V}, \mathbf{M}, \widehat{\mathbf{x}}^\kappa)$
		
		$\mathbf{K}^\kappa \gets \left(^W\mathbf{J}^{\kappa^\top} \mathbf{R}^{\kappa^{-1}} {}^W\mathbf{J}^\kappa + \widehat{\mathbf{P}}^{^{-1}}\right)^{-1} {}^W{\mathbf{J}^{\kappa}}^\top \mathbf{R}^{\kappa^{-1}}$

        $\delta \mathbf{x}^{\kappa+1}_{\kappa} \gets \left(\!-\mathbf{K}^\kappa {}^W\mathbf{z}^\kappa \!\!-\!(\mathbf{I} \!-\! \mathbf{K}^\kappa {}^W\mathbf{J}^\kappa)(\mathbf{A}^\kappa)^{\!-1}\!(\widehat{\mathbf{x}}^\kappa \!\boxminus\! \widehat{\mathbf{x}}) \right)$

        $({}^W\boldsymbol{\mathcal{D}}, \ {}^W\mathbf{V}) \gets \texttt{DetectDegeneracy}({}^W\mathbf{J}^\kappa)$
        
        \If{$\operatorname{any}({}^W \boldsymbol{\mathcal{D}}(q) == \textnormal{``Degenerate"})$}{
           $\delta \mathbf{x}^{\kappa+1}_{\kappa} \gets \texttt{AntiDegeneration}(\delta \mathbf{x}^{\kappa+1}_{\kappa}, {}^W\mathbf{V}, \widehat{\mathbf{P}})$
        }
		
		$\widehat{\mathbf{x}}^{\kappa+1} \!\gets\! \widehat{\mathbf{x}}^\kappa \boxplus \delta \mathbf{x}^{\kappa+1}_{\kappa}$\!
	}
    $\widehat{\mathbf{x}} \gets \widehat{\mathbf{x}}^{\kappa+1}$;\quad $\widehat{\mathbf{P}} \gets (\mathbf{I} - \mathbf{K}^\kappa\, {}^W\mathbf{J}^\kappa) \, \widehat{\mathbf{P}}$

    \tcp*{\textcolor{orange}{Degenerate State Update in ESKF.
    (Lower module in Fig.~\hyperref[fig:pipeline]{\textcolor{orange}{\ref*{fig:pipeline}}})}}
    $({}^B\mathbf{a}_{1:K}, \ {}^B\mathbf{w}_{1:K}) \gets \texttt{CompensateIMU}({}^B\mathbf{a}_{m, 1:K}, {}^B\mathbf{w}_{m, 1:K}, \widehat{\mathbf{x}})$

    ${}^W\widehat{\mathbf{R}}_{1:K} \gets \texttt{Interpolate}(\widehat{\mathbf{x}}, \mathcal{B}_I)$
    
    $\texttt{UpdateBuffer}({}^B\mathbf{a}_{1:K}, {}^B\mathbf{w}_{1:K}, ^W\widehat{\mathbf{R}}_{1:K}, \mathcal{B}_I)$
    
    \eIf {$\operatorname{any}({}^W\boldsymbol{\mathcal{D}}(q) == \textnormal{``Degenerate"})$}{
        $({}^B\mathbf{a}_{1:N}, \ {}^B\mathbf{w}_{1:N}, {}^W\widehat{\mathbf{R}}_{1:N}) \gets \mathcal{B}_I$

        ${}^B\boldsymbol{\xi}_{1:N} \gets \mathrm{Log}({}^W\widehat{\mathbf{R}}_{N}^\top{}^W\widehat{\mathbf{R}}_{1:N})$

        $({}^B\mathbf{v}^\text{NN}_M, \boldsymbol{\sigma}^\text{NN}_M) \gets f_\theta({}^B\mathbf{a}_{1:N}, \ {}^B\mathbf{w}_{1:N},{}^B\boldsymbol{\xi}_{1:N})$

        $(\overline{\mathbf{x}}, \overline{\mathbf{P}}) \gets \texttt{ESKF}
        ({}^B\mathbf{v}^\text{NN}_M, \boldsymbol{\sigma}^\text{NN}_M, \widehat{\mathbf{x}}, \widehat{\mathbf{P}},{}^W\boldsymbol{\mathcal{D}},{}^{W}\mathbf{V})$
    }{
        $\overline{\mathbf{x}}\gets \widehat{\mathbf{x}},\, \overline{\mathbf{P}}\gets \widehat{\mathbf{P}}$;
    }
	\Return{$\overline{\mathbf{x}}, \overline{\mathbf{P}}$}.
\end{algorithm}
Furthermore, we need to determine the measurement noise $\boldsymbol{\Sigma}_v$.
Because the state covariance $\mathbf{P}$ is affected by the scale of LiDAR measurement and IMU process noise, we consider $\boldsymbol{\Sigma}_v = \sigma_v \mathbf{\Sigma}^\text{NN}_{v, M}$ with a scaling factor $\sigma_v$.
Subsequently, this measurement noise is projected onto the degenerate space.
However, ${}^B\mathbf{V}_\mathcal{D}$ is singular because of its rank deficiency, which leads to problems when computing its inverse in the ESKF.
To prevent singularity, an additional term is introduced in the non-degenerate subspace, following the Levenberg--Marquardt regularization scheme:
\begin{equation}    \label{eq29}
    \boldsymbol{\Sigma}_{\mathcal{D}} = {}^B\mathbf{V}_\mathcal{D} \boldsymbol{\Sigma}_v {}^B\mathbf{V}_\mathcal{D}^\top + (\mathbf{I}_3-{}^B\mathbf{V}_\mathcal{D})(\sigma \mathbf{I}_3)(\mathbf{I}_3-{}^B\mathbf{V}_\mathcal{D})^\top, 
\end{equation} 
where, because ${}^B\mathbf{V}_\mathcal{D}$ is involved with the body-frame degenerate space, the projection matrix onto the non-degenerate space can be expressed as $\mathbf{I}_3 - {}^B\mathbf{V}_\mathcal{D}$, and we set $\sigma = \sigma_v$.
Then, the ESKF updates the state as follows:
\begin{equation}    \label{eq30}
    \begin{aligned} 
        \mathbf{K} =\ & \widehat{\mathbf{P}}\mathbf{J}_\mathcal{D}^\top(\mathbf{J}_\mathcal{D} \widehat{\mathbf{P}} \mathbf{J}_\mathcal{D}^\top + \boldsymbol{\Sigma}_{ \mathcal{D}})^{-1}, \\
        \overline{\mathbf{x}} =\  &\widehat{\mathbf{x}}\  \boxplus \ \mathbf{K} \bm{e}_{\mathcal{D}}, \\
        \overline{\mathbf{P}} =\  & (\mathbf{I}_{18} - \mathbf{KJ}_\mathcal{D})\widehat{\mathbf{P}}.
    \end{aligned}
\end{equation}

In this way, all state components are updated by considering the uncertainties from both the system and the neural network, thereby alleviating the effects of degeneracy and ensuring system stability.
The overall degeneracy-aware procedure described here is summarized in Algorithm~\ref{alg:degeneracy_aware_optim}, and further details of the LIO pipeline can be found in~\cite{fast_lio2, pv_lio}.

\section{Performance Evaluation} \label{sec:4_evaluation}
Experiments were conducted to evaluate the effectiveness of the proposed method and validate our key claims:
(i)~Our approach shows strong robustness across various platforms and degenerate environments, achieving the lowest error on the majority of the evaluated degenerate sequences, including against methods specifically designed for degeneracy robustness (Section~\ref{subsec:42_robustness})
(ii)~By integrating deep-learning–based velocity into the ESKF, consistent updates of the full state variables are achieved, resulting in enhanced performance (Section~\ref{subsec:43_eval_state_update}).
(iii)~Our method maintains reliable odometry under simultaneously geometrically and photometrically sparse conditions, where visual measurements also degrade (Section~\ref{subsec:44_comparison_lvio}).
(iv)~By utilizing state estimates from LIO, the generalization capability of the neural network is improved (Section~\ref{subsec:45_impact_deep_lio_output}).

Additional evaluations, including detailed neural network performance, real-time analysis, and a sensitivity analysis of the scaling factor $\sigma_v$, are provided in Sections~\ref{appendix:neural_network_evaluation}, \ref{appendix:real_time_performance}, and~\ref{appendix:sigma_ablation}, respectively.

\subsection{Experimental Setup}
Following AirIO~\cite{airio}, we utilized IMU data collected over 5\,s.
Datasets using a Livox LiDAR sensor cover various platforms, including handheld devices, drones, and unmanned ground vehicles, with a 200\,Hz built-in IMU, resulting in $N=1{,}000$.
Other datasets were collected using cars as platforms, equipped with onboard IMUs operating at 100, 200, or 400\,Hz, with $N=500$, $1{,}000$, or $2{,}000$, respectively.
In addition, $M$ was determined by the neural network, with $M= 56$, $112$, or $223$ for $N = 500$, $1{,}000$, or $2{,}000$, respectively.
The parameters of our system are summarized in Table~\ref{tab:parameter}, and except for $\sigma_v$, all other parameters were set to the same values as in~\cite{airio,x_icp}. The LiDAR configuration parameters, such as the voxel and downsampling sizes, follow~\cite{genzlio}.
To train the neural network, we used open datasets and our private datasets, and GT was constructed in well-conditioned environments using PV-LIO, as mentioned in Section~\ref{subsec:32_neural_network}.
Furthermore, datasets collected with a Livox LiDAR sensor were used to train a single model.
For cars, a separate model was trained for each dataset because of their limited degrees of freedom (DoF) and dependence on the IMU mounting pose.
\begin{table*}[t]
\centering
\small
\caption{Performance of LiDAR-Based Odometry on Degenerate Scenarios (ATE $\downarrow$ [m]).}
\label{tab:lio_evaluation_ape}
\begin{tabular}{l|>{\centering\arraybackslash}p{1.3cm} >{\centering\arraybackslash}p{1.3cm} >{\centering\arraybackslash}p{1.3cm} >{\centering\arraybackslash}p{1.3cm} >{\centering\arraybackslash}p{1.3cm} >{\centering\arraybackslash}p{1.3cm} >{\centering\arraybackslash}p{1.3cm} >{\centering\arraybackslash}p{1.3cm} >{\centering\arraybackslash}p{1.3cm}}
\toprule[1pt]
\textbf{Sequence} & \makecell{GenZ- \\ ICP\\\cite{genzicp}} & \makecell{CV \\ w/ X-ICP\\ \cite{x_icp}}  & \makecell{FAST- \\ LIO2\\\cite{fast_lio2}} &  \makecell{D-LIO\\\cite{dlio}}  & \makecell{PV-LIO\\\cite{pv_lio}}& \makecell{PV-LIO \\ w/ CO\\\cite{relead}}  &
\makecell{Ours \\ w/o CO} & \makecell{Ours} \\
\midrule
\rowcolor{gray!10}\textbf{Car}   &                     &         &               &              &                      &              &                       & \\
\texttt{vehicle\_highway\_0}     & $\times$            & $\times$       & 1085.34       & \tcol 943.08 & $\times$                    & $\times$            & \scol 377.58          & \fcol \textbf{375.75} \\
\texttt{vehicle\_tunnel\_0}      & 510.43              & $\times$       & 2378.02       & $\times$            & \tcol387.25          & 403.56       & \scol 56.64           & \fcol \textbf{53.05}  \\
\texttt{HK-CHTunnel}             & 730.45              & 710.53  & \tcol 656.58  & $\times$            & $\times$                    & $\times$            & \scol 153.59          & \fcol \textbf{100.85} \\
\texttt{HK-Whampoa}             & \fcol \textbf{2.48} & $\times$       & 5.63          & 6.99         & 5.22                 & 5.59         & \tcol 4.23            & \scol 3.87 \\
\texttt{Urban\_Tunnel\_1}        & $\times$                   & $\times$       & $\times$             & $\times$            & $\times$                    & $\times$            & $\times$                     & $\times$ \\
\texttt{Urban\_Tunnel\_2}        & 383.59              & $\times$       & 436.84        & 258.46       & 137.27               & \tcol 127.86 & \scol 106.41          & \fcol \textbf{104.50}  \\
\texttt{Urban\_Tunnel\_3}        & 344.55              & $\times$       & 1164.30       & $\times$            & \fcol \textbf{81.92} & \tcol 86.79  & \scol 84.81           & 88.21  \\
\texttt{Bridge\_2}               & 738.79              & 539.65  & 1323.73       & $\times$            & 137.15               & \tcol 134.62 & \scol 111.54          & \fcol \textbf{109.44} \\
\texttt{Bridge\_3}               & \tcol 373.40        & $\times$       & 572.04        & 382.29       & $\times$                    & $\times$            & \fcol \textbf{109.94} & \scol 110.49 \\
\midrule
Average                             &      --             &  --     &          --   &   --         &  --                  & --           & \scol 125.59          & \fcol \textbf{118.27}  \\
\midrule
\rowcolor{gray!10}\textbf{Others}&     &    &              &        &    &   &                      &             \\
\texttt{ShieldTunnel\_1}         & $\times$   & $\times$  & \tcol 89.88  & $\times$      & $\times$   & $\times$ & \fcol \textbf{12.67} & \scol 12.69 \\
\texttt{ShieldTunnel\_2}         & $\times$   & $\times$  & \tcol 100.86 & 196.38 & $\times$   & $\times$ & \fcol \textbf{21.60} & \scol 22.95 \\
\texttt{ShieldTunnel\_3}         & $\times$   & $\times$  & \tcol 72.64  & 134.60 & $\times$   & $\times$ & \scol 12.68          & \fcol \textbf{10.73} \\
\texttt{ShieldTunnel\_4}         & $\times$   & $\times$  & $\times$            & $\times$      & $\times$   & $\times$ & \fcol \textbf{21.84} &\scol  21.98 \\
\texttt{ShieldTunnel\_5}         & $\times$   & $\times$  & $\times$            & $\times$      & $\times$   & $\times$ & \fcol \textbf{6.55}  & \scol 8.73  \\
\texttt{ShieldTunnel\_6}         & $\times$   & $\times$  & \tcol 37.08  & $\times$      & $\times$   & $\times$ & \scol 6.40           & \fcol \textbf{4.24}  \\
\texttt{FlatGroundSmooth}        & $\times$   & $\times$  & $\times$            & $\times$      & $\times$   & $\times$ & $\times$                    & $\times$           \\
\texttt{FlatGroundAggressive}     & $\times$   & $\times$  & $\times$            & $\times$      & $\times$   & $\times$ & $\times$                    & $\times$           \\
\midrule
Average                             &  -- &  --&    --        &  --    &  --& -- & \scol 13.62          & \fcol \textbf{13.55} \\
\bottomrule[1pt]
\multicolumn{9}{l}{\footnotesize ``$\times$'': complete divergence (velocity grows excessively or the estimate stalls); ``$-$'': average not computed (excluding sequences in which all methods diverged).}\\
\multicolumn{9}{r}{\footnotesize Best results are marked as \ftext{\textbf{first}}, \stext{second}, and \ttext{third}.}\\
\end{tabular}
\vspace{-2mm}
\end{table*}
\begin{figure*}[t!]
    \centering
    \vspace{-1mm}
    \includegraphics[width=0.7\linewidth]{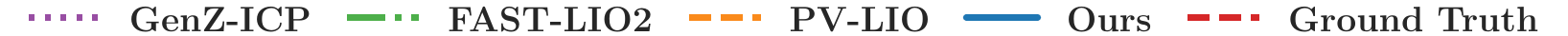} \\[1mm]
    \subfigure[\texttt{vehicle\_tunnel\_0}] {\includegraphics[width=0.237\linewidth]{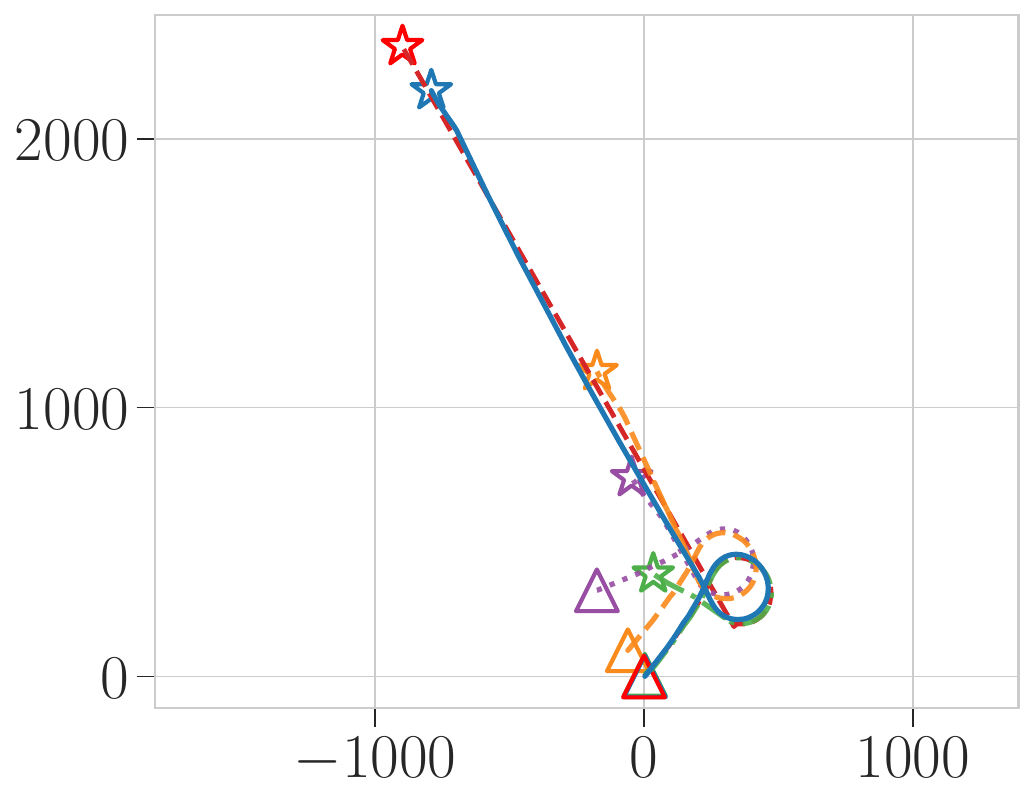}}
    \subfigure[\texttt{HK-CHTunnel}] {\includegraphics[width=0.25\linewidth]{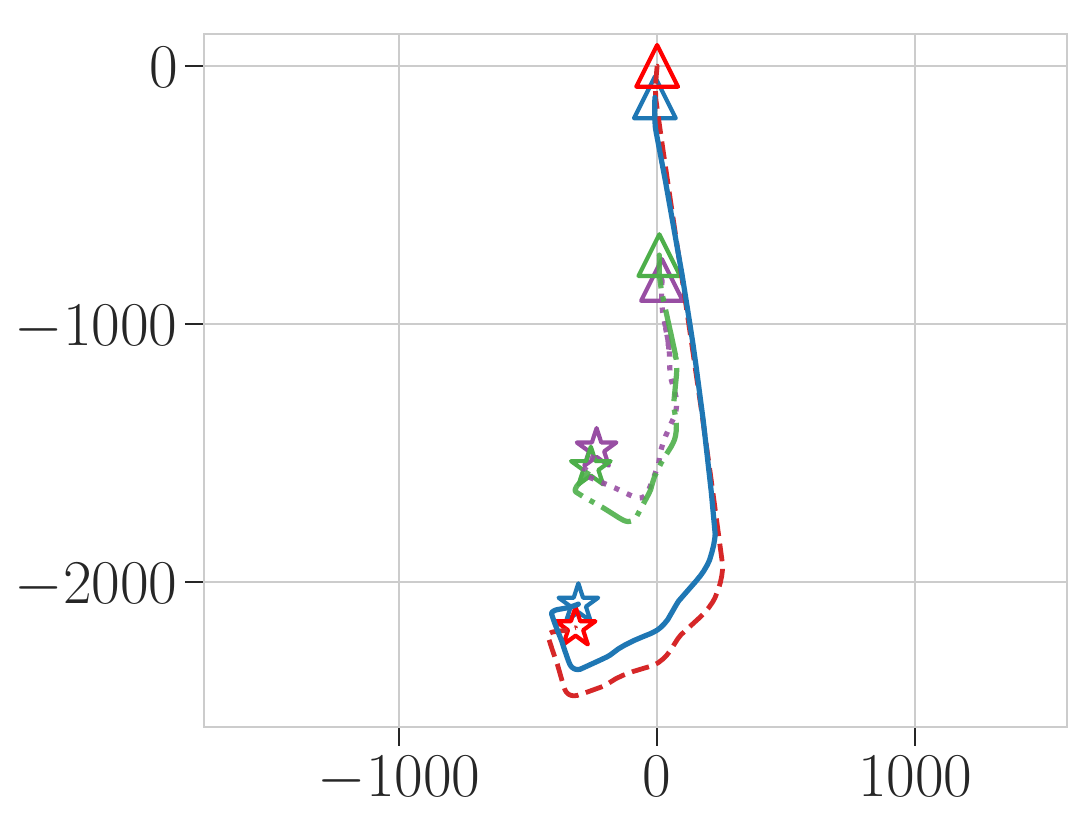}}
    \subfigure[\texttt{Urban\_Tunnel\_3} ]{\includegraphics[width=0.25\linewidth]{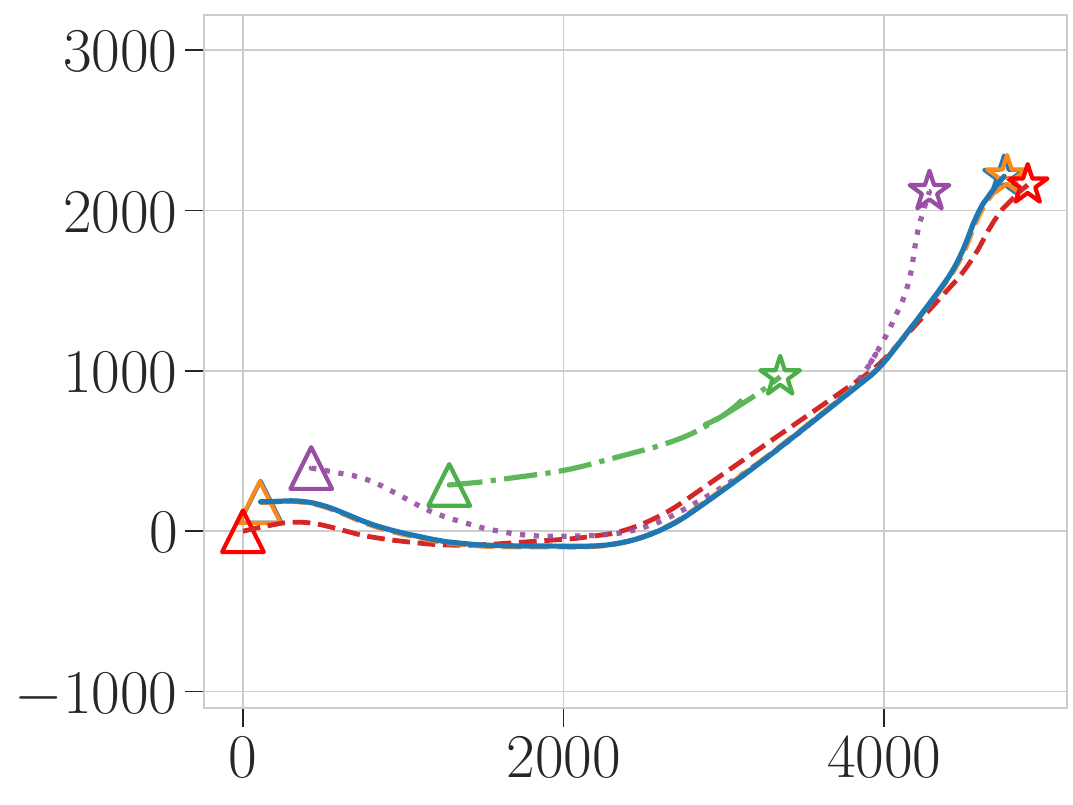}}
    \subfigure[\texttt{Bridge\_2}]{\includegraphics[width=0.237\linewidth]{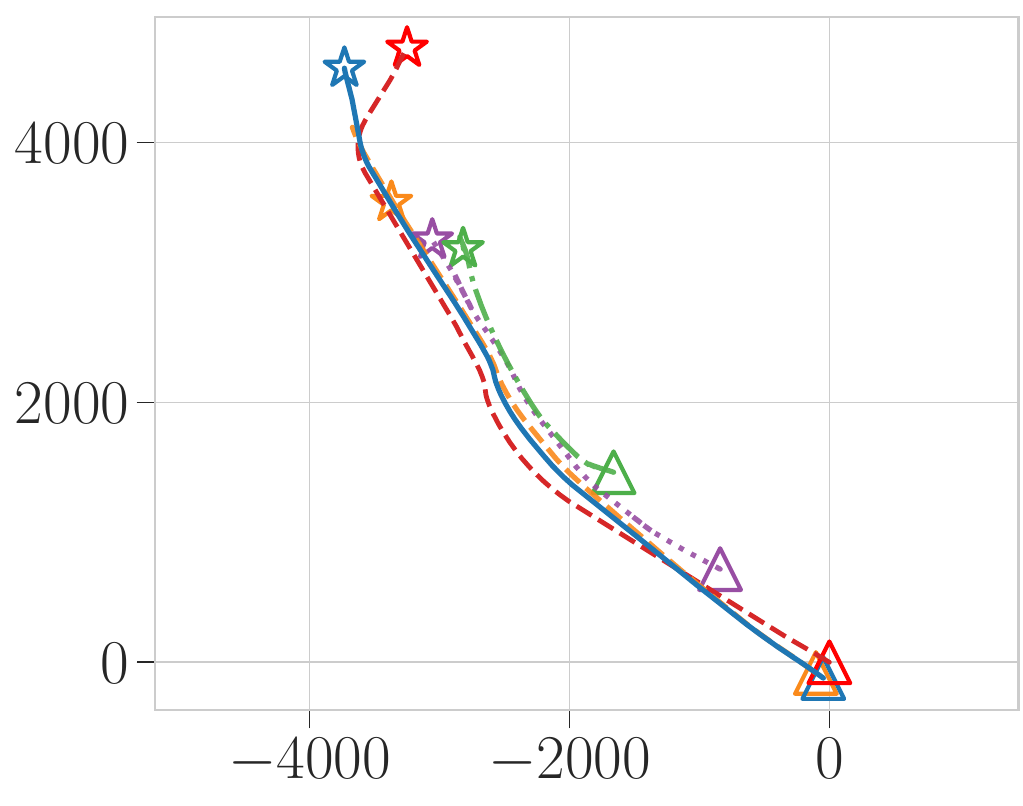}}

    \caption{Vehicle motion: Estimated trajectory visualization under degenerate conditions (X–Y plane in meters). The symbols $\triangle$ and ☆ denote the start and end points, respectively, of each trajectory.}
    \label{fig:lio_output_vehicle}
    \vspace{-5mm}
\end{figure*}

To evaluate the training performance, a subset of sequences from the well-conditioned dataset was used as the evaluation set, and the entire well-conditioned dataset was used to train the model before its deployment in our LIO framework.
Additionally, we constructed additional degeneracy datasets using our handheld device.
Our training, evaluation, and degeneracy datasets are summarized in Table~\ref{tab:dataset}.
To evaluate the neural network, we used the absolute velocity error (AVE) and absolute translation error (ATE) as evaluation metrics.
To evaluate the LIO system, datasets with GT were assessed using ATE via EVO~\cite{evo}.
For vehicle trajectories, the results were projected onto the X-Y plane.
For datasets without GT, including our private datasets, the end-to-end error was used as the evaluation metric, which quantifies the drift as the discrepancy between the estimated start and end poses of a trajectory whose start and end coincide at the same physical location.
Details of our handheld setup can be found in~\cite{park2025dataset}, and our private datasets are described in Section~\ref{appendix:private_dataset_detail}.

\subsection{Robustness under Degeneracy on Various Platforms}\label{subsec:42_robustness}
\begin{table*}[t]
\centering
\small
\caption{Performance of LiDAR-Based Odometry on Degenerate Scenarios (End-to-End Error $\downarrow$ [m]).}
\label{tab:lio_evaluation_e2e}
\begin{tabular}{l|>{\centering\arraybackslash}p{1.3cm} >{\centering\arraybackslash}p{1.3cm} >{\centering\arraybackslash}p{1.3cm} >{\centering\arraybackslash}p{1.3cm} >{\centering\arraybackslash}p{1.3cm} >{\centering\arraybackslash}p{1.3cm} >{\centering\arraybackslash}p{1.3cm} >{\centering\arraybackslash}p{1.3cm} >{\centering\arraybackslash}p{1.3cm}}
\toprule[1pt]
\textbf{Sequence} & \makecell{GenZ- \\ ICP\\\cite{genzicp}} & \makecell{CV \\ w/ X-ICP\\ \cite{x_icp}}  & \makecell{FAST- \\ LIO2\\\cite{fast_lio2}} &  \makecell{D-LIO\\\cite{dlio}}  & \makecell{PV-LIO\\\cite{pv_lio}}& \makecell{PV-LIO \\ w/ CO\\\cite{relead}}  &
\makecell{Ours \\ w/o CO} & \makecell{Ours}\\
\midrule
\texttt{degenerate\_seq\_0}          & $\times$      & 23.66 & 7.53                & 8.89  & \tcol 5.11          & 5.89                & \scol 2.26           & \fcol \textbf{2.04}\\
\texttt{degenerate\_seq\_1}          & 10.15  & 27.60 & 6.14                & 37.27 & 3.50                & \tcol 1.98          & \fcol \textbf{0.01}  & \scol 0.10\\
\texttt{degenerate\_seq\_2}          & $\times$      & 11.36 & 19.32               & $\times$     & 12.59               & \tcol 9.03          & \scol 1.38           & \fcol \textbf{0.38}\\
\texttt{LiDAR\_Degenerate}          & 11.5   & $\times$     & 6.09                & 6.03  & \scol 0.02          & 2.07                & \fcol \textbf{0.01}  & \tcol 0.64\\
\texttt{Bright\_Screen\_Wall}       & $\times$      & $\times$     & \fcol \textbf{0.02} & $\times$     & \scol 0.03          & \tcol 0.69          & 2.01                 & 1.26\\
\texttt{CBD\_Building\_2}           & 17.66  & $\times$     & 1.95                & $\times$     & \fcol \textbf{0.01} & \tcol 1.66          & \scol 0.02           & \fcol \textbf{0.01}\\
\texttt{CBD\_Building\_3}           & 32.49  & $\times$     & \tcol 11.86         & $\times$     & \fcol \textbf{0.01} & \fcol \textbf{0.01} & \scol 0.83           & \fcol \textbf{0.01}\\
\texttt{HIT\_Graffiti\_Wall\_1}     & $\times$      & $\times$     & \tcol 6.71          & $\times$     & \fcol 0.10          & \scol 0.20          & 32.87                & 19.52\\
\texttt{HIT\_Graffiti\_Wall\_2}     & $\times$      & $\times$     & $\times$                   & $\times$     & $\times$                   & \tcol 83.13          & \fcol \textbf{11.50} & \scol 11.76\\
\texttt{HIT\_Graffiti\_Wall\_3}     & $\times$      & $\times$     & $\times$                   & $\times$     & $\times$                   & $\times$                   & \scol 17.84          & \fcol \textbf{17.08}\\
\texttt{HIT\_Graffiti\_Wall\_4}     & $\times$      & $\times$     & \tcol 3.01          & $\times$     & \scol 1.15          & \scol 1.15          & \fcol \textbf{0.42}  & 5.67\\
\texttt{HKU\_Cultural\_Center\_1}   & $\times$      & $\times$     & \tcol 0.09          & $\times$     & \fcol \textbf{0.02} & \fcol \textbf{0.02} & \scol 0.03           & \scol 0.03\\
\texttt{HKU\_Cultural\_Center\_2}   & 15.68  & $\times$     & 5.50                & $\times$     & \fcol \textbf{0.03} & \fcol \textbf{0.03} & \tcol 0.90           & \scol  0.88\\
\midrule
Average                                &   --  & --  & --  & --  & --  & --   & \tcol 5.40 & \fcol \textbf{4.57} \\
\bottomrule[1pt]

\end{tabular}
\vspace{-4mm}
\end{table*}
We evaluated the performance of the proposed method and comparison baselines to support the claim that our approach effectively mitigates degeneracy and enhances the robustness of the LIO system under various degenerate scenarios.
For comparison, we adopted several LIO algorithms: FAST-LIO2~\cite{fast_lio2} and PV-LIO~\cite{pv_lio}, which employ the point-to-plane error metric; and D-LIO~\cite{dlio}, which is based on the Generalized ICP.
Furthermore, we also included algorithms designed to be robust to degeneracy, namely GenZ-ICP~\cite{genzicp} and X-ICP~\cite{x_icp}.
Specifically, X-ICP relies on a joint encoder and IMU to obtain legged odometry as the pose prior, which limits its applicability across different platforms.
For fair comparison, we re-implemented GenZ-ICP to use only the point-to-plane error metric, performing degeneracy detection and constrained optimization in the same manner as X-ICP.
In this implementation, a constant velocity (CV) model was adopted as the pose prior, and we refer to this method as CV w/ X-ICP.
COIN-LIO~\cite{coin_lio} is not suitable for narrow field-of-view (FOV) LiDARs such as the Livox Avia and was therefore not included in the comparison.
Furthermore, constrained optimization (CO), as formulated in (\ref{eq22})--(\ref{eq25}), was applied in PV-LIO.
To analyze the impact of CO, we also evaluated a variant of our system without CO.

\subsubsection{Evaluation on Car-Mounted Platforms}
As listed in the upper part of Table~\ref{tab:lio_evaluation_ape}, either variant of our method (with or without CO) achieves the lowest ATE in all but two of the non-diverged car sequences.
Furthermore, Fig.~\ref{fig:lio_output_vehicle} shows that the endpoint estimated by our method was closer to GT than those produced by the other methods.
In contrast, the ICP algorithms rapidly failed in odometry estimation once degeneracy occurred. Although the LIO algorithms relied on IMU motion, their estimates either diverged or lagged as errors accumulated over time.
In contrast, our approach successfully navigated these challenging scenarios.
Furthermore, our method achieved enhanced performance on most datasets compared to the version without CO.
We observed that, in these scenarios, other vehicles acting as dynamic obstacles can introduce erroneous constraints, leading to backward or forward drift in the estimated position. By preventing such incorrect updates, the incorporation of CO contributed to the observed performance improvement.
Notably, the impact was most pronounced in \texttt{HK-CHTunnel}, leading to a considerable performance gap of 52.74\,m.
However, in specific sequences, although streetlamps could serve as useful geometric features, CO neglects them, resulting in incorrect state updates and a noticeable performance decline of 3.4\,m, as seen in \texttt{Urban\_Tunnel\_3}.
In some sequences, other methods showed better performance: GenZ-ICP and PV-LIO exhibited the best performance in \texttt{HK-Whampoa} and \texttt{Urban\_Tunnel\_3}, respectively.
In the former sequence, the degeneracy duration was short, and the presence of streetlamps provided constraints, resulting in GenZ-ICP achieving the best performance.
In the latter sequence, the vehicle passed through the tunnel with minimal variation in velocity.
This result was attributed to the accurate horizontal orientation estimates in this sequence, resulting in minimal accumulation of gravity-induced errors in the acceleration measurements.

\begin{figure}[t!]
    \centering
    \vspace{-1mm}
    \includegraphics[width=1\linewidth]{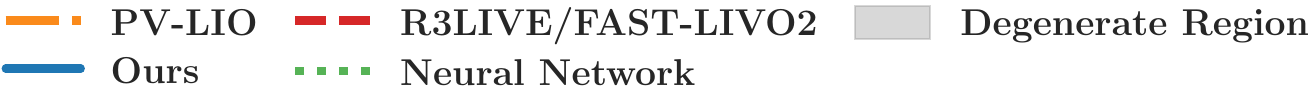} \\[1mm]
    \includegraphics[width=1\linewidth]{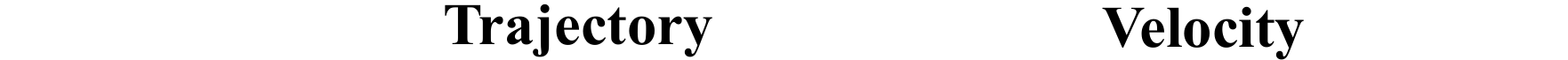} \\[1mm]
    \vspace{-3mm}
    \subfigure[\texttt{degenerate\_seq\_2}]{\includegraphics[width=1.0\linewidth]{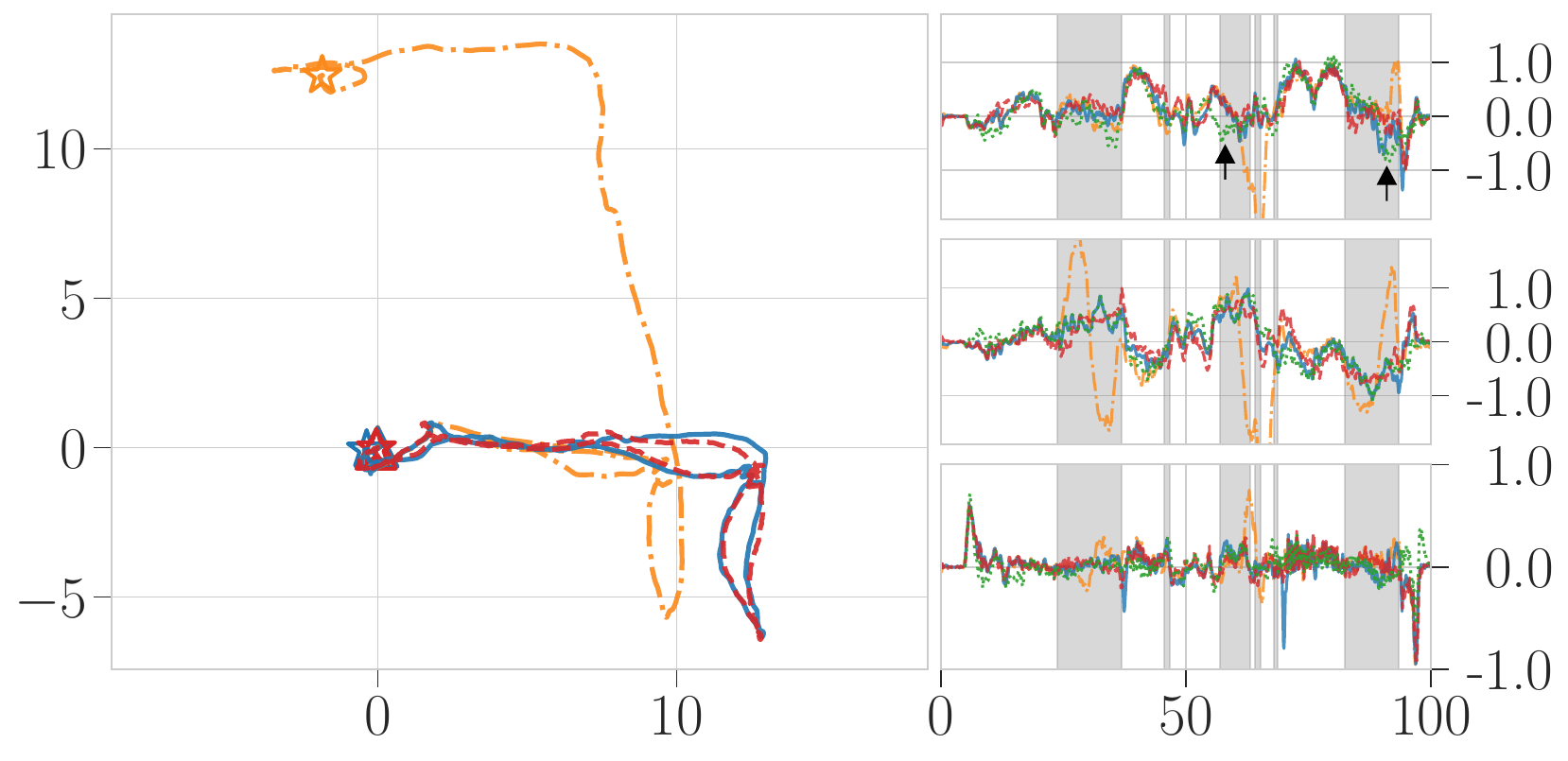}}
    \subfigure[\texttt{HIT\_Graffiti\_Wall\_1}]{\includegraphics[width=1.0\linewidth]{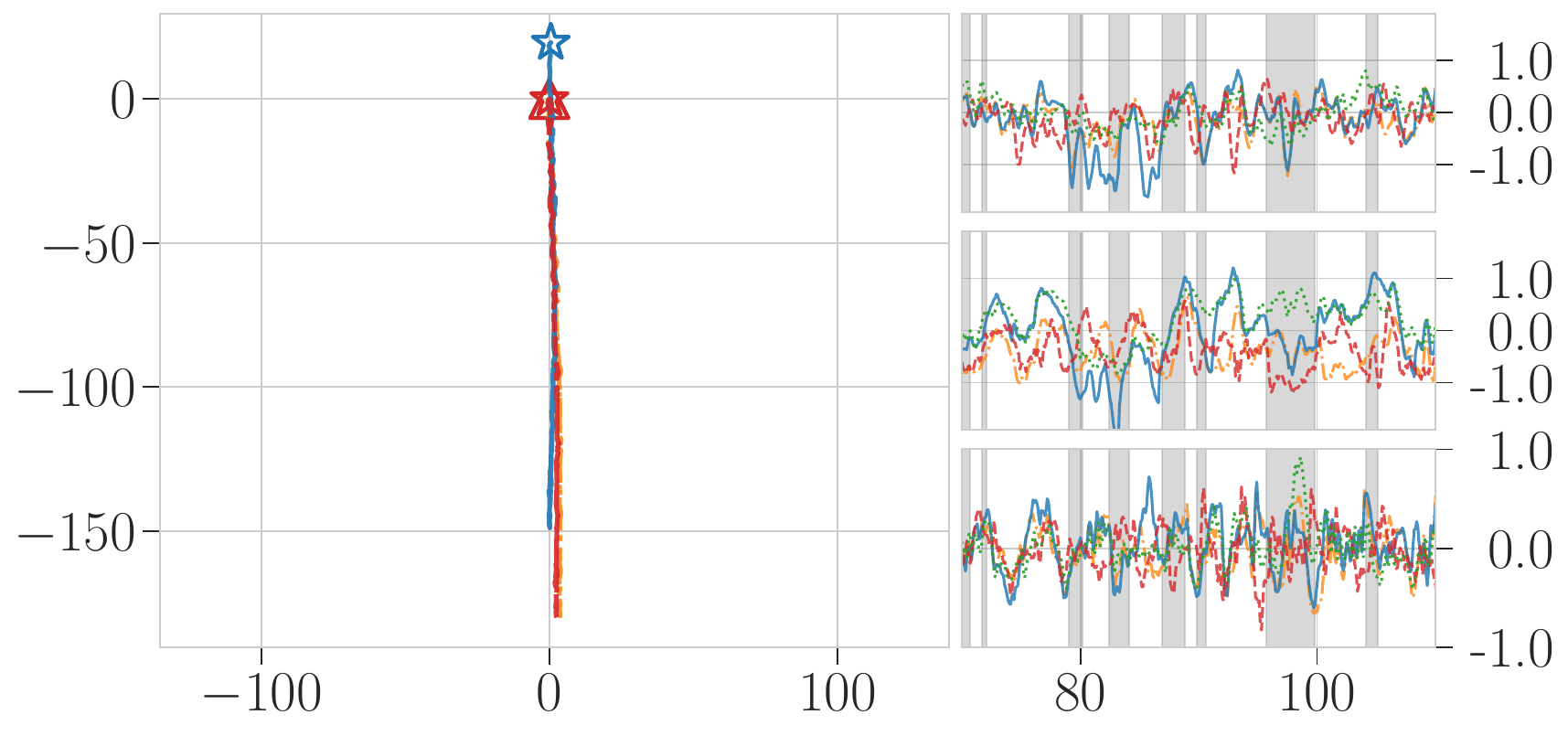}}

    \caption{Handheld motion: Estimated trajectory and velocity visualization on degenerate cases. The left column shows the trajectory (X–Y plane in meters), and the right column presents the body-frame velocities (velocity in meters per second over time in seconds). The symbols $\triangle$ and ☆ denote the start and end points, respectively, of each trajectory.}
    \label{fig:lio_output_handheld}
    \vspace{-6mm}
\end{figure}

\subsubsection{Evaluation on Handheld Platforms}
As listed in the lower part of Table~\ref{tab:lio_evaluation_ape}, during the handheld traversal of \texttt{ShieldTunnel\_1-6}, all algorithms except FAST-LIO2 and D-LIO experienced divergence.
In particular, the ICP methods were affected by the rounded tunnel geometry, resulting in continuous roll rotation.
Although FAST-LIO2 and D-LIO did not diverge, they produced backward state estimates even during forward motion, leading to erroneous odometry.
By contrast, our approach substantially mitigated the effects of degeneracy in these sequences.
Notably, \texttt{ShieldTunnel\_4} involved handheld shaking during traversal, and \texttt{ShieldTunnel\_5} included motions along the tunnel side wall, demonstrating that our method effectively alleviates degeneracy not only during forward motion but also under diverse motion conditions.

Although our method mitigated some challenging scenarios, it still failed to prevent odometry divergence in some scenarios.
In \texttt{Urban\_Tunnel\_1}, the erroneous detection of degeneracy prevented the triggering of ESKF correction, i.e., the system behaved identically to the baseline algorithm.
Across all \texttt{FlatGround} sequences, divergence was avoided, yet the estimated states showed non-negligible discrepancies from the GT.

\subsubsection{Evaluation on Handheld and Drone Platforms Without Ground Truth} \label{subsubsec4b3}
In this experiment, we employed datasets without GT, including our private datasets, and adopted the end-to-end error as the evaluation metric.
First, ICP-based methods rapidly diverged and failed to produce accurate odometry, even over relatively short degenerate durations, as shown in Table~\ref{tab:lio_evaluation_e2e}.
Additionally, D-LIO experienced degraded odometry accuracy in non-degenerate scenarios because of the narrow FOV of the Livox Avia sensor.
By comparison, the results listed in Table~\ref{tab:lio_evaluation_e2e} show that our system remained stable in all sequences.
Remarkably, our method was able to mitigate degeneracy compared to the other approaches in \texttt{degenerate\_seq\_0-2}.
In particular, \texttt{degenerate\_seq\_2} involved lateral motion of the drone while facing a wall, as seen in Fig.~\ref{fig:lio_output_handheld}(a), under which PV-LIO failed to estimate odometry accurately, leading to a substantial end-to-end error.
By contrast, our method produced a trajectory closely matching that of R3LIVE, achieving performance comparable to systems that incorporate vision.
As shown by the estimated velocities on the right of Fig.~\ref{fig:lio_output_handheld}(a), our method aligned closely with R3LIVE, confirming that ${}^B\mathbf{v}^\text{NN}_M$ is quite accurate, even in degenerate regions.
However, ${}^B\mathbf{v}^\text{NN}_M$ was not accurate along all axes.
As observed from the arrows along the $x$-axis on the right side of Fig.~\ref{fig:lio_output_handheld}(a), a slight gap exists because the degenerate direction in this sequence was primarily aligned with the $y$-axis.
By updating solely along the degenerate direction, the limitations of the neural network can be alleviated.

\begin{figure*}[t!]
    \centering
    \vspace{-2mm}
    \includegraphics[width=0.48\linewidth]{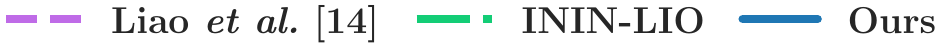} \\[1mm]
    \includegraphics[width=0.497\linewidth]{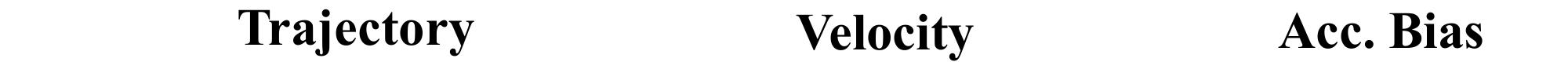} 
    \includegraphics[width=0.497\linewidth]{Figures/traj_title_ablation.pdf} \\[-2mm]
    \subfigure[\texttt{ShieldTunnel\_1}]{\includegraphics[width=0.497\linewidth]{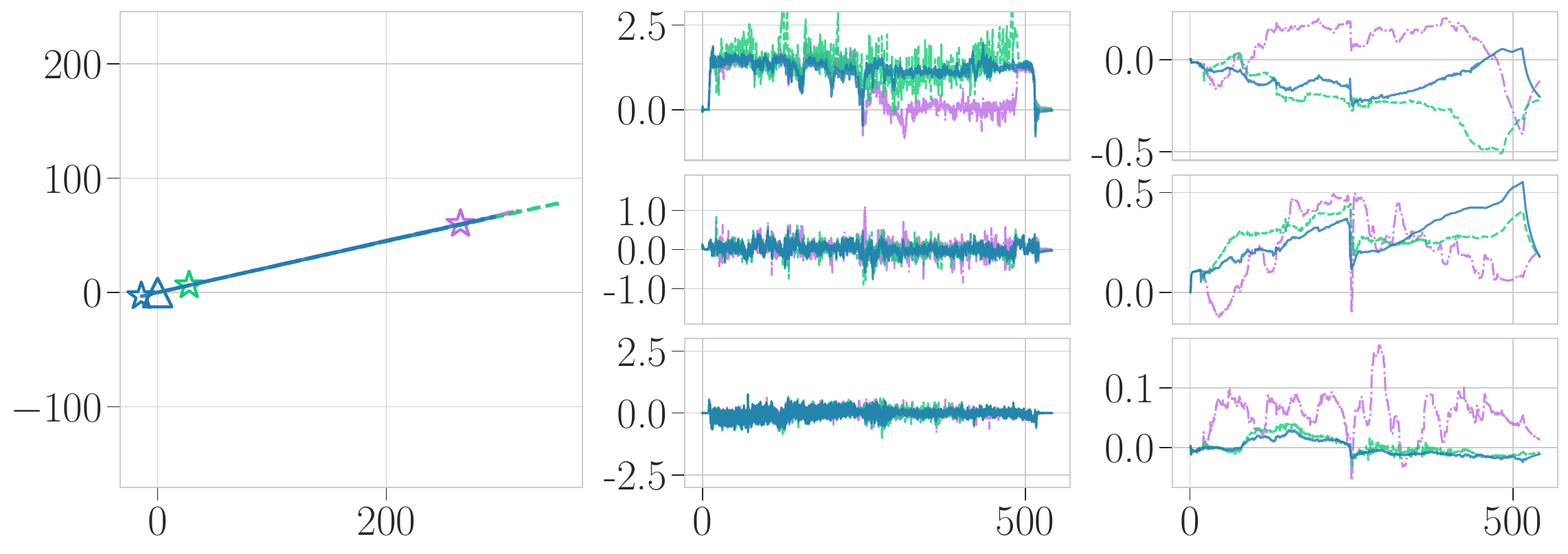}}
    \subfigure[\texttt{ShieldTunnel\_2}]{\includegraphics[width=0.497\linewidth]{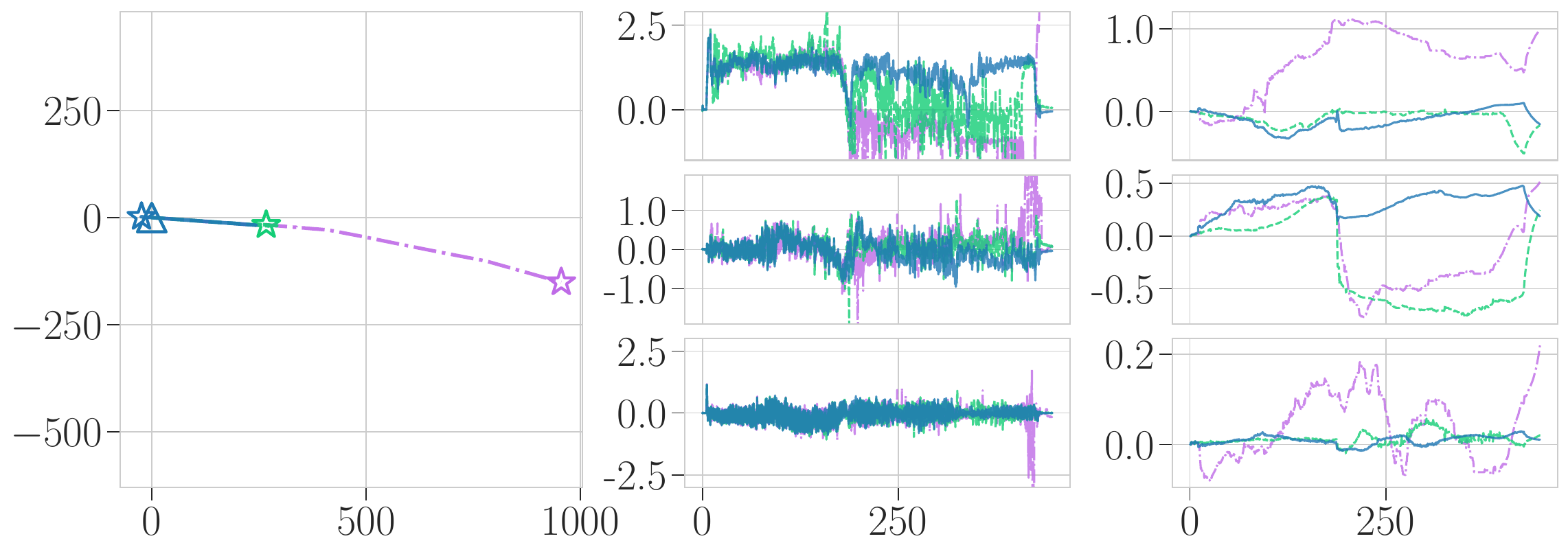}}
    \vspace{-1mm}
    \subfigure[\texttt{ShieldTunnel\_3}]{\includegraphics[width=0.497\linewidth]{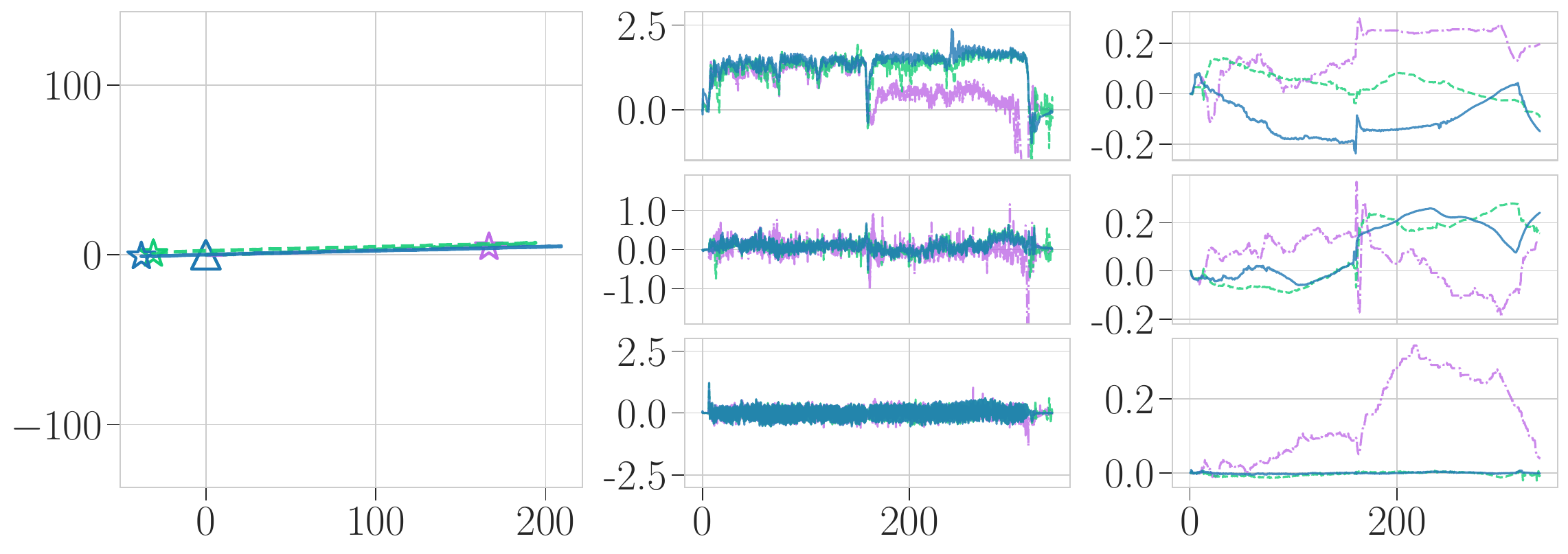}}
    \subfigure[\texttt{ShieldTunnel\_4}]{\includegraphics[width=0.497\linewidth]{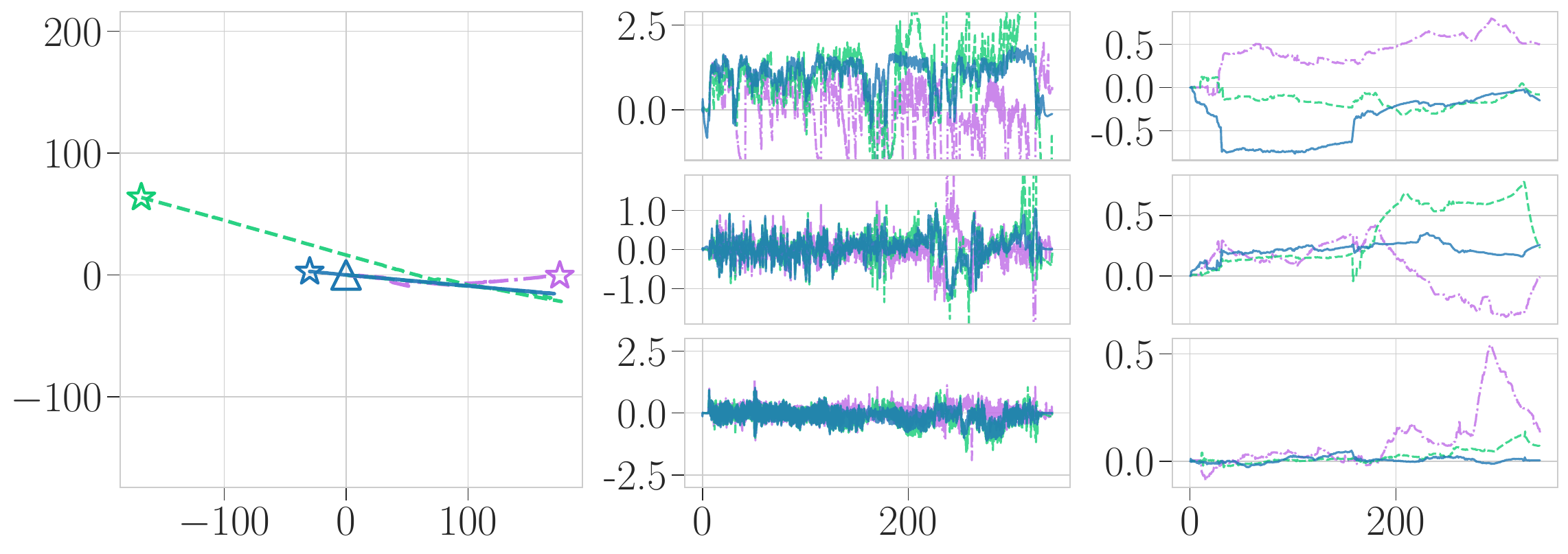}}
    \vspace{-1mm}
    \subfigure[\texttt{ShieldTunnel\_5}]{\includegraphics[width=0.497\linewidth]{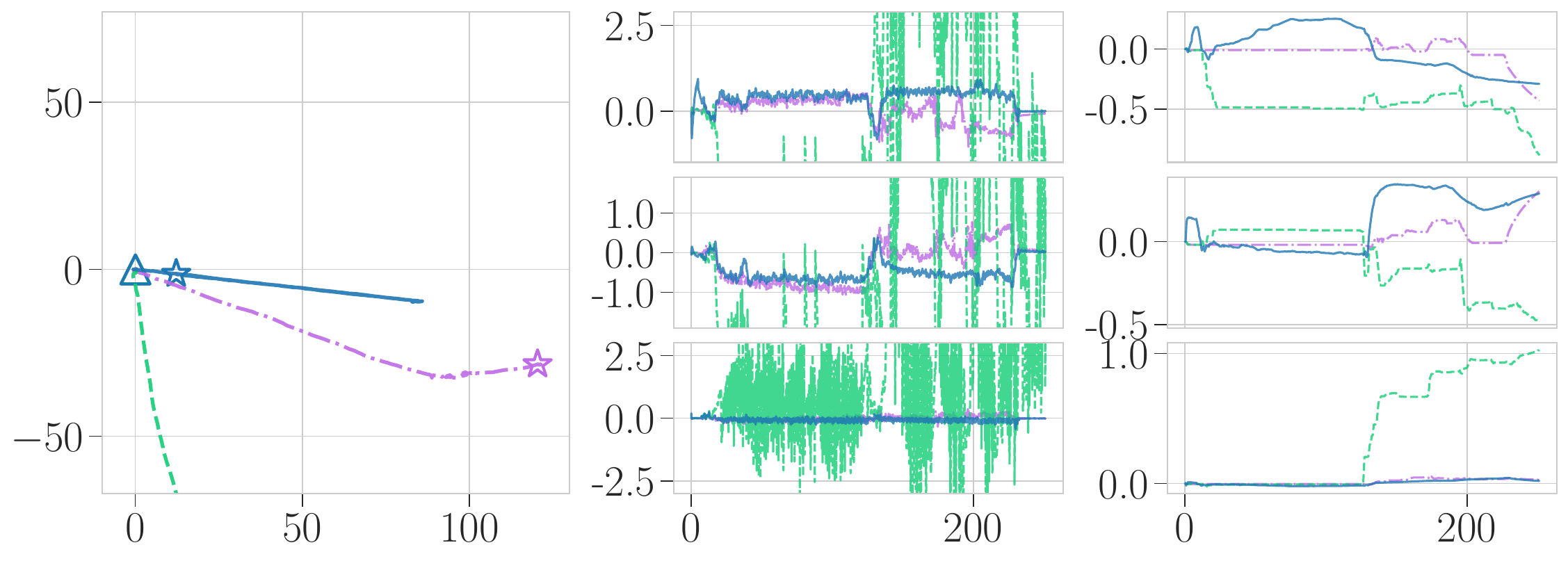}}
    \subfigure[\texttt{ShieldTunnel\_6}]{\includegraphics[width=0.497\linewidth]{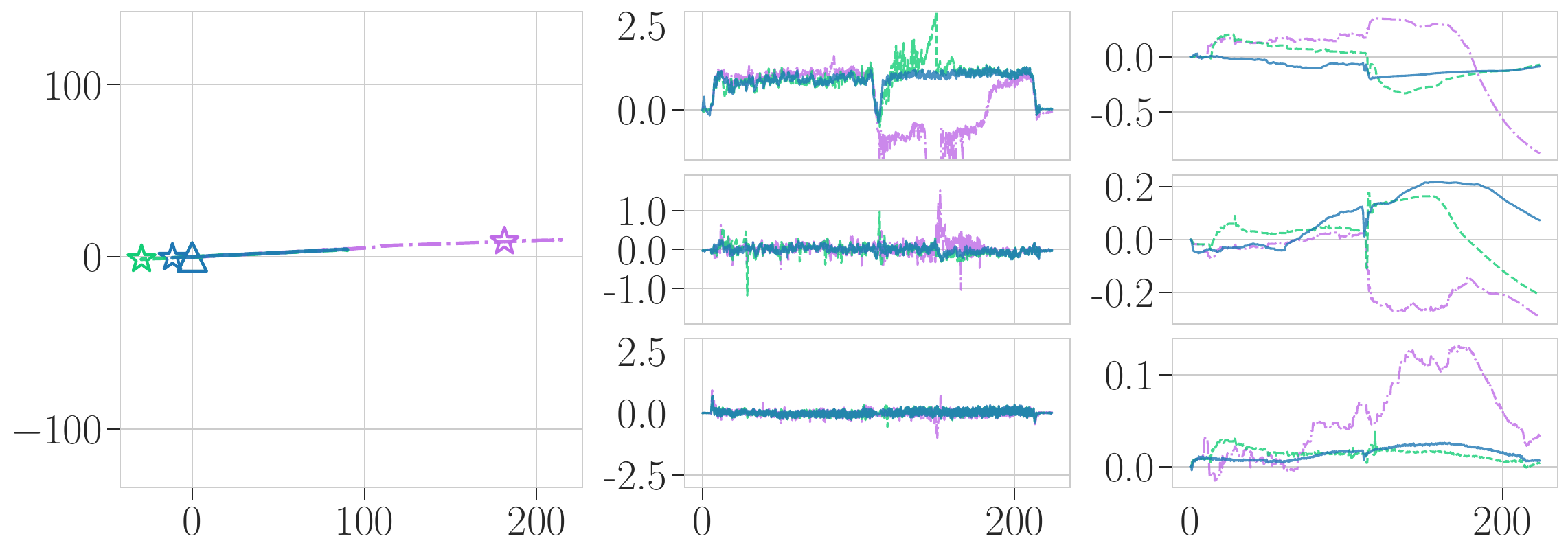}}

    \caption{Plots of the estimated trajectories, velocities, and acceleration biases for degenerate cases. The left column shows the trajectory (X–Y plane in meters), the center displays the body-frame velocities (velocity in meters per second over time in seconds), and the right column presents the estimated acceleration biases (m/s$^2$ over time). The symbols $\triangle$ and ☆ denote the start and end points, respectively, of each trajectory.}
    \label{fig:lio_output_state_update}
    \vspace{-5mm}
\end{figure*}

Some sequences reveal a limitation of the proposed method: The deep-learning-based velocity estimator fails to accurately predict velocities under complex motions, such as frequent rotations, which leads to degraded odometry accuracy.
\texttt{Bright\_Screen\_Wall} includes in-place yaw rotations, which negatively impact velocity prediction.
This effect is more pronounced in \texttt{HIT\_Graffiti\_Wall\_1-4}, where frequent handheld rotations led to substantial performance degradation.
Fig.~\ref{fig:lio_output_handheld}(b) illustrates that in \texttt{HIT\_Graffiti\_Wall\_1}, the velocities exhibited substantial fluctuations, which led to considerable degradation in estimation performance both within degenerate regions and across the entire sequence.
Despite exhibiting the same limitation in the \texttt{HIT\_Graffiti\_Wall\_2-3} sequences, the proposed method prevented odometry estimation failure, whereas the baseline algorithms suffered from odometry divergence.
Specifically, these sequences involved extended periods of facing a featureless wall, creating severe geometric degeneracy that caused other baselines to fail.
In contrast, our method maintained a relatively stable velocity estimation, thereby preventing odometry divergence.

Although our approach did not achieve the best performance across all evaluated datasets, two variants (w/ and w/o CO) showed the best results in 21 out of 27 sequences (excluding sequences where all algorithms diverged), effectively demonstrating their ability to mitigate degeneracy.
Furthermore, our w/ CO variant achieved lower average ATE and end-to-end errors than its w/o CO counterpart, demonstrating the improved performance of the proposed method.

\subsection{Performance Gain from ESKF-Based Integration} \label{subsec:43_eval_state_update}
Experiments showed that integrating deep-learning-based velocity into the ESKF enables consistent updates of all state components, yielding enhanced performance compared to other methods for updating the state.
Two additional approaches were implemented for comparison, as introduced in Section~\ref{sec:1_intro} and illustrated in Fig.~\ref{fig:state_update}.
In both approaches, the same deep-learning velocity output as ours is used, and only the state-update scheme is modified, so that the comparison isolates only the effect of the state-update scheme while keeping the neural network unchanged.
Liao~et al.~\cite{zongbo_tits} utilized the eigenvalues of degeneracy to compute a weighted mean between IMU motion prediction and LiDAR odometry.
For comparison, we implemented this approach by updating the state based on the velocity, where all components of ${}^B\mathbf{v}^\text{NN}_M$ along each axis were utilized.
The formulation of this method is
\begin{align}
	\overline{\mathbf{v}} &= \frac{{}^W\lambda_{\text{min}}}{{}^W\lambda_{\text{max}}} \widehat{\mathbf{v}} + (1-\frac{{}^W\lambda_{\text{min}}}{{}^W\lambda_{\text{max}}}) ({}^W \widehat{\mathbf{R}} {}^B\mathbf{v}^\text{NN}_M),\label{eq31} \\
    \overline{\mathbf{P}}_\text{vel} &= (\frac{{}^W\lambda_{\text{min}}}{{}^W\lambda_{\text{max}}})^2 \widehat{\mathbf{P}}_\text{vel} + (1-\frac{{}^W\lambda_{\text{min}}}{{}^W\lambda_{\text{max}}})^2 ({}^W\widehat{\mathbf{R}} \mathbf{\Sigma}^\text{NN}_{v, M} {}^W\widehat{\mathbf{R}}^\top)\label{eq32},
\end{align}
where $(\cdot)_\text{vel}$ refers to the velocity component of the state.
The original method is not based on a filter-based LIO framework and does not update the state covariance; thus, we constructed (\ref{eq32}) consistently with (\ref{eq31}).

Second, the output of ININ-LIO~\cite{inin_lio} is originally formulated as a relative position in the world frame, which differs from our body-frame velocity. Accordingly, the difference in the output is derived as follows, where the degenerate directions are replaced by ${}^B\mathbf{v}^\text{NN}_M$ and the non-degenerate directions by the LIO output:
\begin{align}
    {}^W\mathbf{V}_\mathcal{D} &= {}^W\widehat{\mathbf{R}}{}^{B}\mathbf{V}_\mathcal{D}{}^W\widehat{\mathbf{R}}^\top,\, {}^W\mathbf{V}_\mathcal{O} = (\mathbf{I}_3 - {}^W\widehat{\mathbf{R}}{}^{B}\mathbf{V}_\mathcal{D}{}^W\widehat{\mathbf{R}}^\top),\nonumber \\
    {}^W\overline{\mathbf{v}} &=  {}^W\mathbf{V}_\mathcal{O}\,{}^W\widehat{\mathbf{v}}+ {}^W\mathbf{V}_\mathcal{D} ({}^W \widehat{\mathbf{R}} {}^B\mathbf{v}^\text{NN}_M), \nonumber\\
    \overline{\mathbf{P}}_\text{vel} &=  {}^W\mathbf{V}_\mathcal{O} \widehat{\mathbf{P}}_\text{vel}  {}^W\mathbf{V}_\mathcal{O}^\top +  {}^W\mathbf{V}_\mathcal{D}({}^W\widehat{\mathbf{R}} \mathbf{\Sigma}^\text{NN}_{v, M} {}^W\widehat{\mathbf{R}}^\top)  {}^W\mathbf{V}_\mathcal{D}^\top, \nonumber
\end{align}
where ${}^W\mathbf{V}_\mathcal{O}$ denotes the projection matrix onto the non-degenerate space in the world frame.

\begin{table}[t!]
\vspace{-2mm}

\caption{Comparison Between Odometry Updates (ATE $\downarrow$ [m]).}
\centering
\small
\begin{tabular}{l|ccc}
\toprule[1pt]
\textbf{Sequence} & \makecell{Liao \\ et al.~\cite{zongbo_tits}} & \makecell{ININ-LIO \\\cite{inin_lio}} & \makecell{Ours} \\
\midrule
\texttt{ShieldTunnel\_1}  & 101.99 & 22.58  & \fcol \textbf{12.67} \\
\texttt{ShieldTunnel\_2}  & 162.67 & 105.94 & \fcol \textbf{21.60} \\
\texttt{ShieldTunnel\_3}  & 58.97  & 14.34  & \fcol \textbf{12.68} \\
\texttt{ShieldTunnel\_4}  & 79.98  & 27.76  & \fcol \textbf{21.84} \\
\texttt{ShieldTunnel\_5}  & 41.12  & $\times$ & \fcol \textbf{6.55}  \\
\texttt{ShieldTunnel\_6}  & 74.93  & 10.86  & \fcol \textbf{6.40}  \\
\midrule	
Average                      & 86.61  & --     & \fcol \textbf{13.55}  \\
\bottomrule[1pt]
\end{tabular}
\label{tab:ablation_state_update}
\par\vspace{1mm}\noindent
\raggedleft
{\footnotesize Each best result is highlighted as \ftext{\textbf{first}}.}
\vspace{-4mm}
\end{table}

As shown in Table~\ref{tab:ablation_state_update}, our ESKF-based strategy exhibited enhanced performance.
Updating the velocity across all axes~\cite{zongbo_tits} led to a marked performance decrease.
As previously discussed in Section~\ref{subsubsec4b3}, the neural velocity cannot be accurately predicted in all directions, which introduces errors.
Because the LIO system updates the state from prior estimates, incorrect updates result in accumulated estimation errors, leading to degraded performance.
In this sense, Fig.~\ref{fig:lio_output_state_update} illustrates that the velocity prediction performance gradually declined across all sequences.

Compared with ININ-LIO~\cite{inin_lio}, our approach offers three main advantages.
First, our approach updates all variables of the state $\mathbf{x}$ simultaneously, thereby maintaining consistency among them.
Fig.~\ref{fig:lio_output_state_update}(a) illustrates that ININ-LIO experiences substantial velocity oscillations, resulting in degraded odometry estimation accuracy, with a difference of 9.91\,m compared with our method, as noted in Table~\ref{tab:ablation_state_update}.
Substantial oscillations in the estimated velocity were also observed in Figs.~\ref{fig:lio_output_state_update}(b), (c), (d), and (f) compared to our method.
This result can be attributed to the biases and gravity, which should be jointly updated by the system but are not; for this reason, using them as inputs to the neural network may lead to unstable inference. In turn, because the ESKF jointly updates the biases with the other states, a reasonably accurate neural velocity drives the biases toward accurate estimates, forming a closed loop in which improved states yield improved network inputs at the subsequent step. However, this coupling has a limit. When the neural velocity becomes unreliable under prolonged degeneracy, the bias may likewise degrade and propagate back into the network input. The drift is therefore mitigated but not fully eliminated, which we regard as a remaining limitation.
Moreover, such oscillations can also be induced by the discrepancy between the LIO-estimated velocity and the neural velocity.
This discrepancy becomes pronounced when an inaccurate degeneracy detection makes the decision alternate frequently between the degenerate and well-conditioned states, which is analyzed in Section~\ref{appendix:switching_stability}.

Second, ININ-LIO relies entirely on ${}^B\mathbf{v}^\text{NN}_M$ in the degenerate directions, which may distort the trajectory, potentially introducing discontinuities.
In the trajectory of Fig.~\ref{fig:lio_output_state_update}(c), ININ-LIO slightly deviated from our method, likely because of its full reliance on the neural network during the pronounced velocity changes near 180\,s.
As shown in the center of Fig.~\ref{fig:lio_output_state_update}(c), although the difference between our estimated velocity and that of ININ-LIO is not pronounced, the subtle deviations suggest that fully trusting the neural velocity may be problematic.
Moreover, while the oscillation issue discussed previously can be mitigated through the proposed Kalman gain-based weighting to suppress noise effects, the velocity-only update in ININ-LIO does not provide this benefit.

Finally, our approach shows stability even in rotational motions that the neural network fails to predict accurately.
The \texttt{ShieldTunnel\_4} sequence involved substantial rotational motion, where ININ-LIO exhibited substantial performance degradation, as depicted in the left and center of Fig.~\ref{fig:lio_output_state_update}(d).
As shown in Fig.~\ref{fig:lio_output_state_update}(e), the estimated odometry diverged during a sudden rotation toward the wall at the start, which was not the case with our method.

To summarize, as shown in Table~\ref{tab:ablation_state_update}, the proposed method achieved the highest performance across all sequences, demonstrating that consistently and continuously updating all components of the state through the ESKF provides increased robustness in degenerate scenarios.

\subsection{Comparison with Vision-Integrated Methods Under Normal and Visually Degenerate Conditions}\label{subsec:44_comparison_lvio}
We conducted further experiments to support our third claim regarding the method's potential under simultaneously geometrically and photometrically sparse conditions.
We conducted experiments on our platform in the same geometry but under different illumination conditions, using a handheld setup (see Section~\ref{appendix:private_dataset_detail} for detailed information on the sequences).
The comparison was performed with the baseline PV-LIO~\cite{pv_lio} and the vision-integrated FAST-LIVO2~\cite{fast_livo2}.

\begin{figure}[t!]
    \centering
    \vspace{-1mm}
    \includegraphics[width=0.9\linewidth]{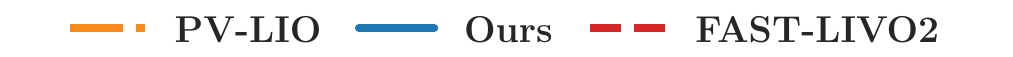} \\[1mm]
    \includegraphics[width=1.0\linewidth]{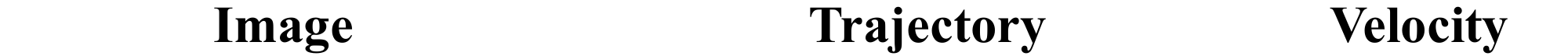} \\[1.0mm]
    \vspace{-3mm}
    \subfigure[\texttt{handheld\_light}]{\includegraphics[width=1.0\linewidth]{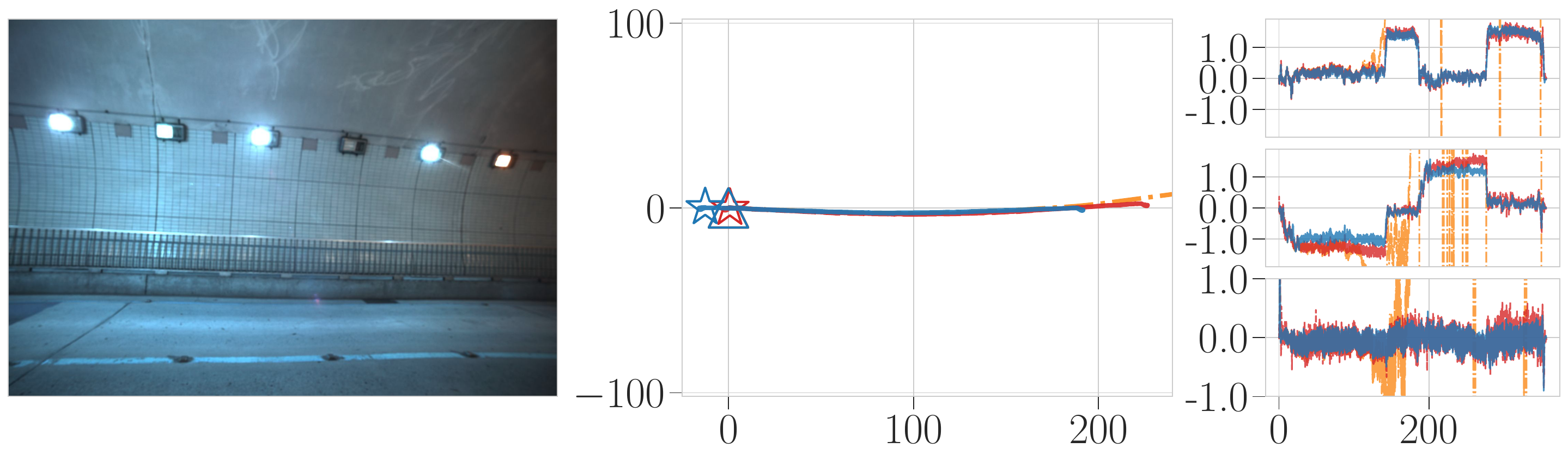}}
    \subfigure[\texttt{handheld\_dark}]{\includegraphics[width=1.0\linewidth]{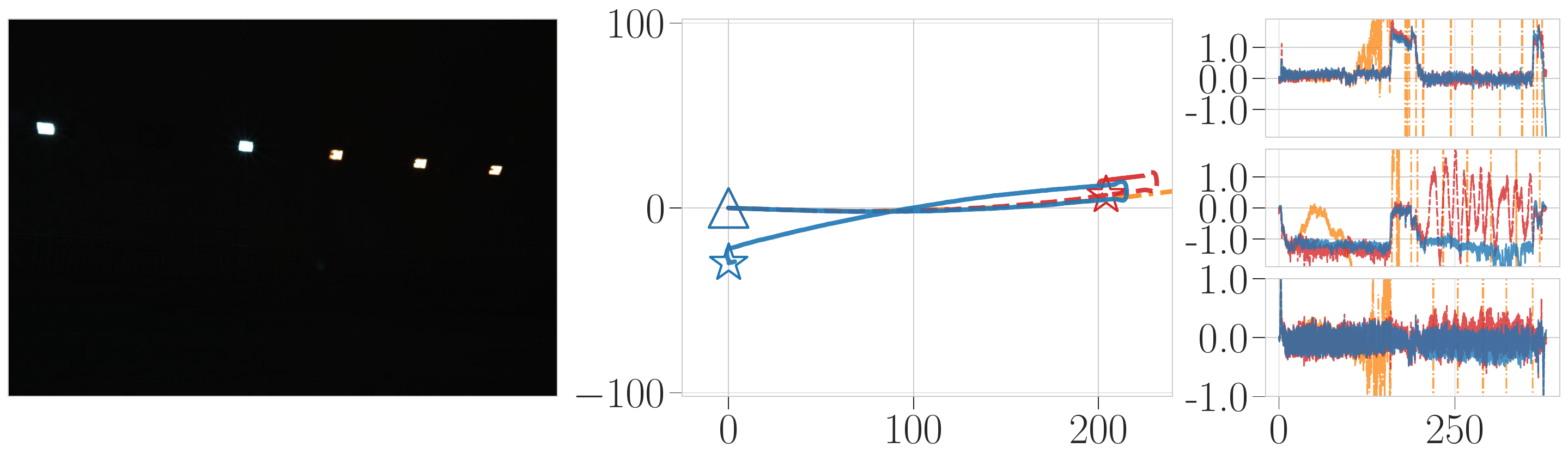}}
    
    \caption{Performance comparison under abundant and sparse visual features in geometrically featureless environments: Plots of estimated trajectory and velocity visualization on degenerate cases from the private dataset. The left column shows a single frame from the sequence, the middle column presents the estimated trajectory (X–Y plane in meters), and the right column displays the body-frame velocities (velocity in meters per second over time in seconds).
    }
    \label{fig:lio_output_private}
    \vspace{-4mm}
\end{figure}

As shown in Table~\ref{tab:evaluation_private_dataset}, PV-LIO diverged on both sequences, confirming that they were LiDAR-degenerate.
For \texttt{handheld\_light}, both our method and FAST-LIVO2 successfully estimated the poses without divergence, as shown in Table~\ref{tab:evaluation_private_dataset} and Fig.~\ref{fig:lio_output_private}(a).
FAST-LIVO2 achieved an end-to-end error of approximately 0.72\,m, whereas our method exhibited a larger error of 12.65\,m, indicating comparatively lower performance.
This result can be attributed to the fact that FAST-LIVO2 benefits from abundant visual features in the environment, allowing it to better perceive the surroundings.
In contrast, the slightly incorrect velocity estimates produced by our system accumulated over time, leading to the observed discrepancy in end-to-end error.

\begin{table}[t]
\centering
\small
\caption{Performance Evaluation on Private Datasets (End-to-End Error $\downarrow$ [m]).}
\label{tab:evaluation_private_dataset}
\begin{tabular}{l|>{\centering\arraybackslash}p{1.5cm}>{\centering\arraybackslash}p{1.5cm}>{\centering\arraybackslash}p{1.5cm}}
\toprule[1pt]
\multirow{1}{*}{\textbf{Sequence}} & \makecell{PV-LIO \\ \cite{pv_lio}} & \makecell{FAST-\\LIVO2~\cite{fast_livo2}} & \makecell{Ours} \\
\midrule
\texttt{handheld\_light}     & $\times$  & \fcol \textbf{0.72}    & 12.65 \\
\texttt{handheld\_dark}     & $\times$   & 205.35                 & \fcol \textbf{30.07} \\
\bottomrule[1pt]
\end{tabular}
\vspace{-4mm}
\end{table}

For \texttt{handheld\_dark}, as shown in the right side of Fig.~\ref{fig:lio_output_private}(b), FAST-LIVO2 failed to estimate the position near 200\,s, i.e., at the return point.
Despite the lack of visual features, FAST-LIVO2 estimated the position up to this point.
This result can be explained by the images shown in the left panel of Fig.~\ref{fig:lio_output_private}(b), where lighting served as a visual cue that facilitated traversal.
However, the failure of FAST-LIVO2 at the return point can be attributed to its use of LiDAR rays to determine the 3D positions of visual patches.
When the platform rotates to return, the LiDAR rays become approximately parallel to the visual patches, leading to incorrect 3D position estimates and ultimately causing the observed failure.
The issue arises in dark environments because few visual features are present.
In contrast, our system was able to prevent divergence.
Although the end-to-end error reached 30.07\,m, the trajectory shown in the middle panel of Fig.~\ref{fig:lio_output_private}(b) indicates that this error was caused mainly by accumulated rotational drift in the odometry.
Considering this result, the start and end points remained approximately close.
Hence, our results imply that the proposed method can mitigate degeneracy even when LiDAR and camera measurements are sparse.

\begin{table}[t]
\centering
\small
\caption{Effect of Input Representation Choices (AVE $\downarrow$ [m/s], ATE $\downarrow$ [m]).}
\label{tab:network_input}
\begin{tabular}{l|>{\centering\arraybackslash}p{0.7cm}>{\centering\arraybackslash}p{0.7cm}>{\centering\arraybackslash}p{0.7cm}>{\centering\arraybackslash}p{0.7cm}}
\toprule[1pt]
\multirow{2}{*}{\textbf{Sequence}} & \multicolumn{2}{c}{\makecell{Not \\ Compensated}} & \multicolumn{2}{c}{\makecell{${}^W\mathbf{g} , \ \mathbf{b}_{a},\  \mathbf{b}_{w}$ \\ Compensated}} \\
\cmidrule(lr){2-5}
& AVE & ATE & AVE & ATE \\
\midrule
\rowcolor{gray!10}\textbf{Car}                     &    &     &   &   \\
\texttt{vehicle\_campus\_1}      & 0.09 & 4.75                & \fcol \textbf{0.07}  & \fcol \textbf{4.54}  \\
\texttt{vehicle\_multilayer\_0}  & 0.06 & \fcol \textbf{3.81} & \fcol \textbf{0.04}  & 3.96  \\
\midrule
Average                             & 0.07 & 4.28                & \fcol \textbf{0.06}  & \fcol \textbf{4.25}  \\
\midrule
\rowcolor{gray!10}\textbf{Others}                  &    &     &   &   \\
\texttt{hku\_campus\_seq\_3}     & 0.15                & 3.55                & \fcol \textbf{0.14}  & \fcol \textbf{2.52}  \\
\texttt{hku\_park\_1}            & 0.20                & \fcol \textbf{7.26} & \fcol \textbf{0.19}  & 7.35  \\
\texttt{hkust\_campus\_3}        & 0.23                & 60.57               & \fcol \textbf{0.20}  & \fcol \textbf{45.54}  \\
\texttt{hku2}                    & 0.26                & 6.20                & \fcol \textbf{0.21}  & \fcol \textbf{5.95}  \\
\texttt{HKU\_Lecture\_Center\_2} & \fcol \textbf{0.24} & 3.58                & 0.26                 & \fcol \textbf{3.15}  \\
\texttt{HKU\_Main\_Building}     & 0.46                & 6.20                & \fcol \textbf{0.43}  & \fcol \textbf{4.33}  \\
\texttt{SYSU\_2}                 & 0.72                & 11.50               & \fcol \textbf{0.64}  & \fcol \textbf{11.49}  \\
\texttt{OffRoad\_5}              & 0.15                & \fcol \textbf{2.55} & \fcol \textbf{0.14}  & 4.31  \\
\texttt{OffRoad\_6}              & 0.13                & 5.51                & \fcol \textbf{0.12}  & \fcol \textbf{2.67}  \\
\texttt{OffRoad\_7}              & 0.14                & 0.83                & \fcol \textbf{0.12}  & \fcol \textbf{0.70}  \\
\texttt{TunnelingTunnel\_4}      & 0.14                & 5.59                & \fcol \textbf{0.11}  & \fcol \textbf{3.97}  \\
\texttt{TunnelingTunnel\_5}      & 0.11                & \fcol\textbf{2.67}  & \fcol \textbf{0.10}  & 3.36  \\
\midrule
Average                             & 0.24                & 9.67                & \fcol \textbf{0.22}  & \fcol \textbf{7.94}  \\
\bottomrule[1pt]
\end{tabular}
\vspace{-4mm}
\end{table}

\subsection{Impact of Utilizing LIO States on Network Generalization}\label{subsec:45_impact_deep_lio_output}
Although the prior study~\cite{inin_lio} introduced the exclusion of 
LIO-estimated biases from learning (while treating gravity as a constant), it did not examine how removing both the estimated biases and gravity affects the generalization performance of deep-learning models.
Thus, we analyzed the impact of utilizing LIO-estimated values on the input processing of the neural network, thereby supporting our fourth and final key claim.

As illustrated in Table~\ref{tab:network_input}, compensating for both biases and gravity consistently improves performance across most datasets.
Although the performance difference in AVE appears to be as small as 0.01\,m/s in most datasets, the odometry estimation errors became more substantial over time.
This result highlights the importance of considering the average ATE, which shows a 1.73\,m difference in the ``Others'' dataset.
In addition, as seen in \texttt{SYSU\_2}, which involved rotational motion (see Fig.~\ref{fig:network_output}), the 0.08\,m/s difference in AVE can be regarded as relatively substantial.
In contrast, for mostly linear-motion scenarios or low-DoF automotive platforms, the performance remained approximately identical.

Although compensating for gravity and biases has a relatively small impact in simple motions, leveraging LIO-estimated components in more complex motions enhances the system’s generalization capability, as discussed previously. Moreover, although this systematic bias cannot be entirely eliminated since the GT is not used, by explicitly excluding the LIO-estimated biases and gravity from the network inputs, we reduce the systematic bias with respect to the training dataset.

\subsection{Failure Cases and Performance Degradation Analysis}\label{subsec:46_failure_analysis}
For completeness, we discuss the cases in which the proposed method failed or showed degraded accuracy. First, the failure on \texttt{Urban\_Tunnel\_1} (Table~\ref{tab:lio_evaluation_ape}) is not attributable to our degeneracy handling: the scan-to-map registration of the LiDAR odometry failed from the very beginning of the sequence, so the pose estimate was already incorrect. Since our degeneracy detection is derived from registration process, its early failure also corrupted the detection, leaving the system at the baseline level; thus, the reliability of detection is bounded by that of the registration rather than by the learning-based fusion. In particular, dynamic objects can degrade both the registration and the resulting degeneracy detection; such cases can be alleviated by dynamic object removal, for which we refer the reader to~\cite{docking_bus, mos}.

Second, the learning-based velocity becomes less reliable under aggressive or in-place rotational motions (Sections~\ref{subsubsec4b3} and~\ref{subsec:43_eval_state_update}; e.g., \texttt{HIT\_Graffiti\_Wall} and \texttt{ShieldTunnel\_4}, Figs.~\ref{fig:lio_output_handheld} and~\ref{fig:lio_output_state_update}). Nevertheless, since the neural velocity is fused only along the degenerate directions, its influence remains bounded and its divergence is still prevented even under severe degeneracy; on the \texttt{FlatGround} sequences, our method avoids divergence but shows non-negligible discrepancies from the GT. The precise impact of rotational motion is difficult to quantify, as such motion is hard to learn given scarce training data and its high diversity; a detailed analysis of this behavior is provided in Section~\ref{appendix:rotation_reliability}.

Despite these limitations, ALIVE-LIO nevertheless consistently maintains robust and competitive performance across the broad range of degenerate scenarios that constitute our focus.

\section{Concluding Remarks} \label{sec:5_conclusion}
In this paper, we present ALIVE-LIO, a degeneracy-aware LiDAR–inertial odometry framework that integrates learning-based velocity estimation into a classical ESKF.
By addressing degeneracy, ALIVE-LIO enables reliable state correction along degenerate directions while preserving the probabilistic consistency and interdependencies among all state components.
Unlike existing heuristic approaches that loosely combine data-driven and model-based methods, ALIVE-LIO achieves tight integration by performing full-state updates within the ESKF, rather than correcting only the velocity.
Additionally, IMU inputs, which are compensated using bias and gravity estimates from the LIO system, are employed to enhance the generalization performance of learning-based velocity estimation. Extensive experiments across multiple platforms, including ground vehicles, aerial drones, and handheld devices, showed that ALIVE-LIO effectively mitigates LiDAR degeneracy in challenging real-world environments.
Furthermore, results on the private datasets suggest that, in scenarios where both geometric and photometric features are insufficient and visual measurements become unreliable, ALIVE-LIO can serve as a practical alternative, conditions under which conventional LVIO systems often struggle.

Despite these promising results, some limitations remain. ALIVE-LIO exhibited reduced robustness under aggressive rotational motion and may fail to trigger corrective updates when degeneracy is not successfully detected.
Additional practical challenges observed in real-world deployments include slow recovery from degeneracy, sensitivity to sensor placement and operating conditions, and performance degradation under severe mechanical vibration.
Future work will focus on improving the reliability of degeneracy detection, enhancing robustness under highly dynamic motions, and addressing the identified practical limitations to further extend the applicability of ALIVE-LIO in diverse and challenging environments.

\section*{Supplementary Information}
The Supplementary Material appended to this paper reports additional evaluations that support the results presented here.
These cover the deep learning-based velocity network and the system runtime (Sections~\ref{appendix:neural_network_evaluation} and~\ref{appendix:real_time_performance}), the scaling factor $\sigma_v$ and the pseudo-ground-truth labels (Sections~\ref{appendix:sigma_ablation} and~\ref{appendix:pvlio_reliability}), the reliability of the neural velocity under rotational motion, the temporal stability of the degeneracy decision, and the calibration of the predicted covariance (Sections~\ref{appendix:rotation_reliability} to~\ref{appendix:covariance_calibration}), and the details of our private datasets (Section~\ref{appendix:private_dataset_detail}).

\bibliographystyle{IEEEtran}
\bibliography{refs}

\begin{thebibliography}{10}
\providecommand{\url}[1]{#1}
\csname url@samestyle\endcsname
\providecommand{\newblock}{\relax}
\providecommand{\bibinfo}[2]{#2}
\providecommand{\BIBentrySTDinterwordspacing}{\spaceskip=0pt\relax}
\providecommand{\BIBentryALTinterwordstretchfactor}{4}
\providecommand{\BIBentryALTinterwordspacing}{\spaceskip=\fontdimen2\font plus
\BIBentryALTinterwordstretchfactor\fontdimen3\font minus
  \fontdimen4\font\relax}
\providecommand{\BIBforeignlanguage}[2]{{%
\expandafter\ifx\csname l@#1\endcsname\relax
\typeout{** WARNING: IEEEtran.bst: No hyphenation pattern has been}%
\typeout{** loaded for the language `#1'. Using the pattern for}%
\typeout{** the default language instead.}%
\else
\language=\csname l@#1\endcsname
\fi
#2}}
\providecommand{\BIBdecl}{\relax}
\BIBdecl

\bibitem{intro_lidar_feat}
C.~Cadena, L.~Carlone, H.~Carrillo, Y.~Latif, D.~Scaramuzza, J.~Neira, I.~Reid,
  and J.~J. Leonard, ``Past, present, and future of simultaneous localization
  and mapping: Toward the robust-perception age,'' \emph{IEEE Trans. Robot.},
  vol.~32, no.~6, pp. 1309--1332, 2016.

\bibitem{intro_survey_lo}
D.~Lee, M.~Jung, W.~Yang, and A.~Kim, ``{LiDAR} odometry survey: recent
  advancements and remaining challenges,'' \emph{Intell. Serv. Robot.},
  vol.~17, no.~2, pp. 95--118, Mar. 2024.

\bibitem{genzicp}
D.~Lee, H.~Lim, and S.~Han, ``{GenZ-ICP}: Generalizable and degeneracy-robust
  {LiDAR} odometry using an adaptive weighting,'' \emph{IEEE Robot. Autom.
  Lett.}, vol.~10, no.~1, pp. 152--159, 2025.

\bibitem{coin_lio}
P.~Pfreundschuh, H.~Oleynikova, C.~Cadena, R.~Siegwart, and O.~Andersson,
  ``{COIN-LIO}: Complementary intensity-augmented {LiDAR} inertial odometry,''
  in \emph{Proc. IEEE Int. Conf. Robot. Automat.}, 2024, pp. 1730--1737.

\bibitem{r3live}
J.~Lin and F.~Zhang, ``{R3LIVE}: A robust, real-time, {RGB}-colored,
  {LiDAR}-inertial-visual tightly-coupled state estimation and mapping
  package,'' in \emph{Proc. IEEE Int. Conf. Robot. Automat.}, 2022, pp.
  10\,672--10\,678.

\bibitem{fast_livo1}
C.~Zheng, Q.~Zhu, W.~Xu, X.~Liu, Q.~Guo, and F.~Zhang, ``Fast-{LIVO}: Fast and
  tightly-coupled sparse-direct {LiDAR}-inertial-visual odometry,'' in
  \emph{Proc. IEEE/RSJ Int. Conf. Intell. Robot. Syst.}, 2022, pp. 4003--4009.

\bibitem{fast_livo2}
C.~Zheng, W.~Xu, Z.~Zou, T.~Hua, C.~Yuan, D.~He, B.~Zhou, Z.~Liu, J.~Lin,
  F.~Zhu, Y.~Ren, R.~Wang, F.~Meng, and F.~Zhang, ``Fast-{LIVO2}: Fast, direct
  {LiDAR}–inertial–visual odometry,'' \emph{IEEE Trans. Robot.}, vol.~41,
  pp. 326--346, 2025.

\bibitem{switch_slam}
J.~Lee, R.~Komatsu, M.~Shinozaki, T.~Kitajima, H.~Asama, Q.~An, and
  A.~Yamashita, ``Switch-{SLAM}: Switching-based {LiDAR}-inertial-visual {SLAM}
  for degenerate environments,'' \emph{IEEE Robot. Autom. Lett.}, vol.~9,
  no.~8, pp. 7270--7277, 2024.

\bibitem{hanbiao_tits_lvio}
H.~Xiao, Z.~Hu, C.~Lv, J.~Meng, J.~Zhang, and J.~You, ``Progressive multi-modal
  semantic segmentation guided {SLAM} using tightly-coupled
  {LiDAR}-visual-inertial odometry,'' \emph{IEEE Trans. Intell. Transp. Syst.},
  vol.~26, no.~2, pp. 1645--1656, 2025.

\bibitem{taku_wheel_lio}
T.~Okawara, K.~Koide, S.~Oishi, M.~Yokozuka, A.~Banno, K.~Uno, and K.~Yoshida,
  ``Tightly-coupled {LiDAR}-{IMU}-wheel odometry with an online neural
  kinematic model learning via factor graph optimization,'' \emph{Robot. Auton.
  Syst.}, vol. 187, p. 104929, 2025.

\bibitem{x_icp}
T.~Tuna, J.~Nubert, Y.~Nava, S.~Khattak, and M.~Hutter, ``{X-ICP}:
  Localizability-aware {LiDAR} registration for robust localization in extreme
  environments,'' \emph{IEEE Trans. Robot.}, vol.~40, pp. 452--471, 2024.

\bibitem{tucan_field_robotics}
T.~Tuna, J.~Nubert, P.~Pfreundschuh, C.~Cadena, S.~Khattak, and M.~Hutter,
  ``Informed, constrained, aligned: A field analysis on degeneracy-aware point
  cloud registration in the wild,'' \emph{IEEE Trans. Field Robot.}, vol.~2,
  pp. 485--515, 2025.

\bibitem{relead}
Z.~Chen, H.~Chen, Y.~Qi, S.~Zhong, D.~Feng, J.~Wu, W.~Wen, and M.~Liu,
  ``{RELEAD}: Resilient localization with enhanced {LiDAR} odometry in adverse
  environments,'' in \emph{Proc. IEEE Int. Conf. Robot. Automat.}, 2024, pp.
  3999--4005.

\bibitem{zongbo_tits}
Z.~Liao, X.~Zhang, T.~Zhang, Z.~Li, Z.~Zheng, Z.~Wen, and Y.~Li, ``A real-time
  degeneracy sensing and compensation method for enhanced {LiDAR} {SLAM},''
  \emph{IEEE Trans. Intell. Transp. Syst.}, vol.~26, no.~3, pp. 4202--4213,
  2025.

\bibitem{inin_lio}
Y.~Gao, L.~Zhao, and B.~Zhang, ``{ININ-LIO}: Informer-based neural inertial
  network-aided {LiDAR}–inertial odometry,'' \emph{IEEE Trans. Instrum.
  Meas.}, vol.~74, pp. 1--10, 2025.

\bibitem{lp_icp}
H.~Yue, Q.~Xu, F.~Chen, J.~Pan, and W.~Chen, ``{LP-ICP}: General
  localizability-aware point cloud registration for robust localization in
  extreme unstructured environments,'' 2025, arXiv:2501.02580.

\bibitem{airio}
Y.~Qiu, C.~Xu, Y.~Chen, S.~Zhao, J.~Geng, and S.~Scherer, ``{AirIO}: Learning
  inertial odometry with enhanced {IMU} feature observability,'' \emph{IEEE
  Robot. Autom. Lett.}, vol.~10, no.~9, pp. 9368--9375, 2025.

\bibitem{tartan_imu}
S.~Zhao, S.~Zhou, R.~Blanchard, Y.~Qiu, W.~Wang, and S.~Scherer, ``{Tartan
  IMU}: A light foundation model for inertial positioning in robotics,'' in
  \emph{Proc. IEEE/CVF Conf. Comput. Vis. and Pattern Recognit.}, 2025, pp.
  22\,520--22\,529.

\bibitem{pv_lio}
C.~Yuan, W.~Xu, X.~Liu, X.~Hong, and F.~Zhang, ``Efficient and probabilistic
  adaptive voxel mapping for accurate online {LiDAR} odometry,'' \emph{IEEE
  Robot. Autom. Lett.}, vol.~7, no.~3, pp. 8518--8525, 2022.

\bibitem{onnx}
\BIBentryALTinterwordspacing
O.~R. developers, ``{ONNX} runtime,'' 2021, {Accessed: Aug. 30, 2025}.
  [Online]. Available: \url{https://onnxruntime.ai/}
\BIBentrySTDinterwordspacing

\bibitem{ptpl_icp}
S.~Rusinkiewicz and M.~Levoy, ``Efficient variants of the {ICP} algorithm,'' in
  \emph{Proc. IEEE Int. Conf. 3D Digital Imaging and Modeling}, 2001, pp.
  145--152.

\bibitem{zhang_icra}
J.~Zhang, M.~Kaess, and S.~Singh, ``On degeneracy of optimization-based state
  estimation problems,'' in \emph{Proc. IEEE Int. Conf. Robot. Automat.}, 2016,
  pp. 809--816.

\bibitem{lion}
A.~Tagliabue, J.~Tordesillas, X.~Cai, A.~Santamaria-Navarro, J.~P. How,
  L.~Carlone, and A.-a. Agha-mohammadi, ``{LION}: {LiDAR}-inertial
  observability-aware navigator for vision-denied environments,'' in
  \emph{Proc. Int. Symp. Exp. Robot.}, 2021, pp. 380--390.

\bibitem{mm_lins}
Y.~Ma, J.~Xu, S.~Yuan, T.~Zhi, W.~Yu, J.~Zhou, and L.~Xie, ``{MM-LINS}: A
  multi-map {LiDAR}-inertial system for over-degenerate environments,''
  \emph{IEEE Trans. Intell. Veh.}, vol.~10, no.~1, pp. 472--482, 2025.

\bibitem{io_survey}
C.~Chen and X.~Pan, ``Deep learning for inertial positioning: A survey,''
  \emph{IEEE Trans. Intell. Transp. Syst.}, vol.~25, no.~9, pp.
  10\,506--10\,523, 2024.

\bibitem{ori_net}
M.~A. Esfahani, H.~Wang, K.~Wu, and S.~Yuan, ``{OriNet}: Robust {3-D}
  orientation estimation with a single particular {IMU},'' \emph{IEEE Robot.
  Autom. Lett.}, vol.~5, no.~2, pp. 399--406, 2020.

\bibitem{airimu}
Y.~Qiu, C.~Wang, C.~Xu, Y.~Chen, X.~Zhou, Y.~Xia, and S.~Scherer, ``{AirIMU}:
  Learning uncertainty propagation for inertial odometry,'' 2024,
  arXiv:2310.04874.

\bibitem{tlio}
W.~Liu, D.~Caruso, E.~Ilg, J.~Dong, A.~I. Mourikis, K.~Daniilidis, V.~Kumar,
  and J.~Engel, ``{TLIO}: Tight learned inertial odometry,'' \emph{IEEE Robot.
  Autom. Lett.}, vol.~5, no.~4, pp. 5653--5660, 2020.

\bibitem{ridi}
H.~Yan, Q.~Shan, and Y.~Furukawa, ``{RIDI}: Robust {IMU} double integration,''
  in \emph{Proc. European Conf. Comput. Vis.}, 2018, pp. 641--656.

\bibitem{dive}
A.~Bajwa, C.~C. Cossette, M.~A. Shalaby, and J.~R. Forbes, ``{DIVE}: Deep
  inertial-only velocity aided estimation for quadrotors,'' \emph{IEEE Robot.
  Autom. Lett.}, vol.~9, no.~4, pp. 3728--3734, 2024.

\bibitem{jin_tits}
Y.~Jin, W.-A. Zhang, H.~Sun, and L.~Yu, ``Learning-aided inertial odometry with
  nonlinear state estimator on manifold,'' \emph{IEEE Trans. Intell. Transp.
  Syst.}, vol.~24, no.~9, pp. 9792--9803, 2023.

\bibitem{fast_lio1}
W.~Xu and F.~Zhang, ``Fast-{LIO}: A fast, robust {LiDAR}-inertial odometry
  package by tightly-coupled iterated {Kalman} filter,'' \emph{IEEE Robot.
  Autom. Lett.}, vol.~6, no.~2, pp. 3317--3324, 2021.

\bibitem{fast_lio2}
W.~Xu, Y.~Cai, D.~He, J.~Lin, and F.~Zhang, ``Fast-{LIO2}: Fast direct
  lidar-inertial odometry,'' \emph{IEEE Trans. Robot.}, vol.~38, no.~4, pp.
  2053--2073, 2022.

\bibitem{geode}
Z.~Chen, Y.~Qi, D.~Feng, X.~Zhuang, H.~Chen, X.~Hu, J.~Wu, K.~Peng, and P.~Lu,
  ``Heterogeneous {LiDAR} dataset for benchmarking robust localization in
  diverse degenerate scenarios,'' \emph{The Int. J. Robot. Res.}, 2025.

\bibitem{marios_kalman_eccomas}
M.~Panias, L.~Caglio, S.~Glavind, A.~Sadeqi, M.~Faber, H.~Stang, and
  E.~Katsanos, ``Tuning of {Kalman} filter noise parameters for uncertainty
  quantification in input-state estimation of {MDOF} systems,'' in \emph{Proc.
  ECCOMAS Thematic Conf. Uncertainty Quan. Comput. Sci. Eng.}\hskip 1em plus
  0.5em minus 0.4em\relax National Technical University of Athens, 2023.

\bibitem{christian_tro}
C.~Forster, L.~Carlone, F.~Dellaert, and D.~Scaramuzza, ``On-manifold
  preintegration for real-time visual--inertial odometry,'' \emph{IEEE Trans.
  Robot.}, vol.~33, no.~1, pp. 1--21, 2017.

\bibitem{imo}
G.~Cioffi, L.~Bauersfeld, E.~Kaufmann, and D.~Scaramuzza, ``Learned inertial
  odometry for autonomous drone racing,'' \emph{IEEE Robot. Autom. Lett.},
  vol.~8, no.~5, pp. 2684--2691, 2023.

\bibitem{urban_nav_hk}
L.-T. Hsu, F.~Huang, H.-F. Ng, G.~Zhang, Y.~Zhong, X.~Bai, and W.~Wen, ``{Hong
  Kong UrbanNav}: An open-source multisensory dataset for benchmarking urban
  navigation algorithms,'' \emph{NAVIGATION: J. Inst. Navig.}, vol.~70, no.~4,
  2023.

\bibitem{urban_nav_hsu}
L.-T. Hsu, N.~Kubo, W.~Wen, W.~Chen, Z.~Liu, T.~Suzuki, and J.~Meguro,
  ``{UrbanNav}: An open-sourced multisensory dataset for benchmarking
  positioning algorithms designed for urban areas,'' in \emph{Proc. Int. Tech.
  Meet. Satell. Division of the Inst. Navig.}, St. Louis, Missouri, Sep. 2021,
  pp. 226--256.

\bibitem{genzlio}
D.~Lee, H.~Lim, S.~Kim, S.~Rho, C.~Lee, S.~Park, J.~Hong, E.~Choi, H.~Jo, and
  S.~Han, ``{GenZ-LIO}: Generalizable {LiDAR}-inertial odometry beyond
  confined--open boundaries,'' 2026.

\bibitem{dlio}
K.~Chen, R.~Nemiroff, and B.~T. Lopez, ``Direct {LiDAR}-inertial odometry:
  Lightweight {LIO} with continuous-time motion correction,'' in \emph{Proc.
  IEEE Int. Conf. Robot. Automat.}, 2023, pp. 3983--3989.

\bibitem{evo}
\BIBentryALTinterwordspacing
M.~Grupp, ``evo: Python package for the evaluation of odometry and {SLAM}.''
  {Accessed: Oct. 12, 2025}. [Online]. Available:
  \url{https://github.com/MichaelGrupp/evo}
\BIBentrySTDinterwordspacing

\bibitem{park2025dataset}
S.~Park, ``{CoCEL} handheld dataset,''
  \url{https://github.com/SanghyunPark01/CoCEL_Handheld-Dataset}, 2025.

\bibitem{docking_bus}
Y.~He, Y.~Gao, Y.~Liu, and X.~Qu, ``{LiDAR-IMU SLAM} framework in autonomous
  modular bus docking systems,'' \emph{Green Energy and Intelligent
  Transportation}, vol.~4, no.~6, p. 100343, 2025.

\bibitem{mos}
H.~Assemlali, S.~Bouhsissin, and N.~Sael, ``Deep learning-driven {CNN} model
  for detection and classification of dynamic obstacles,'' \emph{Green Energy
  and Intelligent Transportation}, vol.~5, no.~3, p. 100334, 2026.

\bibitem{fusion_portable_v2}
H.~Wei, J.~Jiao, X.~Hu, J.~Yu, X.~Xie, J.~Wu, Y.~Zhu, Y.~Liu, L.~Wang, and
  M.~Liu, ``{FusionPortableV2}: A unified multi-sensor dataset for generalized
  slam across diverse platforms and scalable environments,'' \emph{The Int. J.
  Robot. Res.}, vol.~44, no.~7, pp. 1093--1116, 2025.

\end{thebibliography}

\setcounter{section}{0}
\setcounter{subsection}{0}
\setcounter{table}{0}
\setcounter{figure}{0}
\setcounter{equation}{0}
\renewcommand{\thetable}{S\arabic{table}}
\renewcommand{\thefigure}{S\arabic{figure}}
\renewcommand{\thesubsection}{S\Alph{subsection}}
\makeatletter
\def\thesubsectiondis{S\Alph{subsection}.}
\makeatother
\renewcommand{\theHtable}{S\arabic{table}}
\renewcommand{\theHfigure}{S\arabic{figure}}
\renewcommand{\theHsubsection}{S\Alph{subsection}}
\renewcommand{\theHequation}{S\arabic{equation}}

\twocolumn[{%
\section*{Supplementary Material}
\begin{center}
    \centering
    \makeatletter\def\@captype{figure}\makeatother
    \includegraphics[width=0.4\linewidth]{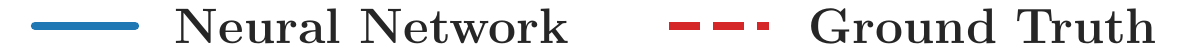}\\
    \includegraphics[width=0.244\linewidth]{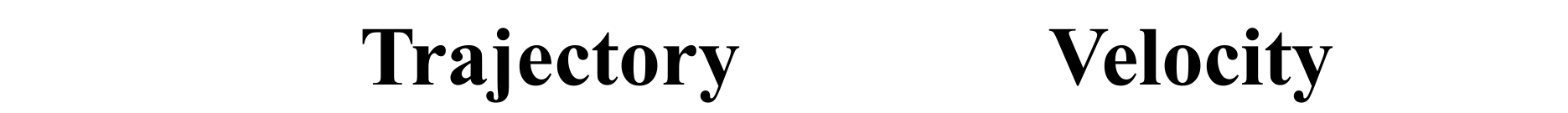}
    \includegraphics[width=0.244\linewidth]{Figures/traj_title_handheld3.pdf}
    \includegraphics[width=0.244\linewidth]{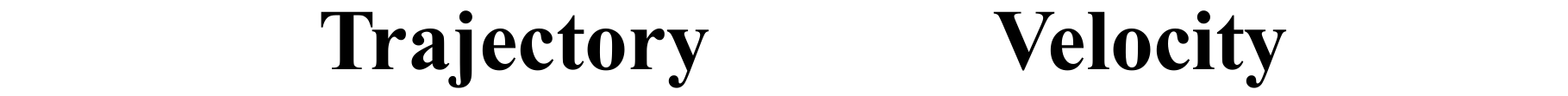}
    \includegraphics[width=0.244\linewidth]{Figures/traj_title_handheld3.pdf} \\
    \subfigure[\texttt{SYSU\_2}]{\includegraphics[width=0.245\linewidth]{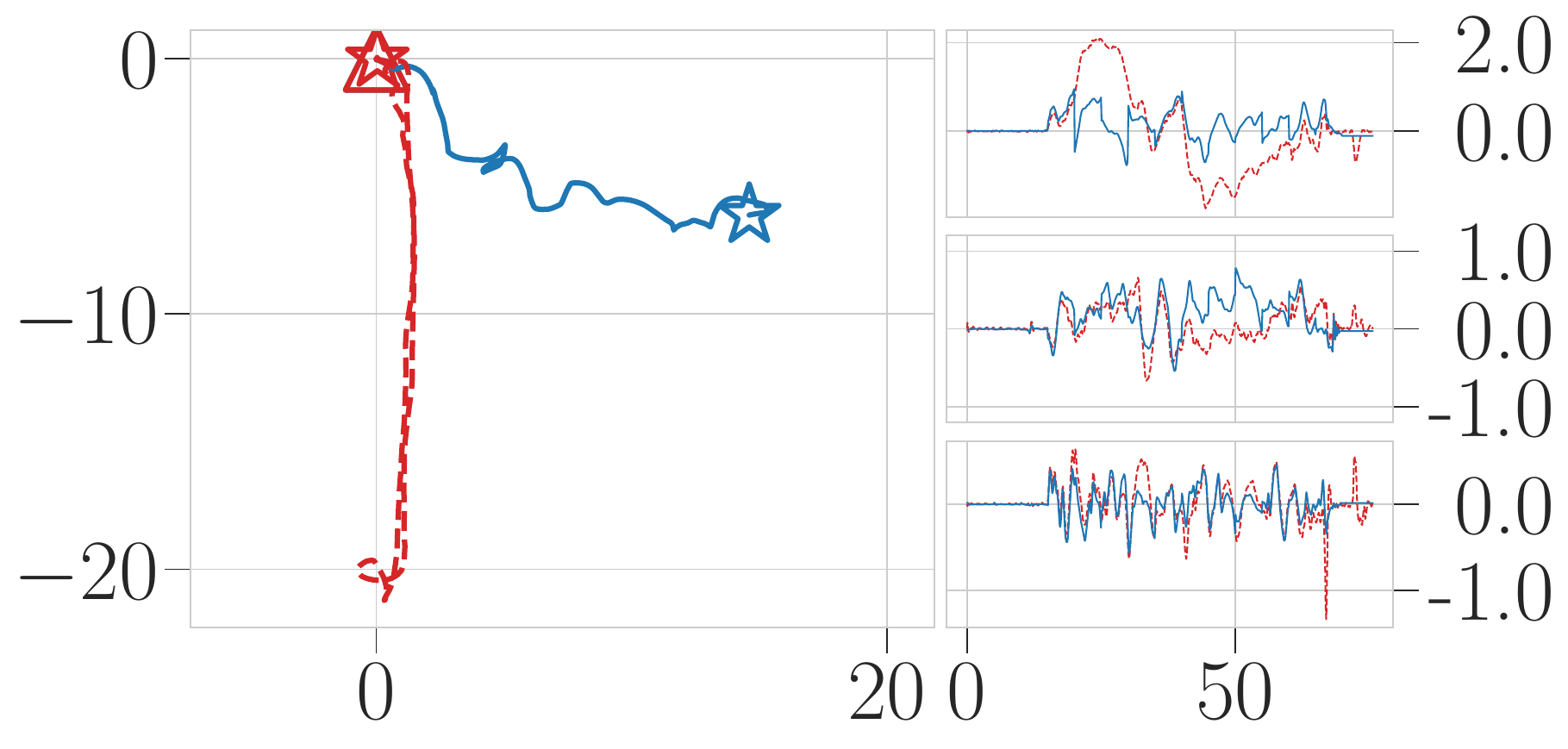}}
    \subfigure[\texttt{hku\_campus\_seq\_3}]{\includegraphics[width=0.246\linewidth]{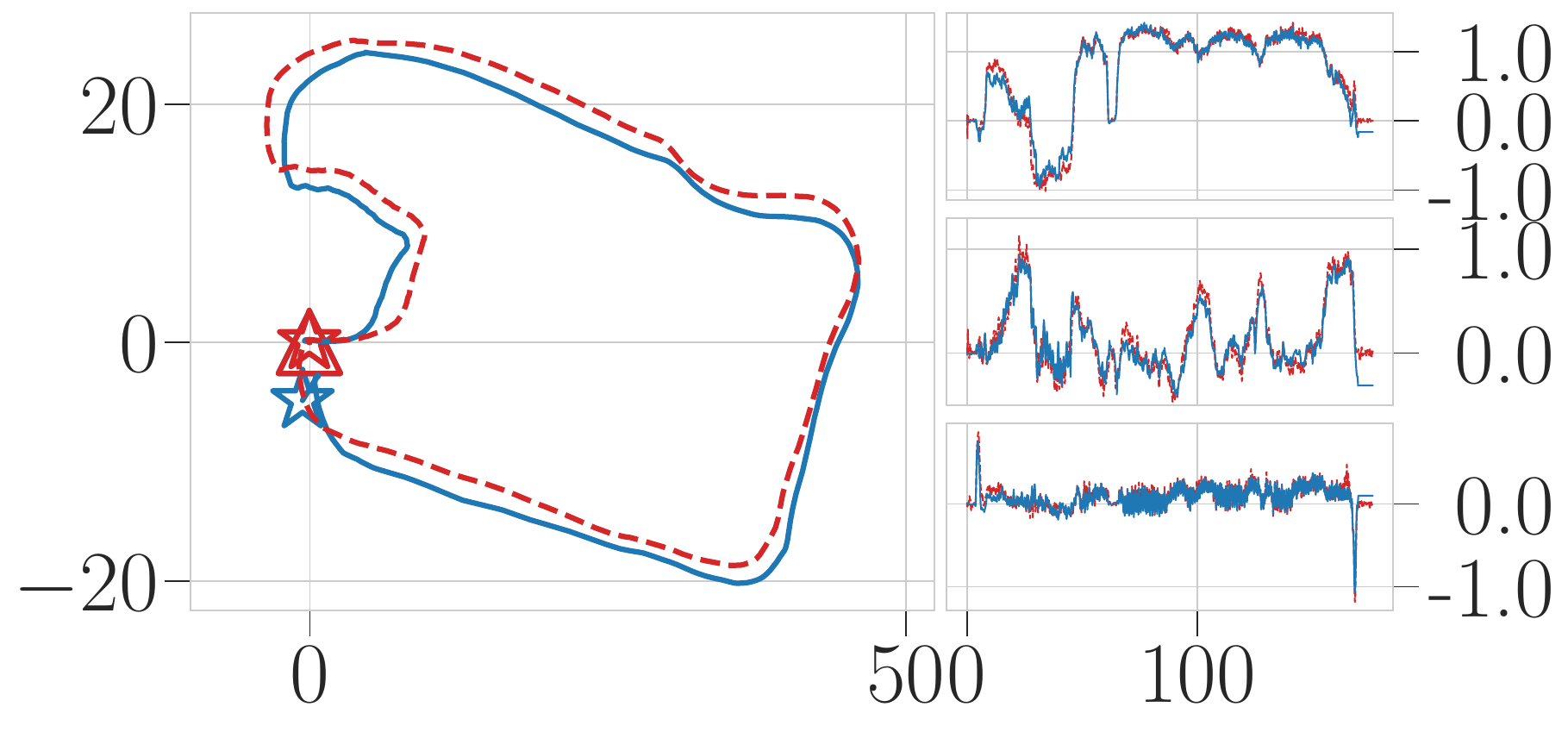}}
    \subfigure[\texttt{OffRoad\_5}]{\includegraphics[width=0.239\linewidth]{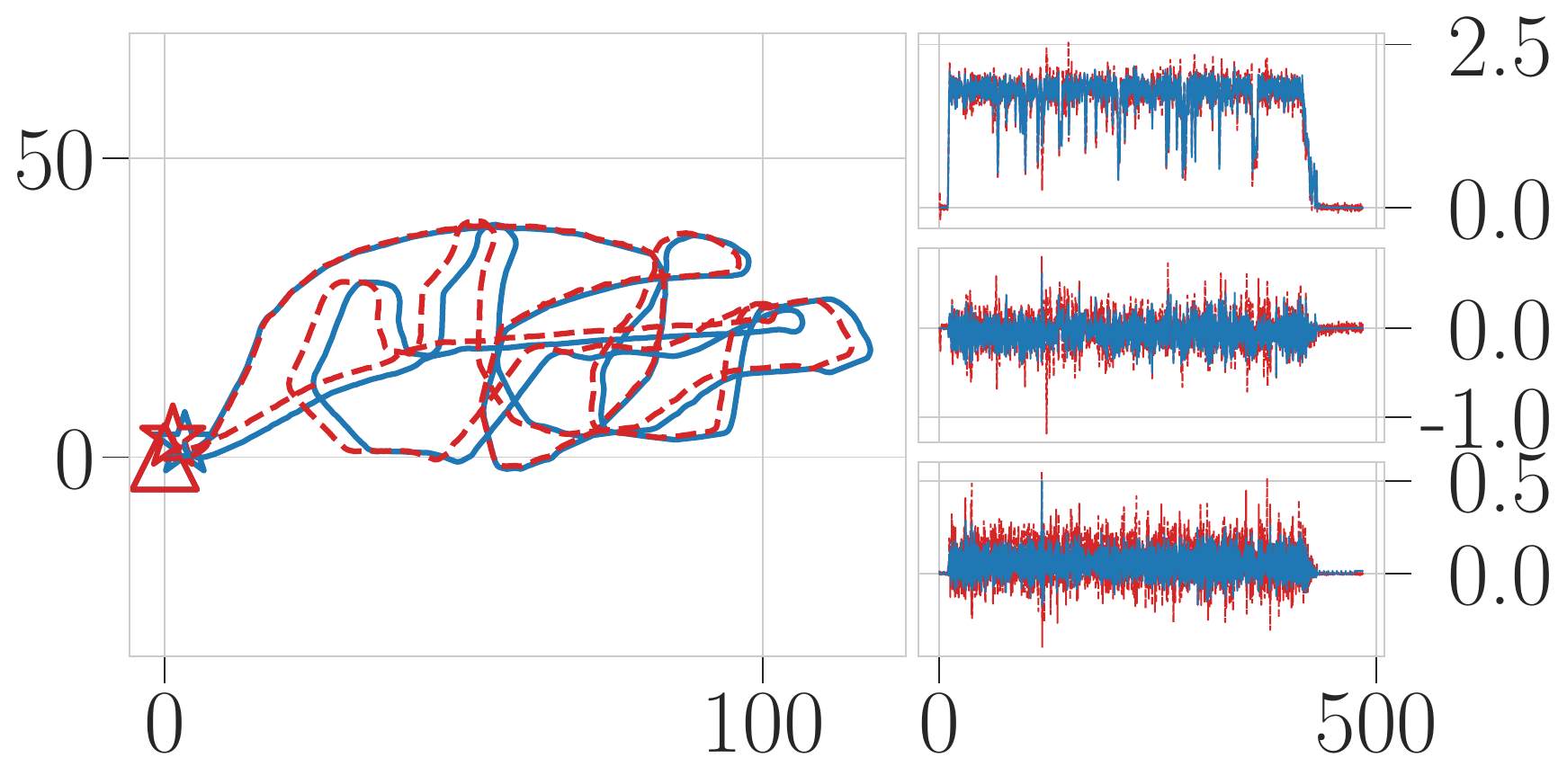}}
    \subfigure[\texttt{vehicle\_campus\_1}]{\includegraphics[width=0.252\linewidth]{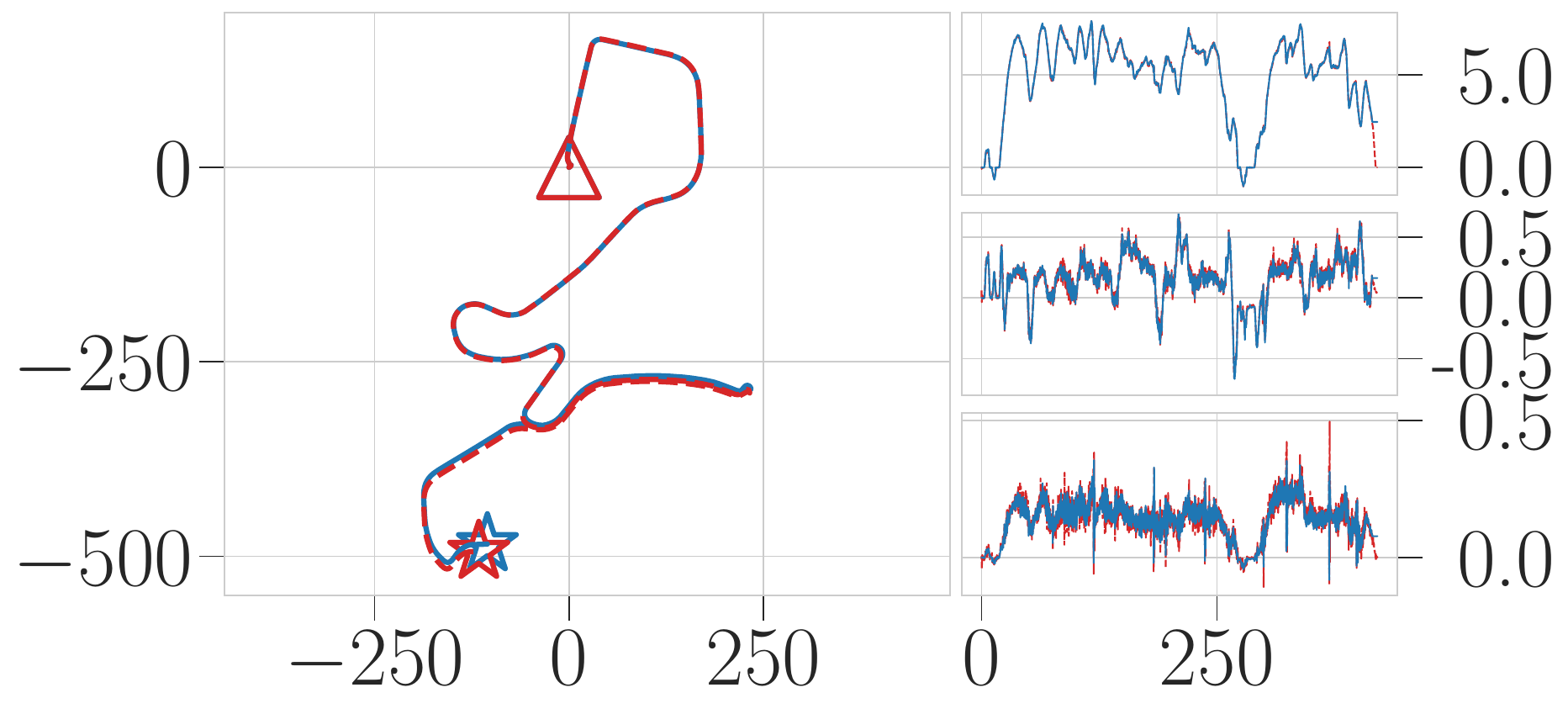}}
    \subfigure[\texttt{HKU\_Main\_Building}]{\includegraphics[width=0.245\linewidth]{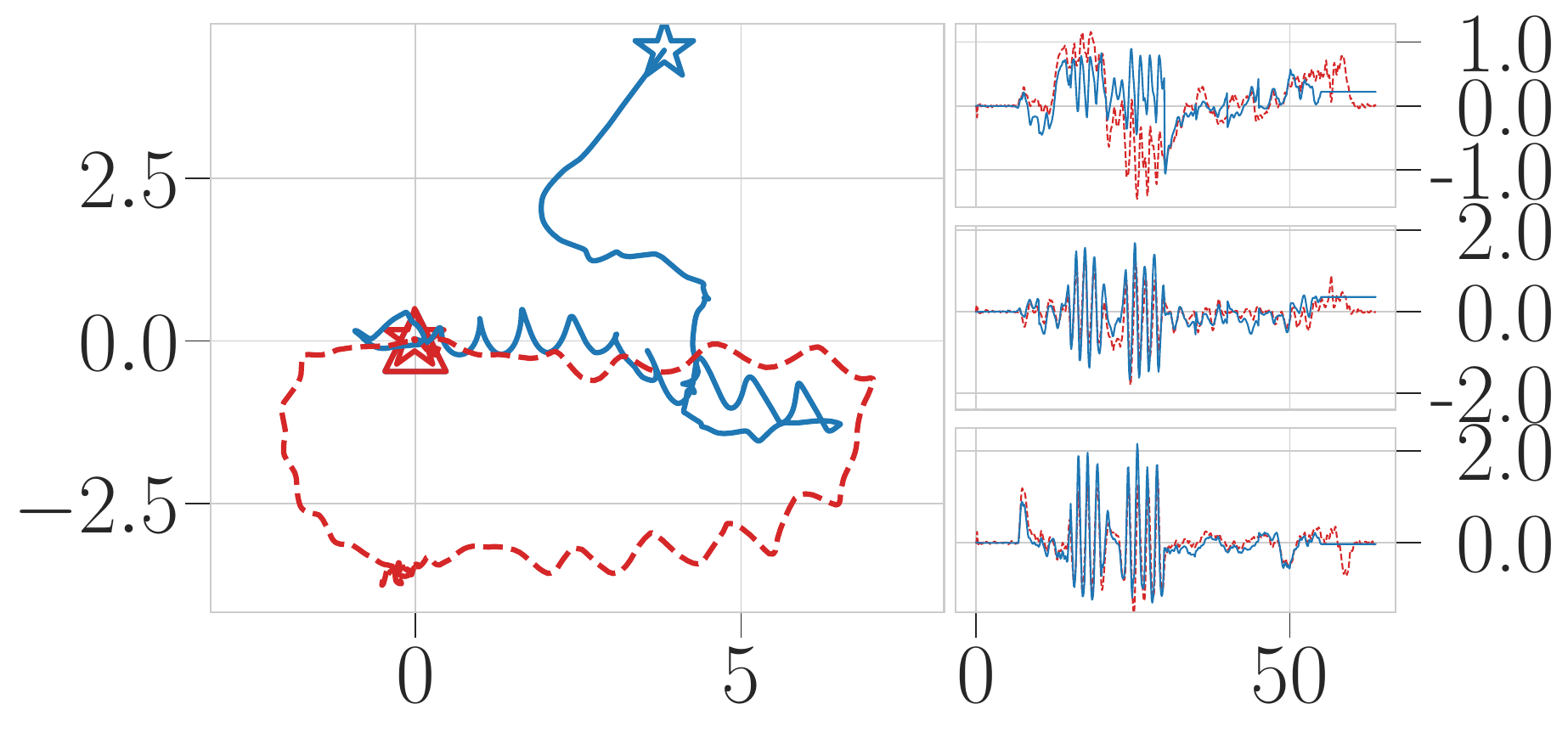}}
    \subfigure[\texttt{hkust\_campus\_3}]{\includegraphics[width=0.245\linewidth]{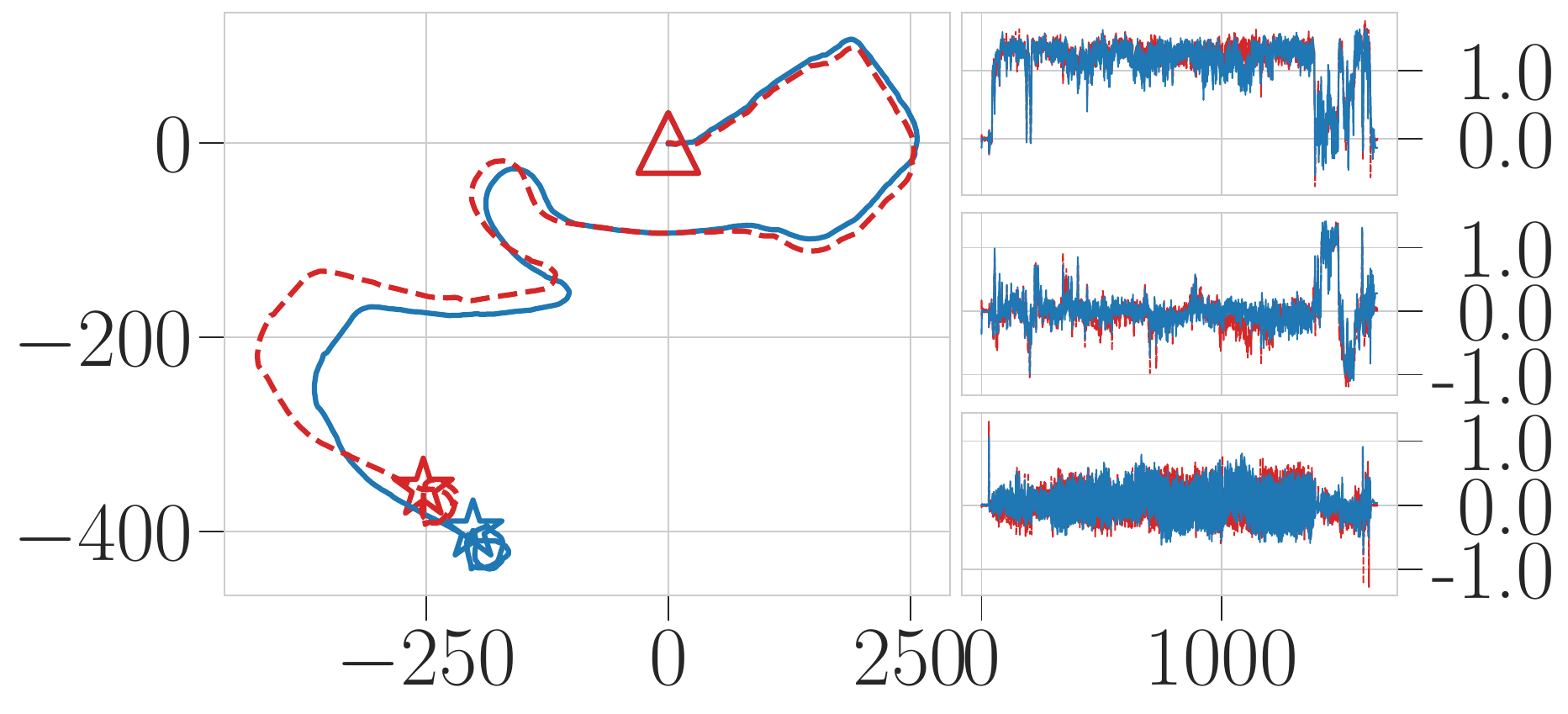}}
    \subfigure[\texttt{OffRoad\_7}]{\includegraphics[width=0.245\linewidth]{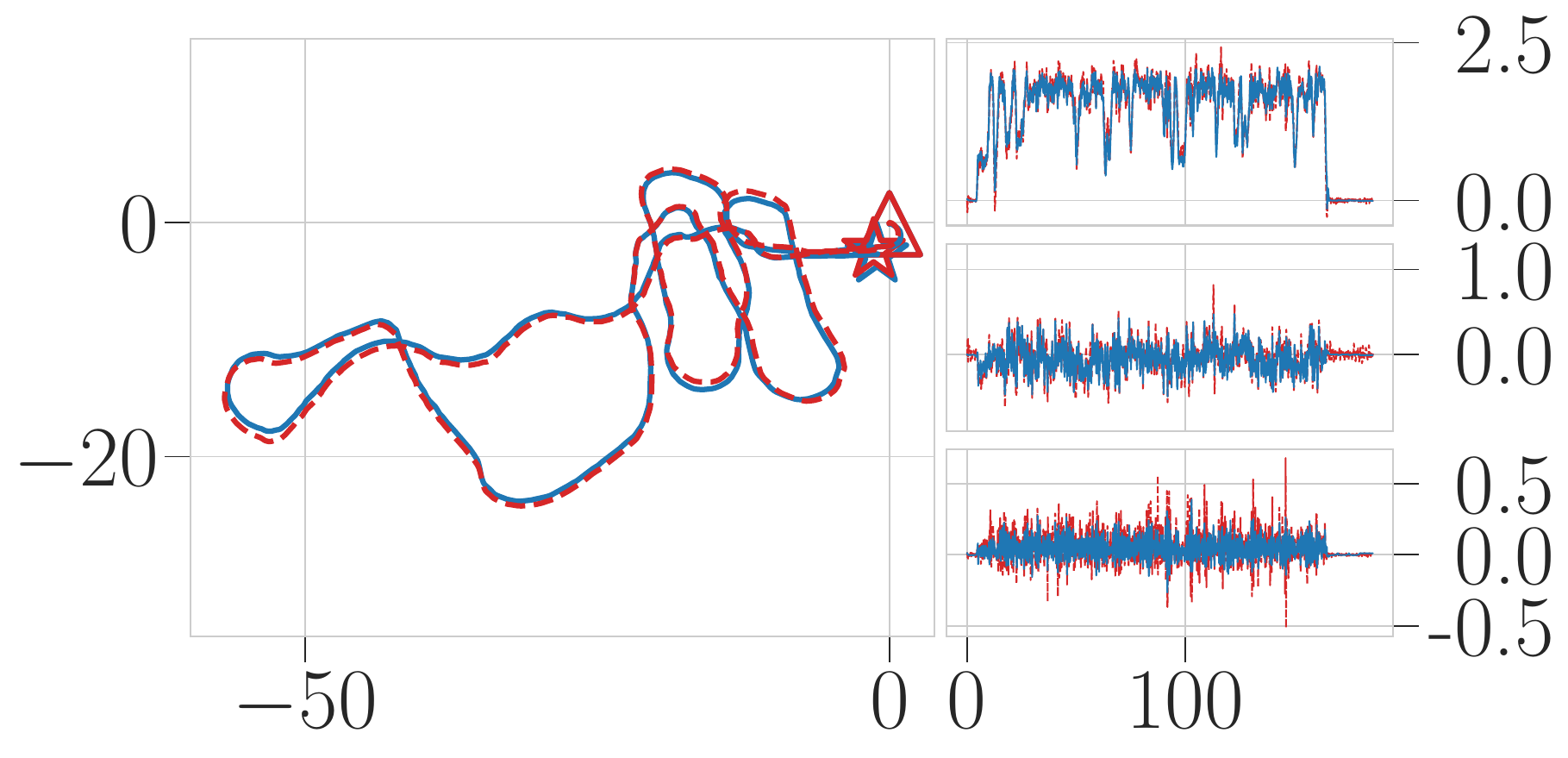}}
    \subfigure[\texttt{vehicle\_multilayer\_0}]{\includegraphics[width=0.245\linewidth]{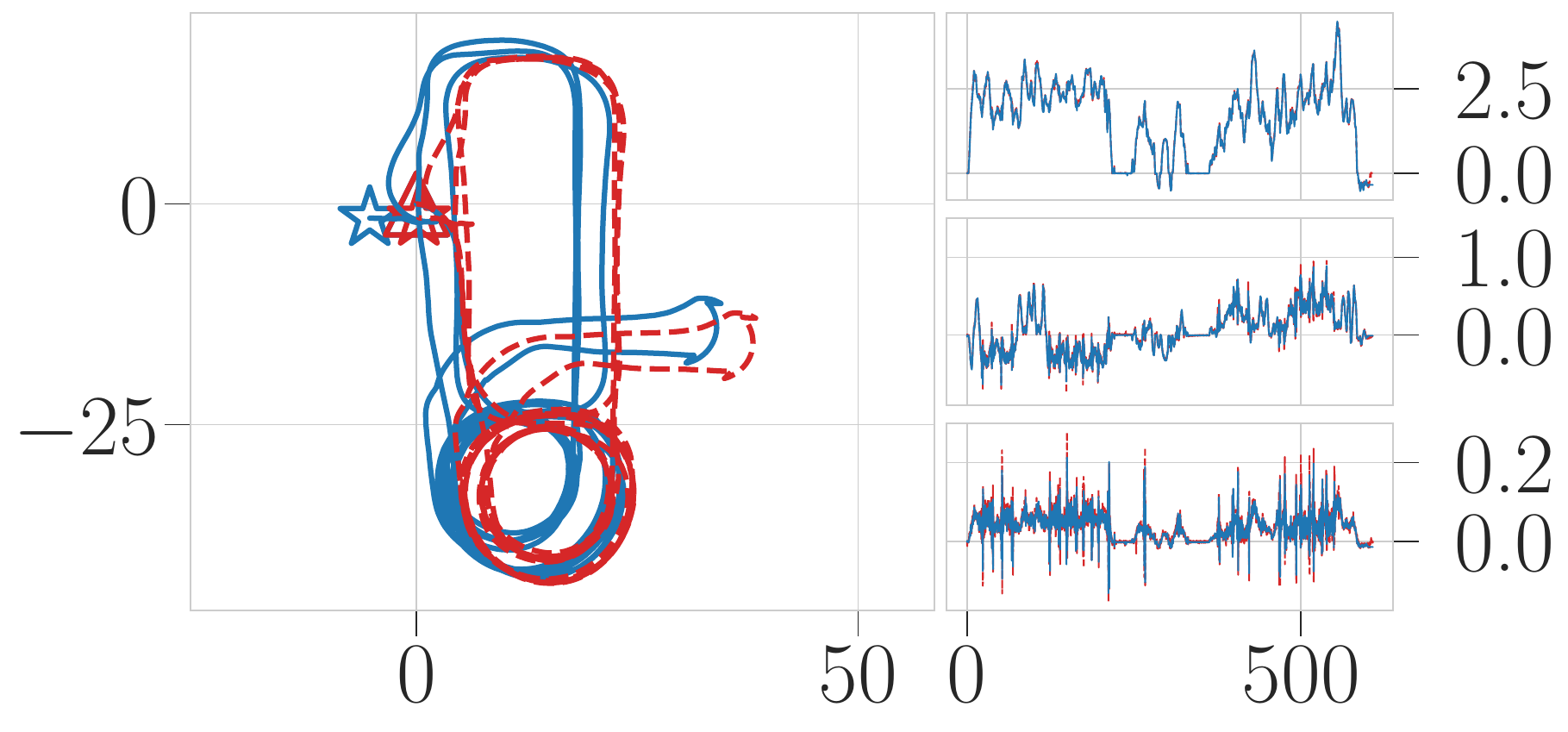}}
    \captionof{figure}{Evaluation of deep learning-based inertial odometry on multiple datasets. The left column shows the trajectory (X--Y plane in meters), while the right column presents the body-frame velocities along the $x$, $y$, and $z$ axes (velocity in m/s over time in seconds). The symbols $\triangle$ and ☆ denote the start and end points of each trajectory, respectively.}
    \label{fig:network_output}
\end{center}

\vspace{2mm}
}]
The supplementary material provides additional details supporting the main text, organized as follows:
\begin{itemize}
\item[\textbf{\ref{appendix:neural_network_evaluation}.}] Detailed evaluation of the deep learning-based inertial velocity network across multiple datasets.
\item[\textbf{\ref{appendix:real_time_performance}.}] Real-time and end-to-end runtime analysis of the full system.
\item[\textbf{\ref{appendix:sigma_ablation}.}] Ablation study on the scaling factor $\sigma_v$.
\item[\textbf{\ref{appendix:pvlio_reliability}.}] Reliability assessment of the LIO-estimated pseudo-GT labels used for training.
\item[\textbf{\ref{appendix:rotation_reliability}.}] Analysis of the neural velocity reliability under rotational motion.
\item[\textbf{\ref{appendix:switching_stability}.}] Temporal stability analysis of the binary degeneracy decision.
\item[\textbf{\ref{appendix:covariance_calibration}.}] Calibration analysis of the network-predicted velocity covariance.
\item[\textbf{\ref{appendix:private_dataset_detail}.}] Details of our private datasets and handheld setup.
\end{itemize}

\subsection{Neural Network Evaluation} \label{appendix:neural_network_evaluation}

\begin{table}[h]
\centering
\small
\caption{Quantitative Results of Neural Network (AVE $\downarrow$ [m/s], ATE $\downarrow$ [m], RTE $\downarrow$ [m]).}
\label{tab:network_evaluation}
\begin{tabular}{l|ccc}
\toprule[1pt]
\multirow{1}{*}{\textbf{Seq.}}   & AVE & ATE & RTE \\
\midrule
\rowcolor{gray!10}\textbf{Car}   &      &      & \\
\texttt{vehicle\_campus\_1}      & 0.07 & 4.54 & 0.43 \\
\texttt{vehicle\_multilayer\_0}  & 0.04 & 3.96 & 0.10 \\
\midrule
Avg.                             & 0.06 & 4.25 & 0.26\\
\midrule
\rowcolor{gray!10}\textbf{Others}                  &    &   &\\
\texttt{hku\_campus\_seq\_3}     & 0.14 & 2.52  & 0.35 \\
\texttt{hku\_park\_1}            & 0.19 & 7.35  & 0.69 \\
\texttt{hkust\_campus\_3}        & 0.20 & \worsecol 45.54 & 0.52 \\
\texttt{hku2}                    & 0.21 & 5.95  & 0.53 \\
\texttt{HKU\_Lecture\_Center\_2} & 0.26 & 3.15  & 0.73 \\
\texttt{HKU\_Main\_Building}     & \badcol 0.43 & 4.33  & \badcol 1.20 \\
\texttt{SYSU\_2}                 & \worsecol 0.64 & \badcol 11.49 & \worsecol 2.11 \\
\texttt{OffRoad\_5}              & 0.14 & 4.31  & 0.19 \\
\texttt{OffRoad\_6}              & 0.12 & 2.67  & 0.15 \\
\texttt{OffRoad\_7}              & 0.12 & 0.70  & 0.17 \\
\texttt{TunnelingTunnel\_4}      & 0.11 & 3.97  & 0.27 \\
\texttt{TunnelingTunnel\_5}      & 0.10 & 3.36  & 0.30 \\
\midrule
Avg.                             & 0.22 & 7.94  & 0.60 \\
\bottomrule[1pt]
\end{tabular}
\par\vspace{1mm}\noindent
\raggedleft
{\footnotesize Worst results are marked as \wtext{worst} and \wstext{second worst}.}
\vspace{-4mm}
\end{table}

As shown in Table~\ref{tab:network_evaluation}, the AVE for the car is below 0.1\,m/s for all cases, and the ATE remains within 5\,m.
As shown in Fig.~\ref{fig:network_output}(d) and (h), since the car exhibits low degrees of freedom (DoF) in its motion, the network predicts primarily along the $x$-direction, resulting in higher prediction accuracy.
For datasets using the built-in IMU of the Livox LiDAR, the ATE for \texttt{hkust\_campus\_3} is relatively large at 45.54\,m. However, as shown in (f), considering the long trajectory and accumulated errors, this value can be regarded as within an acceptable range.
In contrast, for \texttt{HKU\_Main\_Building} and \texttt{SYSU\_2}, both the AVE and RTE are observed to be relatively high.
As depicted in Fig.~\ref{fig:network_output}(a) and (e), the predicted trajectory deviates substantially from the ground truth (GT), and for \texttt{SYSU\_2}, the velocity prediction performance is notably degraded.
This can be attributed to the dataset containing insufficient data, as it includes rotations in place and backward movements.
Furthermore, while velocities along the $y$ and $z$ axes for \texttt{HKU\_Main\_Building} are accurately predicted, the $x$-axis prediction performance deteriorates, likely due to increased velocity variability induced by rotation.
As observed, prediction performance deteriorates for motions involving insufficient data or abrupt rotations. However, for the remaining datasets, the network accurately predicts even on unseen data.

\begin{figure}[t!]
    \centering
    \includegraphics[width=0.8\linewidth]{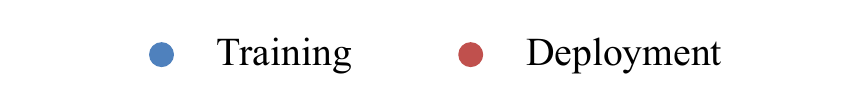} \\[1mm]
    \vspace{-4mm}
    \subfigure[Car]{\includegraphics[width=0.49\linewidth]{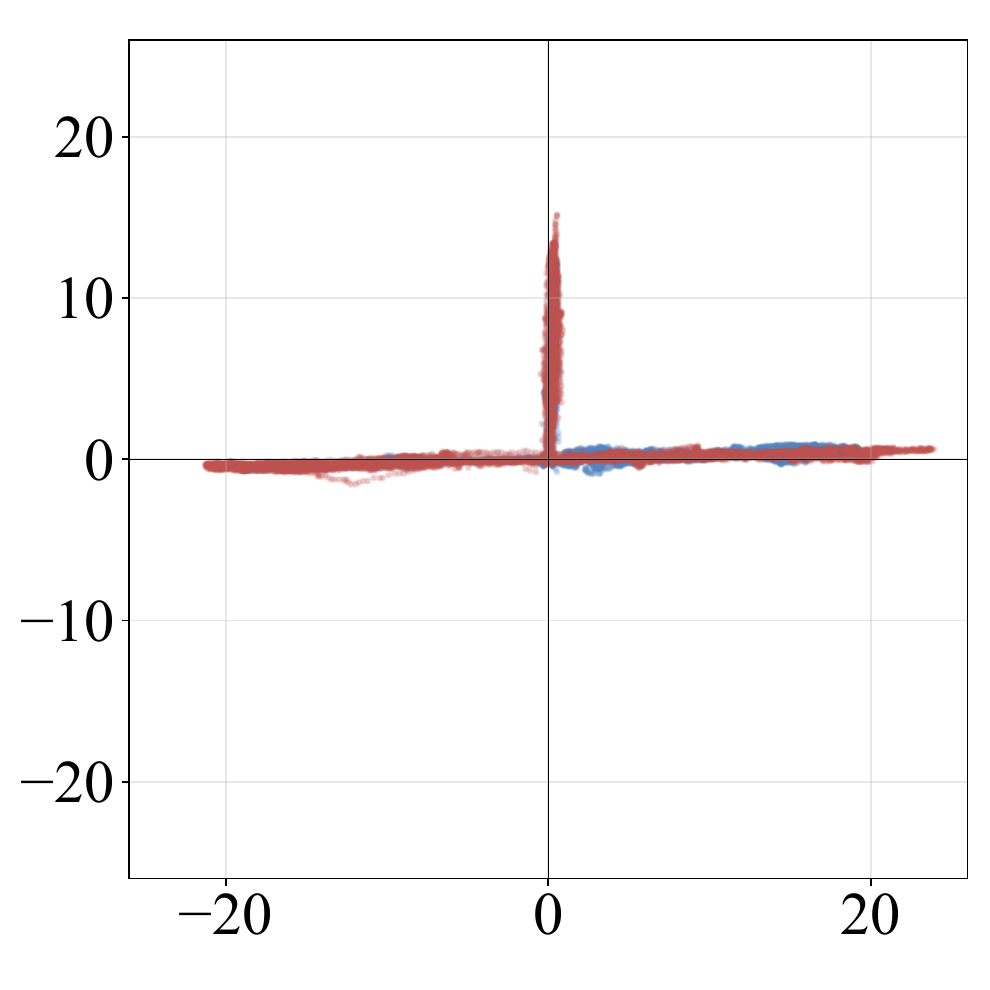}}
    \subfigure[Handheld]{\includegraphics[width=0.49\linewidth]{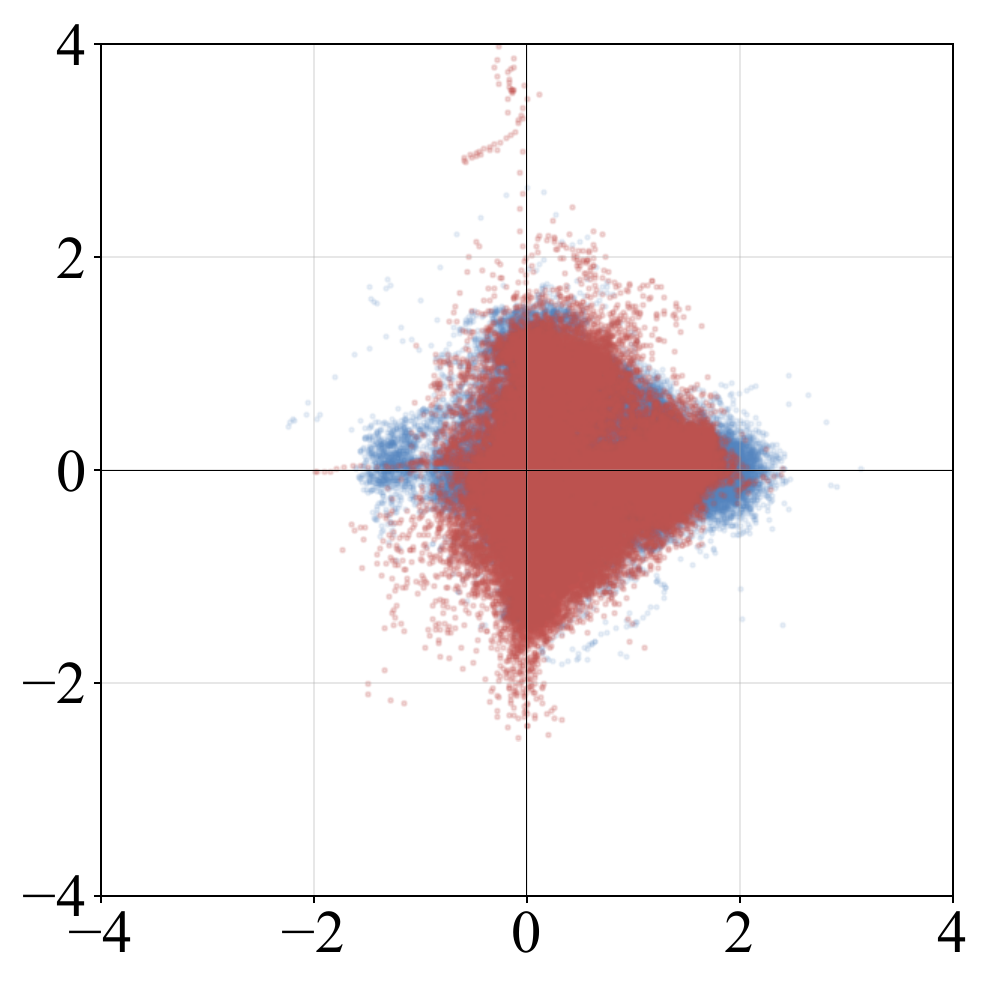}}
    \caption{Distribution of the body-frame velocity for the automotive and portable handheld platforms ($x$-axis: ${}^B\mathbf{v}_x$, $y$-axis: ${}^B\mathbf{v}_y$ [m/s]).}
    \label{fig:velocity_distribution}
    \vspace{-4mm}
\end{figure}

\begin{table*}[t!]
\small
\centering
\caption{Dataset split for training, evaluation, and degeneracy sequences.}
\label{tab:dataset}
\begin{tabular}{c c l l l}
\toprule
\textbf{Platform} & \textbf{Dataset} & \multicolumn{3}{c}{\textbf{Sequence}} \\
\cmidrule(lr){3-5}
& & \textbf{Train} & \textbf{Evaluation} & \textbf{Degeneracy (Deployment)} \\
\midrule
\multirow{10}{*}{Car}
                    & \multirow{3}{*}{FusionPortableV2~\cite{fusion_portable_v2}}
                        & \texttt{vehicle\_campus\_0}     & \texttt{vehicle\_campus\_1}    & \texttt{vehicle\_highway\_0} \\
                    &   & \texttt{vehicle\_highway\_1}    & \texttt{vehicle\_multilayer\_0}& \texttt{vehicle\_tunnel\_0}\\
                    &   & \texttt{vehicle\_street\_0}     &                                &                             \\
                        \cmidrule(lr){2-5}
                    & \multirow{2}{*}{UrbanNav~\cite{urban_nav_hk, urban_nav_hsu}}
                        &  \texttt{HK-TST}                &                               & \texttt{HK-CHTunnel} \\
                    &   & \texttt{HK-Mongkok}             &                               & \texttt{HK-Whampoa} \\
                        \cmidrule(lr){2-5}

                        & \multirow{3}{*}{GEODE~\cite{geode}}
                        & \texttt{Urban\_Tunnel\_1-3} (partial)&                               & \texttt{Urban\_Tunnel\_1-3} \\
                    &   & \texttt{Bridge\_1}              &                               & \texttt{Bridge\_2-3}\\
                    &   & \texttt{Bridge\_2-3} (partial)  &                               &                     \\
                        
                        \cmidrule(lr){1-5}
\multirow{21}{*}{\shortstack{Others}}
                    & \multirow{4}{*}{R3LIVE~\cite{r3live} }
                        &\texttt{hku\_campus\_seq\_0-2}   & \texttt{hku\_campus\_seq\_3}  & \texttt{degenerate\_seq\_0-2} \\
                    &   & \texttt{hku\_main\_building}    & \texttt{hku\_park\_1}         & \\
                    &   &\texttt{hku\_park\_0}            & \texttt{hkust\_campus\_3}     & \\
                    &   &\texttt{hkust\_campus\_0-2}      &                               &  \\
                        \cmidrule(lr){2-5} 
                        & \multirow{2}{*}{FAST-LIVO~\cite{fast_livo1}}
                        & \texttt{hku1}                   & \texttt{hku2}                 & \texttt{LiDAR\_Degenerate} \\
                    &   & \texttt{Visual\_Challenge}      &                               &  \\
                        \cmidrule(lr){2-5}
                        & \multirow{8}{*}{FAST-LIVO2~\cite{fast_livo2}}
                        & \texttt{CBD\_Building\_1}       &\texttt{HKU\_Lecture\_Center\_2}& \texttt{CBD\_Building\_2-3} \\
                    &   & \texttt{HKU\_Centennial\_Garden}& \texttt{HKU\_Main\_Building}  & \texttt{HIT\_Graffiti\_Wall\_1-4} \\
                    &   & \texttt{HKU\_Landmark}          &   \texttt{SYSU\_2}            & \texttt{HKU\_Cultural\_Center\_1-2}\\
                    &   & \texttt{HKU\_Lecture\_Center\_1}&                               & \texttt{Bright\_Screen\_Wall}\\
                    &   & \texttt{Retail\_Street}         &                               &  \\
                    &   & \texttt{HKUST\_Red\_Sculpture} &                                &  \\
                    &   & \texttt{SYSU\_1}               &                                &  \\
                        \cmidrule(lr){2-5}
                        & \multirow{3}{*}{GEODE~\cite{geode}}
                        & \texttt{OffRoad\_1-4}           & \texttt{OffRoad\_5-7}         & \texttt{ShieldTunnel\_1-6} \\
                    &   & \texttt{TunnelingTunnel\_1-3}   & \texttt{TunnelingTunnel\_4-5} & \texttt{FlatGroundSmooth} \\
                    &   &                                 &                               & \texttt{FlatGroundAggressive} \\
                        \cmidrule(lr){2-5}
                        & \multirow{2}{*}{Private}
                        & \texttt{handheld\_train}        &                               & \texttt{handheld\_light} \\
                    &   &                                 &                               & \texttt{handheld\_dark}  \\
                        
\bottomrule[1pt]
\multicolumn{5}{r}{\footnotesize ``partial'' indicates that the sequence is used as a training sequence from its start until the onset of degeneracy.}\\
\end{tabular}
\end{table*}
We additionally clarify our training strategy. The handheld, drone, and mobile-robot platforms used for training operate at speeds that do not exceed approximately 3\,m/s, whereas cars operate at substantially higher speeds; because the motion distributions differ markedly, as illustrated by the body-frame velocity distributions in Fig.~\ref{fig:velocity_distribution}, we train the car model separately from that of the other platforms. Moreover, even among car datasets, the IMU mounting direction differs (rear in GEODE, side in FusionPortable, and front in UrbanNav), which alters the input distribution; for example, the $x$-axis acceleration has opposite signs for front- and rear-mounted IMUs, whereas a side-mounted IMU is bounded along the $y$-axis. For these reasons, a separate model is trained per platform configuration. The detailed configuration of the training, evaluation, and degeneracy datasets is summarized in Table~\ref{tab:dataset}.
\subsection{Real-time Performance} \label{appendix:real_time_performance}
Experiments on open datasets were conducted on a desktop equipped with an Intel Core i9-13900KF CPU (24 cores, 32 threads, 5.8\,GHz), an NVIDIA GeForce RTX 3090 Ti GPU, 64\,GB RAM, and Ubuntu 20.04 LTS.
Experiments on private datasets were conducted on an Intel NUC (Intel Core i7-1165G7 CPU, 4 cores, 8 threads, up to 4.7\,GHz, 16\,GB RAM, x86\_64 architecture, Ubuntu).
\begin{table}[t!]
\caption{The network inference time (ms$\downarrow$).}
\centering
\small
\begin{tabular}{c|c|c|c}
\toprule[1pt]
\textbf{Processor} & \textbf{\makecell{$N=500$}} & \textbf{\makecell{$N=1000$}} & \textbf{\makecell{$N=2000$}}\\
\midrule
Desktop (CPU)        & 1.40 & 3.14 & 3.10 \\
Desktop (GPU)        & 1.93 & 2.25 & 3.10 \\
OnBoard (CPU)        & --   & 4.68 & -- \\
\bottomrule[1pt]
\multicolumn{4}{r}{\footnotesize ``--'' denotes a case that was not measured.}\\
\end{tabular}
\label{tab:latency}
\end{table}

\begin{table}[t!]
\caption{End-to-end per-frame runtime [ms] of the proposed system on representative degenerate sequences.}
\centering
\small
\begin{tabular}{c|c|c}
\toprule[1pt]
\textbf{Sequence} & \textbf{Mean} & \textbf{Max} \\
\midrule
\texttt{ShieldTunnel\_1}        & 41.24 & 89.75 \\
\texttt{Urban\_Tunnel\_2}       & 42.80 & 80.30 \\
\texttt{HIT\_Graffiti\_Wall\_2} & 28.08 & 92.49 \\
\bottomrule[1pt]
\end{tabular}
\label{tab:runtime_endtoend}
\end{table}

We measured real-time performance by averaging the runtime of the deep learning model over 200\,s on each processor.
As summarized in Table~\ref{tab:latency}, the system exhibits low runtimes on both desktop CPUs/GPUs and onboard low-performance CPUs, confirming its real-time capability. Interestingly, at $N=500$, the GPU consumes about $0.5$\,ms more than the CPU; this is possible because the network is highly lightweight and thus already fast enough on the CPU.
We additionally report the end-to-end per-frame runtime of the full system, including point cloud registration, the iterative ESKF, and network inference, in Table~\ref{tab:runtime_endtoend}, measured on the desktop. Both the mean and the maximum per-frame runtime remain within the $100$\,ms period of the LiDAR ($10$\,Hz), confirming real-time operation. The maximum occasionally approaches $100$\,ms when the number of voxelized points exceeds approximately $7{,}000$; in such cases the registration step, specifically the point-to-map correspondence search, dominates the runtime. This cost is inherent to the underlying LIO rather than introduced by our degeneracy-aware fusion, whose inference overhead is small as reported in Table~\ref{tab:latency}. This is further illustrated in Fig.~12 of GenZ-LIO~\cite{genzlio}, where the correspondence search accounts for a large portion of the registration time.
\subsection{Ablation on the Scaling Factor $\sigma_v$} \label{appendix:sigma_ablation}
In this section, an ablation on the scaling factor $\sigma_v$ applied to the neural-velocity covariance $\mathbf{\Sigma}^\text{NN}_{v, M}$ is presented to explain why this scaling is introduced. The ablation uses three sequences with different motions and platforms: \texttt{vehicle\_tunnel\_0} represents the automotive platform, \texttt{ShieldTunnel\_1} is a handheld sequence with predominantly straight motion, and \texttt{ShieldTunnel\_5} is a sequence in which a narrow-FOV LiDAR faces a single lateral plane throughout.

\begin{figure}[t!]
\centering
\includegraphics[width=0.6\linewidth]{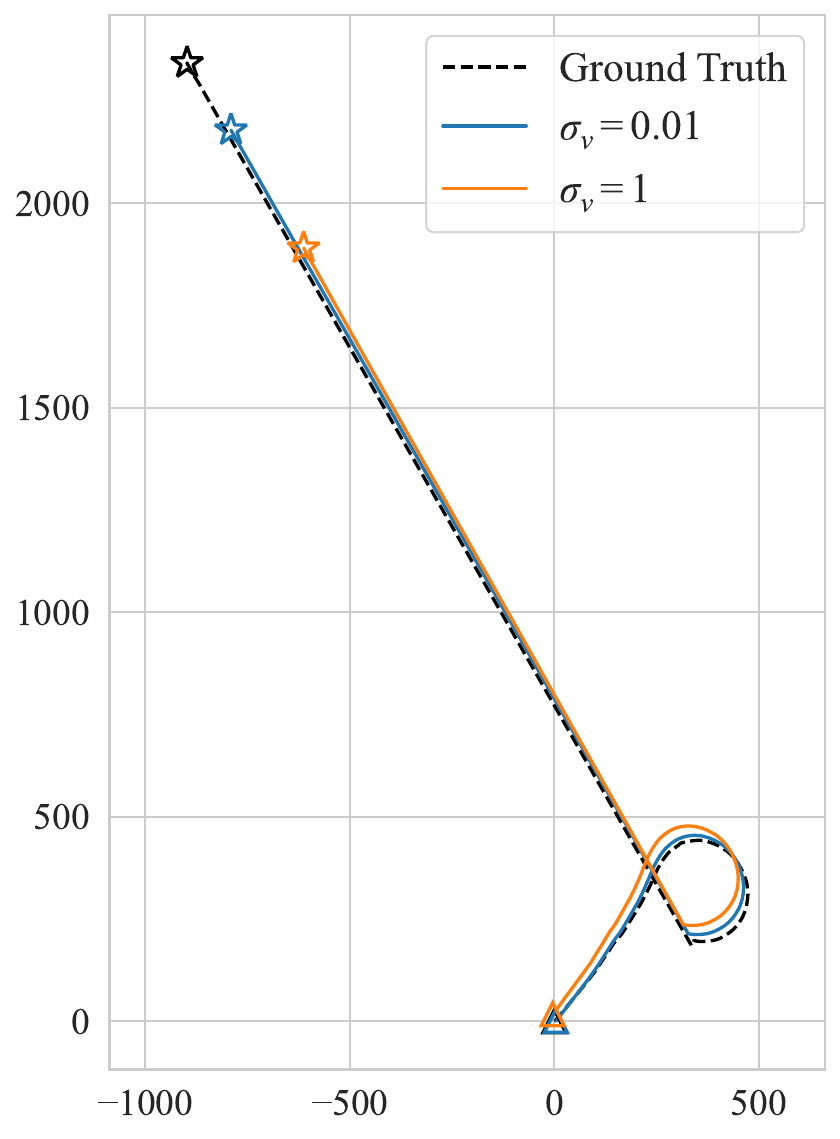}
\caption{Estimated trajectory on \texttt{vehicle\_tunnel\_0} for $\sigma_v=1$ and $\sigma_v=0.01$ (X–Y plane in meters). The symbols $\triangle$ and ☆ denote the start and end points, respectively.}
\label{fig:sigma_ablation}
\vspace{-4mm}
\end{figure}

As summarized in Table~\ref{tab:sigma_sensitivity}, when $\sigma_v=1$, the ATE of \texttt{ShieldTunnel\_1} and \texttt{vehicle\_tunnel\_0} becomes markedly larger compared to the other values. We found that the neural velocity itself is estimated without major issues in this case; however, the state covariance $\mathbf{P}$ becomes smaller than $\mathbf{\Sigma}^\text{NN}_{v, M}$, so the weight assigned to the neural velocity in the ESKF update becomes too small and the neural velocity is barely reflected in the estimate. As shown in Fig.~\ref{fig:sigma_ablation}, the estimated trajectory for $\sigma_v=1$ terminates at a shorter point than for $\sigma_v=0.01$, which is also attributable to this effect.
\begin{table}[t!]
\caption{Sensitivity of the ATE (m$\downarrow$) to the scaling factor $\sigma_v$, evaluated on a logarithmic scale over representative sequences spanning different motions and platforms. We adopt $\sigma_v=0.01$.}
\centering
\small
\begin{tabular}{c|cccc}
\toprule[1pt]
\multirow{2}{*}{\textbf{Sequence}} & \multicolumn{4}{c}{\textbf{Scaling factor $\sigma_v$}} \\
\cmidrule(lr){2-5}
 & $0.001$ & $0.01$ & $0.1$ & $1$ \\
\midrule
\texttt{ShieldTunnel\_1}    & 12.81 & 12.69 & 11.25 & \worsecol 39.87 \\
\texttt{ShieldTunnel\_5}    & \worsecol 10.75 & 8.73 & 7.34 & 7.85 \\
\texttt{vehicle\_tunnel\_0} & 56.73 & 53.05 & 71.01 & \worsecol 148.17 \\
\bottomrule[1pt]
\end{tabular}
\par\vspace{1mm}\noindent
\raggedleft
{\footnotesize The worst case for each sequence is marked as \wtext{worst}.}
\vspace{-4mm}
\label{tab:sigma_sensitivity}
\end{table}

In contrast, \texttt{ShieldTunnel\_5} does not exhibit this degradation, which stems from how long the degeneracy persists. When the degenerate interval is short, the above effect accumulates only mildly; indeed, \texttt{ShieldTunnel\_5} is about $250$\,s long, whereas the other two sequences are more than twice as long. This observation motivates the introduction of $\sigma_v$. Conversely, setting $\sigma_v$ too small amounts to fully trusting the neural velocity, and the performance on \texttt{ShieldTunnel\_5} degrades at $\sigma_v=0.001$.

All experiments in this paper use $\sigma_v=0.01$. Nevertheless, $\sigma_v=0.1$ yields better performance on \texttt{ShieldTunnel\_1} and \texttt{ShieldTunnel\_5}, whereas $\sigma_v=0.01$ is better on \texttt{vehicle\_tunnel\_0}. Because cars generally move at higher speeds, the scale of $\mathbf{\Sigma}^\text{NN}_{v, M}$ formed during training is larger than for handheld platforms, so using a smaller $\sigma_v$ for the car helps match its scale to that of the state covariance. Finally, our system is built on top of PV-LIO, whose LiDAR-measurement noise covariance is algorithmically distributed to be smaller than that of FAST-LIO2 under nominal conditions; the appropriate value of $\sigma_v$ therefore needs to be tuned according to how the underlying LIO sets the LiDAR-measurement noise covariance.

\begin{table*}[t!]
\caption{Reliability of the PV-LIO pseudo-ground-truth labels, evaluated against the dataset ground truth on the well-conditioned training and evaluation sequences (ATE [m], RTE [m], and the average $\|\Delta \mathbf{t}\|_2$ [m]).}
\centering
\small
\begin{tabular}{c|c|c|c|c}
\toprule[1pt]
\textbf{Dataset} & \textbf{Sequence} & \textbf{ATE} & \textbf{RTE} & $\|\Delta \mathbf{t}\|_2$ \\
\midrule
\multirow{5}{*}{FusionPortableV2} & \texttt{vehicle\_campus\_0}     & 10.27 & 1.29 & 0.02 \\
                                  & \texttt{vehicle\_highway\_1}    & 28.81 & 3.17 & 0.08 \\
                                  & \texttt{vehicle\_street\_0}     & 37.74 & 1.73 & 0.02 \\
                                  & \texttt{vehicle\_campus\_1}     & 6.33  & 1.32 & 0.02 \\
                                  & \texttt{vehicle\_multilayer\_0} & 0.31  & 0.01 & 0.01 \\
\midrule
\multirow{2}{*}{UrbanNav} & \texttt{HK-TST}     & 8.06 & 6.51 & 0.24 \\
                          & \texttt{HK-Mongkok} & 2.64 & 1.84 & 0.09 \\
\midrule
\multirow{13}{*}{GEODE} & \texttt{Bridge\_1}          & 41.44 & 1.60 & 0.07 \\
                        & \texttt{OffRoad\_1}         & 0.62 & 0.12 & 0.01 \\
                        & \texttt{OffRoad\_2}         & 0.65 & 0.16 & 0.02 \\
                        & \texttt{OffRoad\_3}         & 0.66 & 0.23 & 0.03 \\
                        & \texttt{OffRoad\_4}         & 0.62 & 0.27 & 0.03 \\
                        & \texttt{OffRoad\_5}         & 0.66 & 0.18 & 0.03 \\
                        & \texttt{OffRoad\_6}         & 0.66 & 0.18 & 0.03 \\
                        & \texttt{OffRoad\_7}         & 0.63 & 0.23 & 0.03 \\
                        & \texttt{TunnelingTunnel\_1} & 0.11 & \textcolor{gray}{\sout{3.24}} & 0.10 \\
                        & \texttt{TunnelingTunnel\_2} & 1.94 & \textcolor{gray}{\sout{1.72}} & 0.08 \\
                        & \texttt{TunnelingTunnel\_3} & 0.31 & \textcolor{gray}{\sout{2.76}} & 0.09 \\
                        & \texttt{TunnelingTunnel\_4} & 0.31 & \textcolor{gray}{\sout{2.43}} & 0.09 \\
                        & \texttt{TunnelingTunnel\_5} & 0.30 & \textcolor{gray}{\sout{1.56}} & 0.06 \\
\bottomrule[1pt]
\multicolumn{5}{r}{\footnotesize \textcolor{gray}{\sout{RTE}} values are not reliable because the dataset provides no rotation.}\\
\end{tabular}
\vspace{-4mm}
\label{tab:pvlio_reliability}
\end{table*}

\subsection{Reliability of the Pseudo-Ground-Truth Labels} \label{appendix:pvlio_reliability}
\begin{figure}[t!]
\centering
\includegraphics[width=0.8\linewidth]{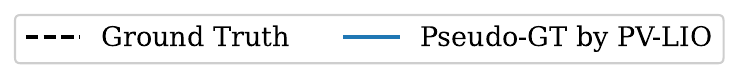} \\[1mm]
\subfigure[Per-axis position over time]{\includegraphics[width=0.54\linewidth]{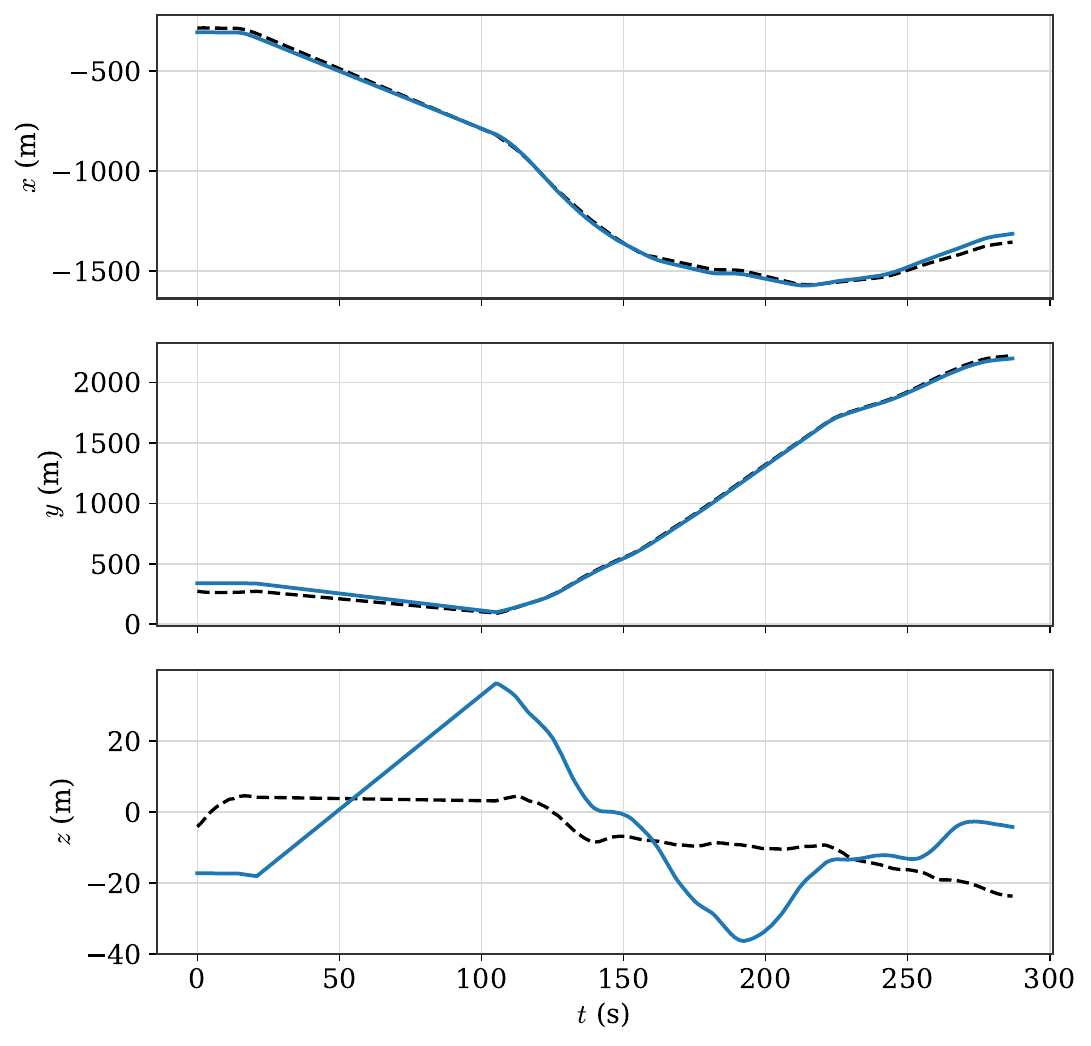}}
\hfill
\subfigure[Estimated trajectory]{\includegraphics[width=0.42\linewidth]{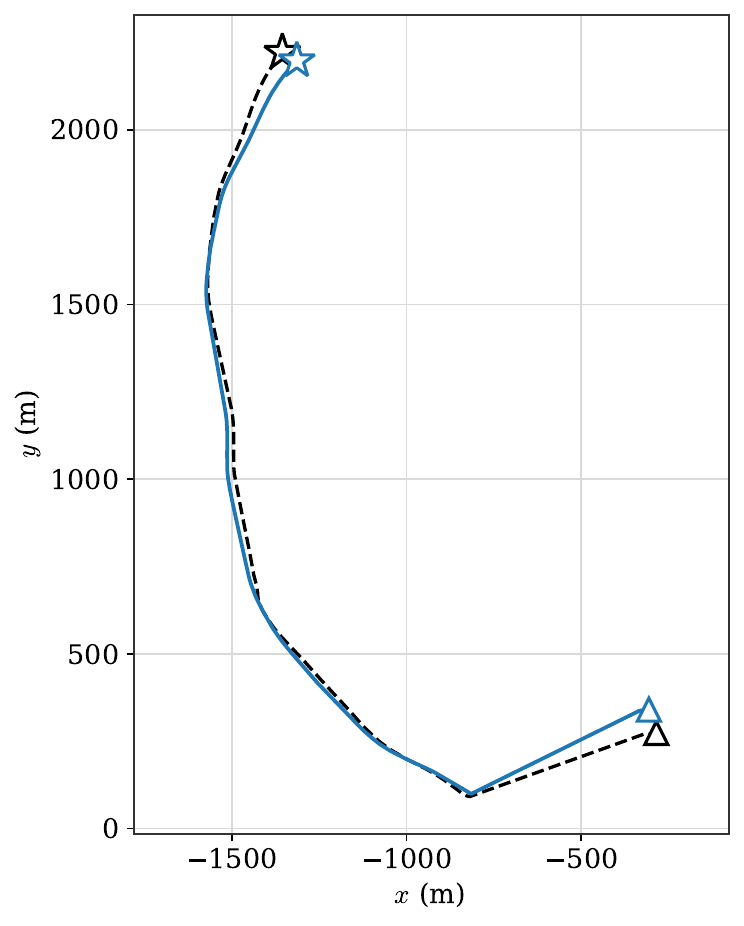}}
\caption{PV-LIO estimate on \texttt{vehicle\_highway\_1}, which has a long trajectory and exhibits pronounced drift along the $z$-axis. The trajectory is shown in the X–Y plane (in meters), where $\triangle$ and ☆ denote the start and end points, respectively.}
\label{fig:highway_drift}
\vspace{-6mm}
\end{figure}

As summarized in Table~\ref{tab:pvlio_reliability}, the global ATE is relatively large on the vehicle sequences, because these sequences not only have long trajectories but also exhibit pronounced drift along the $z$-axis, as shown in Fig.~\ref{fig:highway_drift}. To assess the local accuracy that is relevant to our velocity labels, the RTE should instead be examined, as it reflects local motion rather than the accumulated global drift. On the car sequences, the RTE is relatively small given their long trajectories, and it remains small on the other sequences as well. On some sequences, however, a reliable RTE cannot be obtained: for the UrbanNav dataset the GT is provided at only 1\,Hz, and for the GEODE \texttt{TunnelingTunnel} sequences rotation is not provided in the GT, so an EVO-based RTE evaluation is unreliable or inapplicable. For these sequences we instead report the per-frame translation norm $\|\Delta\mathbf{t}\|_2$, which captures only the magnitude of the per-frame translation and not its body-frame direction; this value remains very small relative to the GT. Although LIO-estimated pseudo-GT labels cannot achieve zero error, they are produced by the LIO system operating in well-conditioned scenarios, and recovering this well-conditioned LIO accuracy under LiDAR degeneracy is precisely the objective of our method, so the labels serve as a reasonable training target. In addition, training on the outputs of a LIO or SLAM system is an established practice: ININ-LIO~\cite{inin_lio} partly uses a LIO algorithm as GT, and TartanIMU~\cite{tartan_imu}, an online deep inertial learning method, is trained on SLAM outputs.

\subsection{Neural Velocity Reliability under Rotational Motion} \label{appendix:rotation_reliability}
\begin{figure*}[t!]
\centering
\subfigure[]{\includegraphics[width=0.32\linewidth]{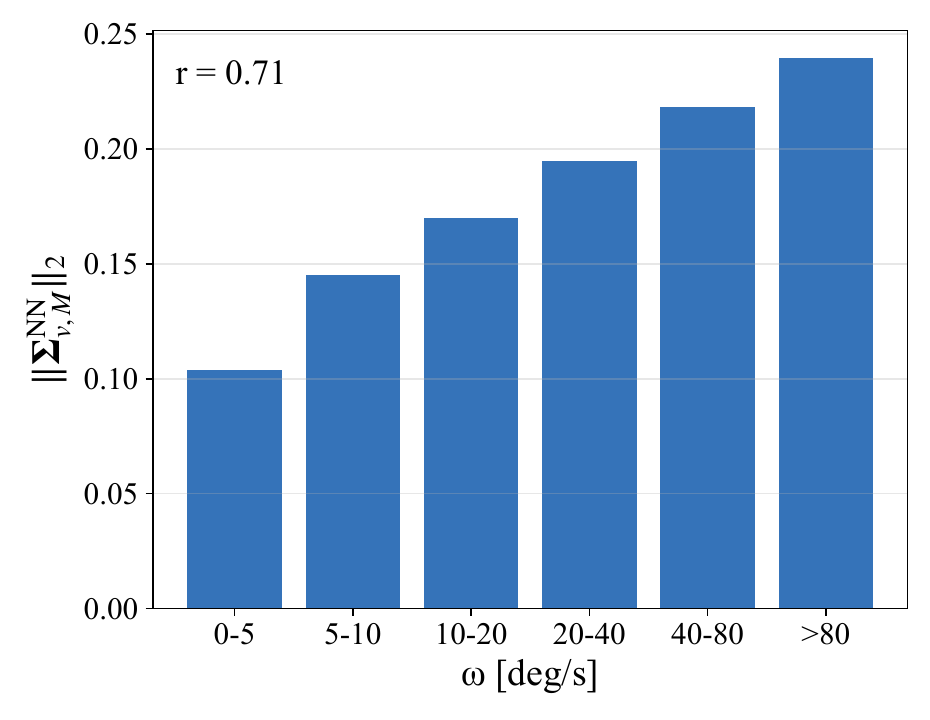}}
\subfigure[]{\includegraphics[width=0.332\linewidth]{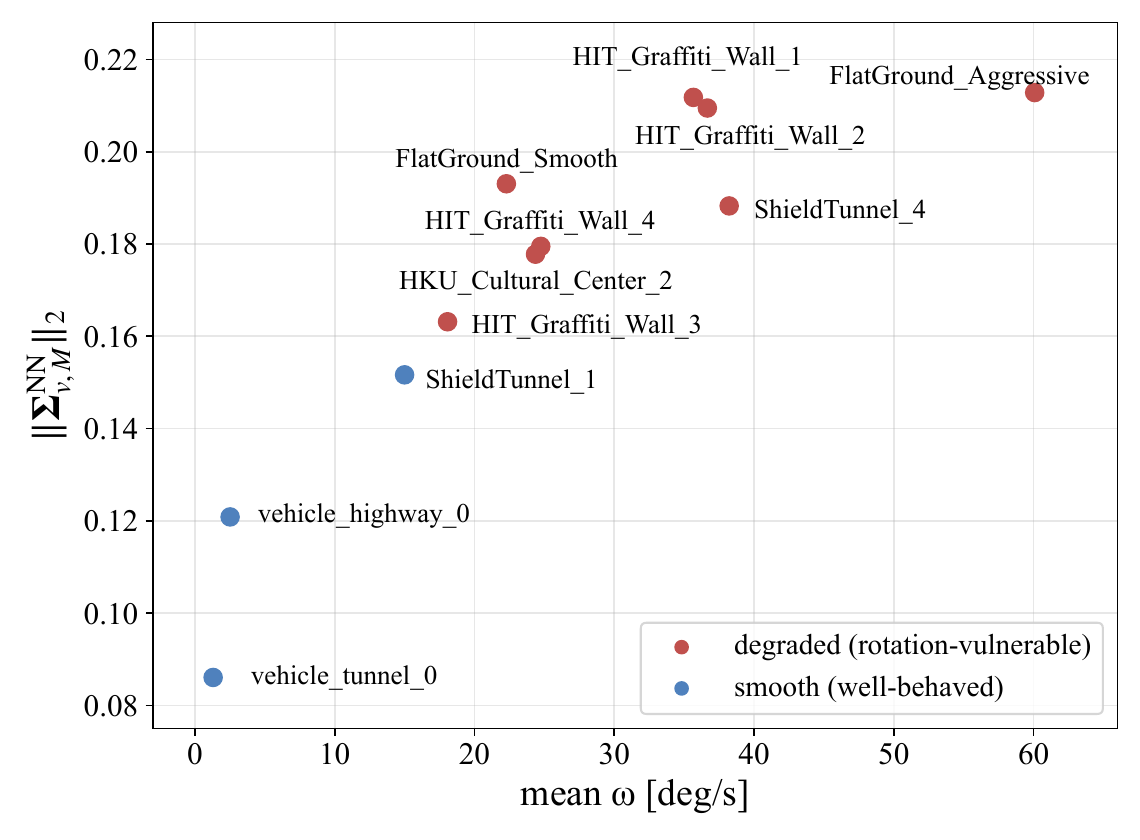}}
\subfigure[]{\includegraphics[width=0.324\linewidth]{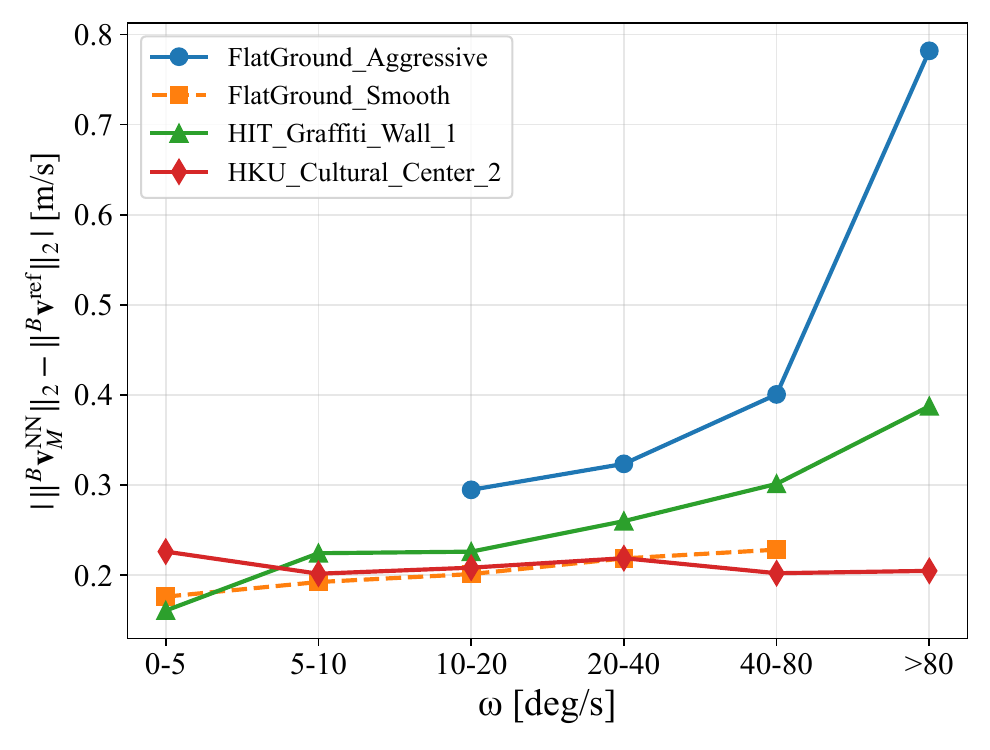}}
\caption{Neural-velocity reliability versus the rotation rate $\omega$. (a) Network-reported uncertainty $\|\mathbf{\Sigma}^\text{NN}_{v, M}\|_2$ pooled over all frames ($r=0.71$). (b) Per-sequence mean $\omega$ versus mean $\|\mathbf{\Sigma}^\text{NN}_{v, M}\|_2$ (red: rotation-vulnerable, blue: smooth). (c) Neural speed error $|\,\|{}^B\mathbf{v}^\text{NN}_M\|_2-\|{}^B\mathbf{v}^\text{ref}\|_2\,|$ within sequences with a reliable reference, where the reference is public GT when available and otherwise FAST-LIVO2 or PV-LIO.}
\label{fig:rot_uncertainty}
\end{figure*}

\begin{figure*}[t!]
\centering
\subfigure[]{\includegraphics[width=0.48\linewidth]{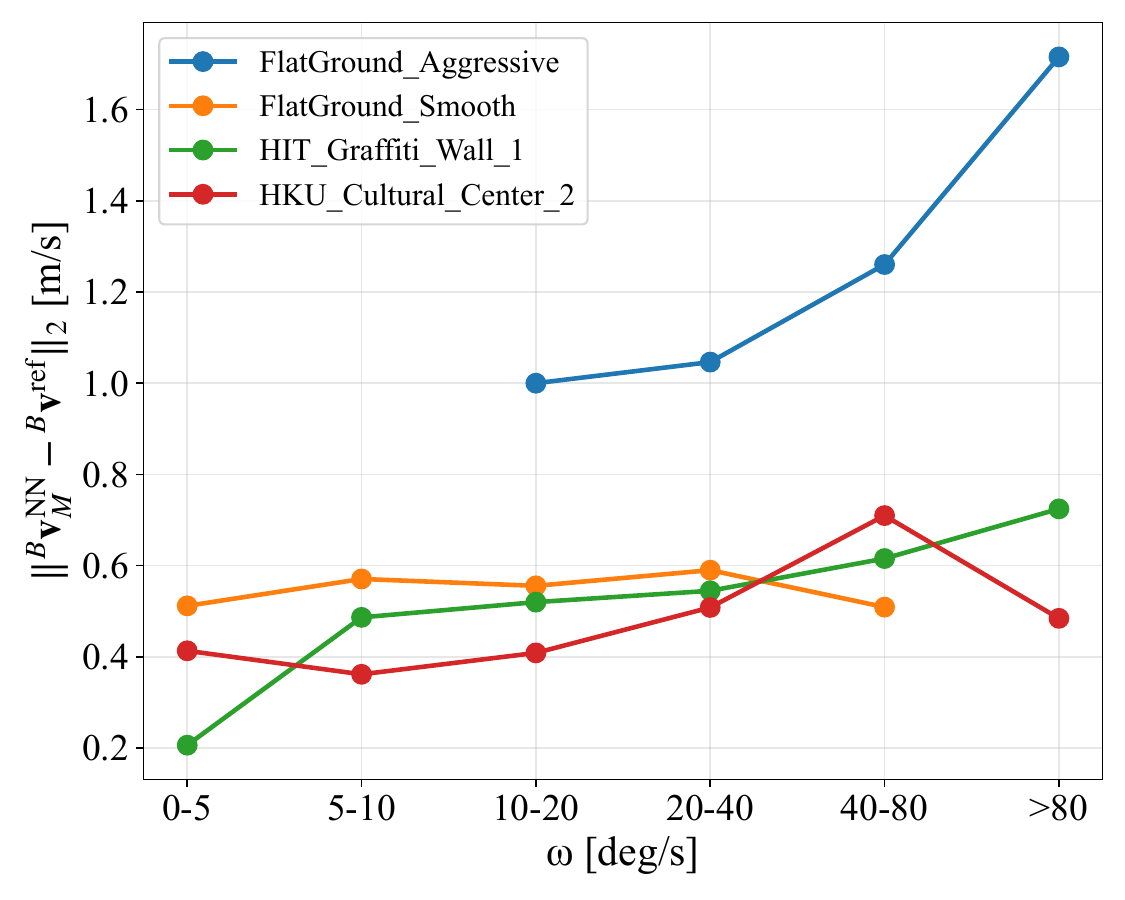}}
\subfigure[]{\includegraphics[width=0.48\linewidth]{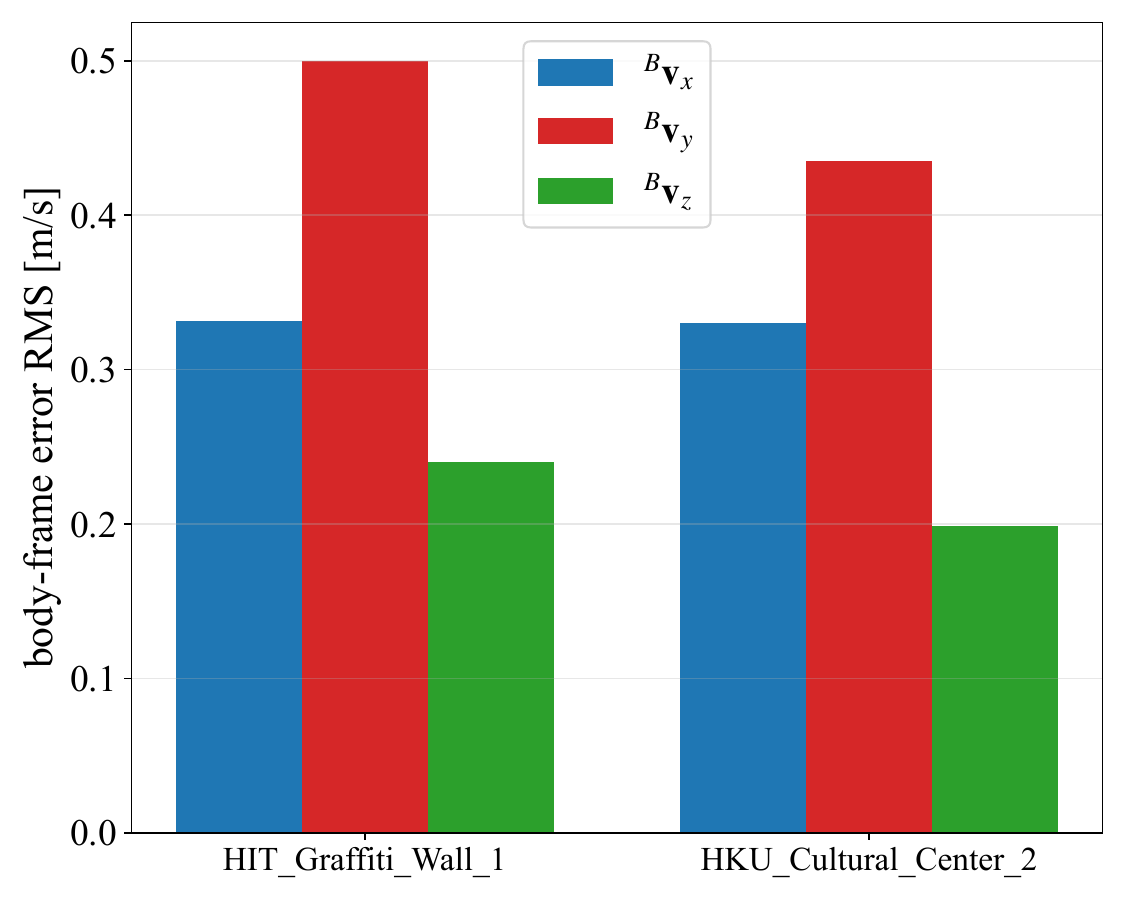}}
\caption{Body-frame velocity error versus the rotation rate $\omega$. (a) Body-frame velocity vector error $\|{}^B\mathbf{v}^\text{NN}_M-{}^B\mathbf{v}^\text{ref}\|_2$. (b) Per-axis error root-mean-square (RMS) along the forward ($x$), lateral ($y$), and up ($z$) axes for sequences sharing the IMU frame with their reference; the lateral axis dominates.}
\label{fig:rot_bodyframe}
\end{figure*}

We quantitatively analyze how the platform's rotation affects the reliability of the learned velocity. The rotation rate $\omega$ is defined as $\omega = \sum\limits_{i}\|{}^B\mathbf{w}_i\|_2$, the sum of the $L_2$ norms of the IMU angular velocity ${}^B\mathbf{w}_i$ over each LiDAR-frame (100\,ms) interval $[t_k, t_{k+1}]$.

First, $\mathbf{\Sigma}^\text{NN}_{v, M}$, the velocity covariance predicted by the network, is constructed as a diagonal matrix from a predicted vector, and its diagonal-element norm $\|\mathbf{\Sigma}^\text{NN}_{v, M}\|_2$ increases monotonically with $\omega$. Pooled over all frames of all sequences, the mean $\|\mathbf{\Sigma}^\text{NN}_{v, M}\|_2$ rises from $0.10$ at $\omega<5$\,deg/s to $0.24$ at $\omega>80$\,deg/s, about $2.4\times$, with a per-frame Pearson correlation of $r(\omega,\|\mathbf{\Sigma}^\text{NN}_{v, M}\|_2)=0.71$ (Fig.~\ref{fig:rot_uncertainty}(a)). At the sequence level (Fig.~\ref{fig:rot_uncertainty}(b)), smooth sequences such as the vehicles cluster in the low-$\omega$, low-uncertainty region, whereas rotation-heavy sequences, such as \texttt{FlatGround}, \texttt{HIT\_Graffiti\_Wall}, and \texttt{ShieldTunnel\_4}, occupy the high-$\omega$, high-uncertainty region, yielding a clear separation between the two groups. In other words, the network predicts a larger velocity covariance as the rotation increases.

This increase in $\|\mathbf{\Sigma}^\text{NN}_{v, M}\|_2$ is reflected in the actual estimation error. On sequences with a reliable reference (pseudo-GT), the speed error $|\,\|{}^B\mathbf{v}^\text{NN}_M\|_2-\|{}^B\mathbf{v}^\text{ref}\|_2\,|$ grows monotonically with $\omega$, reaching $0.78$\,m/s on \texttt{FlatGround\_Aggressive} and $0.39$\,m/s on \texttt{HIT\_Graffiti\_Wall\_1}, while remaining around $0.2$\,m/s for smooth, low-rotation motion (Fig.~\ref{fig:rot_uncertainty}(c)). The body-frame velocity vector error $\|{}^B\mathbf{v}^\text{NN}_M-{}^B\mathbf{v}^\text{ref}\|_2$ follows the same trend (Fig.~\ref{fig:rot_bodyframe}(a)), and a per-axis breakdown shows that the lateral ($y$) axis dominates the error (Fig.~\ref{fig:rot_bodyframe}(b)), indicating that the lateral velocity is the most vulnerable component under rotation.

In summary, the reliability of the neural velocity degrades monotonically as the rotation becomes more aggressive, although its onset depends on the platform speed and the motion type and thus cannot be characterized by a single threshold on $\omega$. Furthermore, because these quantities are aggregated over all frames of a sequence, a pronounced degradation confined to a few frames can be obscured in the aggregate while still affecting the overall performance. We therefore explicitly acknowledge that our neural network can be vulnerable to rotational motion, which quantitatively supports the qualitative discussion in Section~\ref{subsec:46_failure_analysis}.

\definecolor{degenblue}{RGB}{0,114,178}
\definecolor{wellgray}{RGB}{237,237,237}
\begin{figure*}[t!]
\centering
\includegraphics[width=0.92\linewidth]{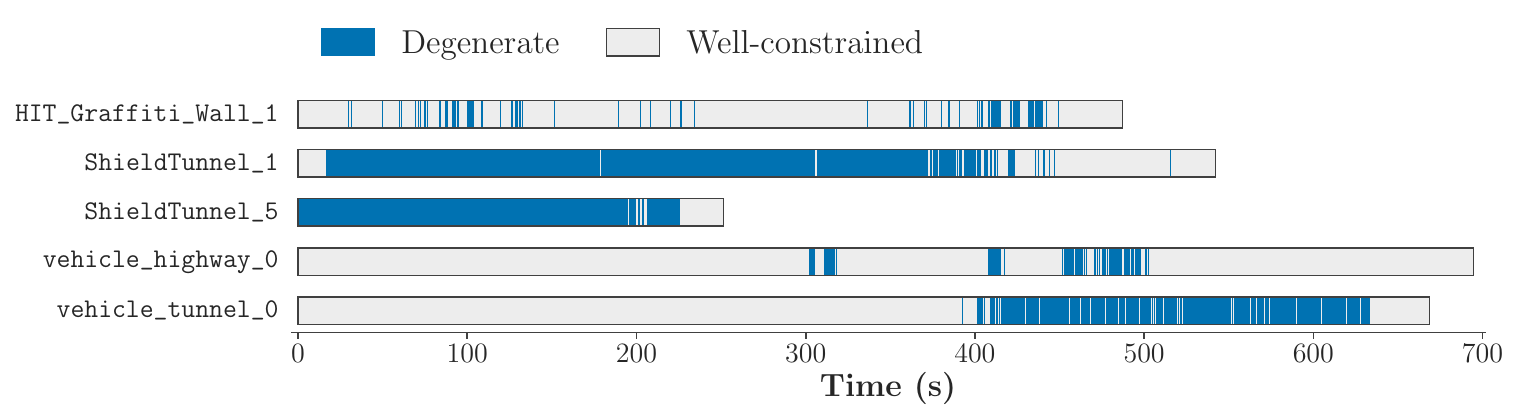}
\caption{Per-scan degeneracy decision over time on the five sequences of Table~\ref{tab:degeneracy_switching}. \colorbox{degenblue}{\textcolor{white}{Blue}} denotes scans classified as degenerate and \colorbox{wellgray}{gray} denotes well-conditioned scans.}
\label{fig:degeneracy_timeline}
\vspace{-2mm}
\end{figure*}

\subsection{Temporal Stability of the Degeneracy Decision} \label{appendix:switching_stability}
Because the degeneracy criterion in (\ref{eq17}) is binary, the decision may change frequently when the localizability contributions lie close to the thresholds.
We therefore analyze the temporal behavior of the decision.
Let $d(t_k) \in \{0, 1\}$ denote the decision for the LiDAR scan at time $t_k$, where $d(t_k) = 1$ if at least one translational direction is classified as degenerate, which is the condition that triggers the state correction in Algorithm~\ref{alg:degeneracy_aware_optim}.
We evaluate $d(t_k)$ on five sequences that span different platforms and environments.

The statistics are summarized in Table~\ref{tab:degeneracy_switching}, and the corresponding decisions over time are shown in Fig.~\ref{fig:degeneracy_timeline}.
\texttt{ShieldTunnel\_1} and \texttt{ShieldTunnel\_5} remain degenerate for $70.8\%$ and $86.4\%$ of their scans, and their longest consecutive degenerate intervals last $162.0$\,s and $195.2$\,s; on \texttt{ShieldTunnel\_5}, this single interval accounts for $195.2$\,s of the $217.1$\,s of total degenerate time, or $89.9\%$.
\texttt{HIT\_Graffiti\_Wall\_1}, \texttt{vehicle\_highway\_0}, and \texttt{vehicle\_tunnel\_0} are instead flagged for $9.0\%$, $5.2\%$, and $24.7\%$ of their scans, and their longest intervals last only $4.6$\,s, $6.8$\,s, and $11.1$\,s.
The switching rate varies accordingly, from $7.88$ transitions per minute on \texttt{ShieldTunnel\_5} to $61.20$ on \texttt{vehicle\_tunnel\_0}; because the decision is evaluated at the $10$\,Hz scan rate, it changes on $1.3\%$ to $10.2\%$ of the scans.

These differences follow from the geometry of each scene rather than from instability at the decision threshold.
On \texttt{HIT\_Graffiti\_Wall\_1}, the handheld platform faces the wall only intermittently while rotating, so the degenerate motion itself is short.
On \texttt{vehicle\_highway\_0}, the decision toggles frequently because the streetlamps along the road periodically enter the field of view and provide geometric constraints.
A related mechanism governs \texttt{vehicle\_tunnel\_0}, in which the degeneracy is confined to a single stretch of the sequence, yet the vehicles passing through the tunnel intermittently add structure to an otherwise featureless scene and thereby clear the degenerate flag.

\begin{figure}[t!]
\centering
\includegraphics[width=\linewidth]{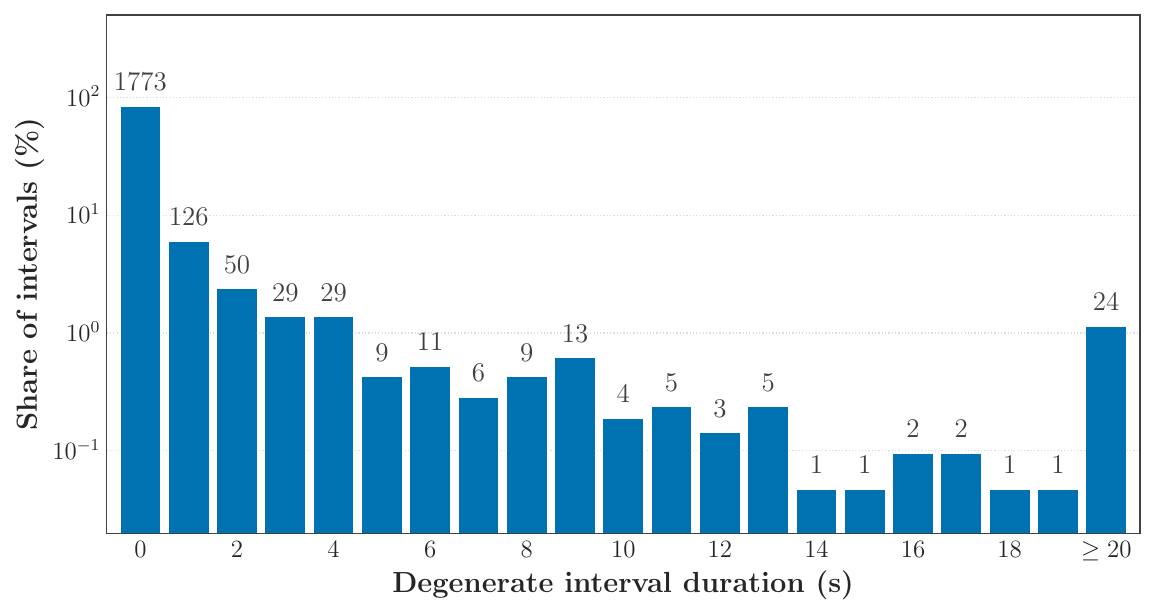}
\caption{Distribution of the consecutive degenerate interval duration, pooled over all evaluated sequences and binned at 1\,s. The vertical axis is the share of intervals on a logarithmic scale, and the number above each bar is the corresponding interval count. Short intervals dominate the interval count, although they account for $12.7\%$ of the total degenerate time.}
\label{fig:degeneracy_runlength}
\vspace{-2mm}
\end{figure}

However, the degenerate and well-conditioned states can also alternate frequently beyond the situations described above.
Fig.~\ref{fig:degeneracy_runlength} shows the interval duration distribution pooled over all evaluated sequences: of the $2{,}104$ degenerate intervals, $1{,}773$ are shorter than $1$\,s, whereas a small number of intervals exceed $20$\,s.
These short intervals nevertheless account for only $12.7\%$ of the total degenerate time.
In other words, short intervals dominate the interval count while long intervals dominate the total degenerate time.
We therefore acknowledge that isolated short decisions do occur, which is an inherent limitation of any binary switching criterion and whose extent depends on the thresholds chosen for that criterion.

\definecolor{neuralorange}{RGB}{213,94,0}
\definecolor{stateblue}{RGB}{0,114,178}
\definecolor{degenshade}{RGB}{224,237,245}
\begin{figure}[t!]
\centering
\includegraphics[width=\linewidth]{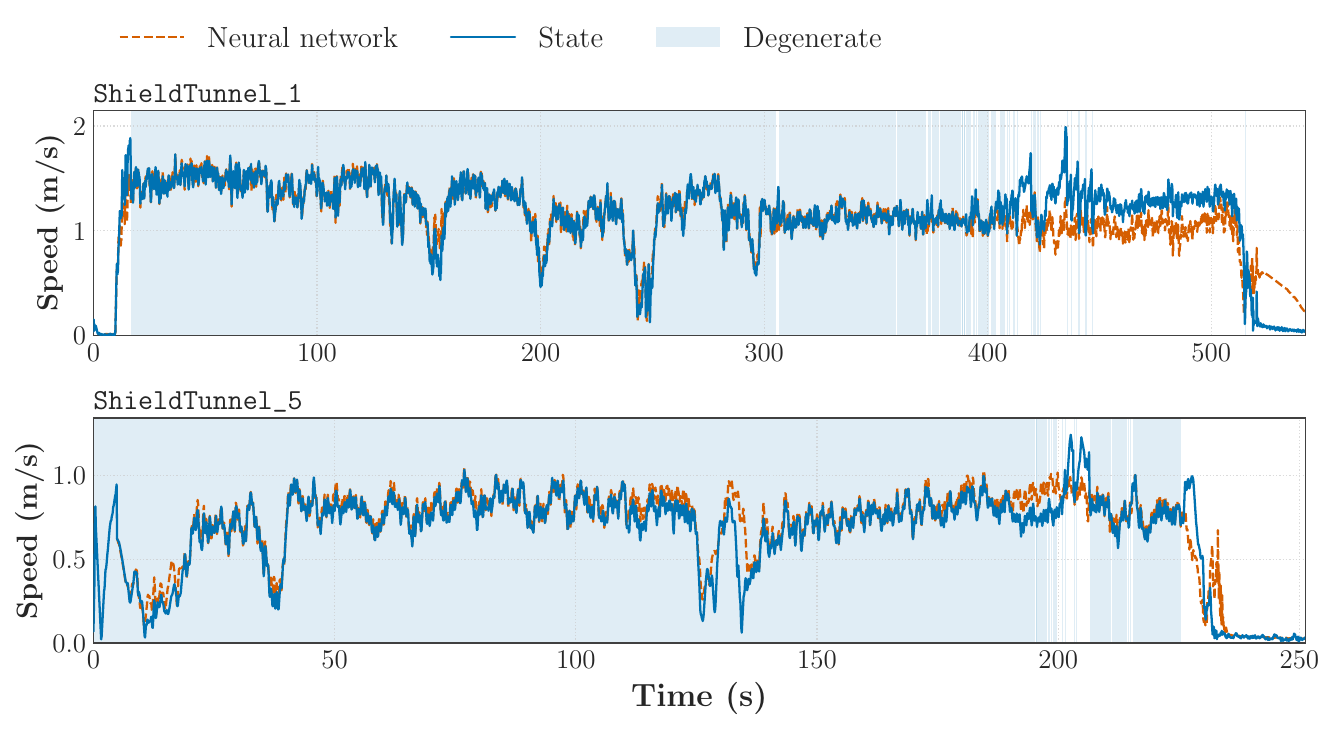}
\caption{\textcolor{neuralorange}{Neural speed} $\|{}^B\mathbf{v}^\text{NN}_M\|_2$ (dashed) and \textcolor{stateblue}{the speed of the state} (solid) on \texttt{ShieldTunnel\_1} (top) and \texttt{ShieldTunnel\_5} (bottom). The \colorbox{degenshade}{shaded regions} denote the scans classified as degenerate.}
\label{fig:degeneracy_neural_gap}
\vspace{-2mm}
\end{figure}

\begin{table}[t!]
\caption{Temporal stability of the binary degeneracy decision on five representative sequences spanning different platforms and environments. Duty denotes the fraction of scans classified as degenerate, and the switching rate counts the degenerate $\leftrightarrow$ well-conditioned transitions. Med., $p_{90}$, and Max denote the median, the 90th percentile, and the maximum duration of a consecutive degenerate interval, respectively.}
\centering
\footnotesize
\setlength{\tabcolsep}{4pt}
\resizebox{\columnwidth}{!}{%
\begin{tabular}{l|ccc|ccc}
\toprule[1pt]
\multirow{2}{*}{\textbf{Sequence}} & \textbf{Dur.} & \textbf{Duty} & \textbf{Switch.} & \multicolumn{3}{c}{\textbf{Degenerate interval [s]}} \\
\cmidrule(lr){5-7}
 & [s] & [\%] & [min$^{-1}$] & \textbf{Med.} & $\bm{p_{90}}$ & \textbf{Max} \\
\midrule
\texttt{HIT\_Graffiti\_Wall\_1} & 487.1 & 9.0  & 19.95 & 0.3 & 1.2 & 4.6 \\
\texttt{ShieldTunnel\_1}        & 542.1 & 70.8 & 10.63 & 0.3 & 5.2 & 162.0 \\
\texttt{ShieldTunnel\_5}        & 251.3 & 86.4 & 7.88  & 0.2 & 6.5 & 195.2 \\
\texttt{vehicle\_highway\_0}    & 694.6 & 5.2  & 13.65 & 0.2 & 0.7 & 6.8 \\
\texttt{vehicle\_tunnel\_0}     & 668.6 & 24.7 & 61.20 & 0.3 & 1.0 & 11.1 \\
\bottomrule[1pt]
\end{tabular}}
\vspace{-2mm}
\label{tab:degeneracy_switching}
\end{table}

\definecolor{underconf}{RGB}{0,114,178}
\definecolor{overconf}{RGB}{140,29,24}
\begin{figure*}[t!]
\centering
\includegraphics[width=\linewidth]{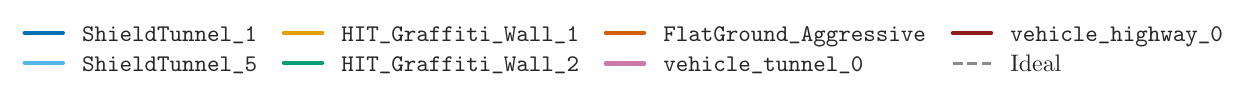} \\[1mm]
\subfigure[$\sigma_v = 1$]{\includegraphics[width=0.36\linewidth]{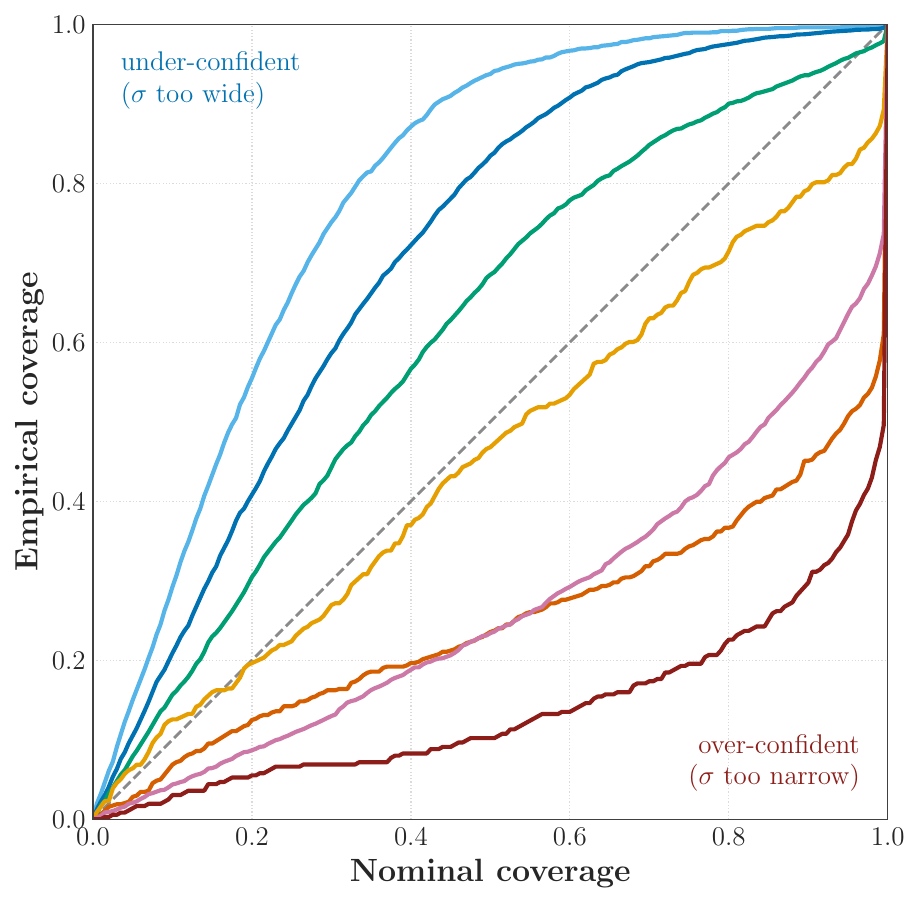}}
\hspace{6mm}
\subfigure[$\sigma_v = 0.01$]{\includegraphics[width=0.36\linewidth]{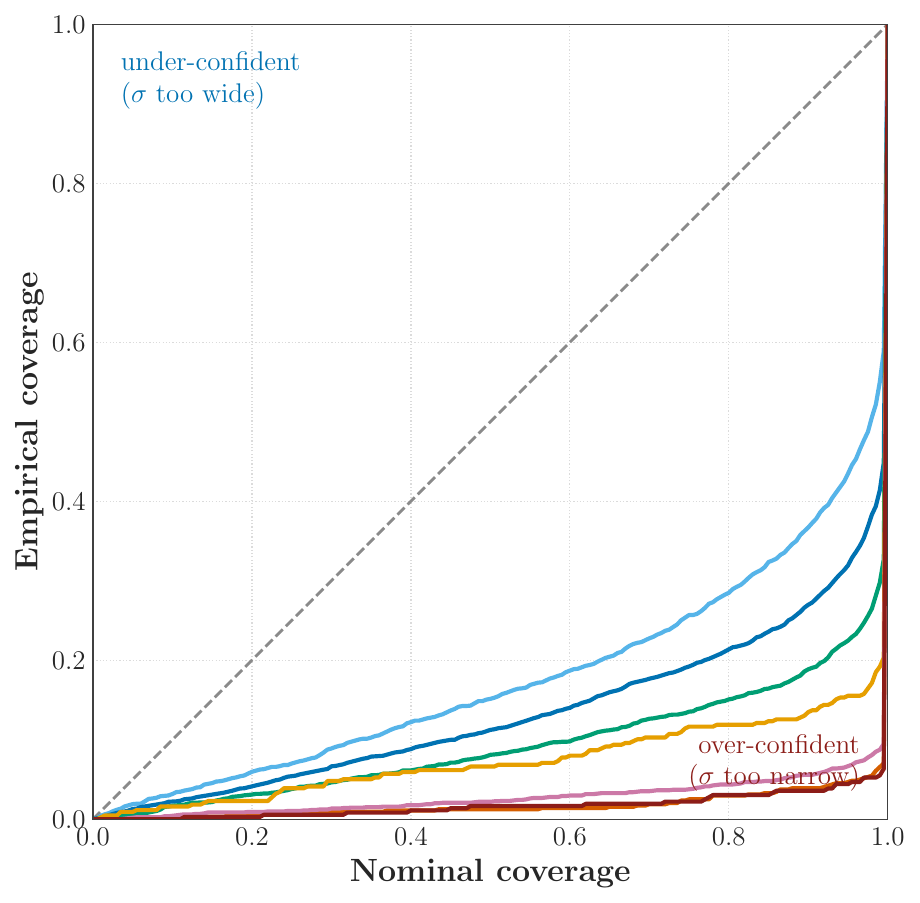}}
\caption{Confidence-interval coverage of the network-predicted velocity covariance along the degenerate directions, evaluated on representative sequences. The horizontal axis is the nominal coverage and the vertical axis is the empirical coverage, so the dashed diagonal denotes a correctly calibrated prediction. A curve above the diagonal indicates that the predicted covariance is too wide (\textcolor{underconf}{under-confident}), and a curve below the diagonal indicates that it is too narrow (\textcolor{overconf}{over-confident}). The value $\sigma_v = 1$ corresponds to the raw network output and $\sigma_v = 0.01$ to the value used in the update.}
\label{fig:covariance_calibration}
\vspace{-2mm}
\end{figure*}

Nevertheless, the influence of these short intervals is attenuated by how the decision is used in the state correction.
Rather than replacing the estimate along the degenerate directions, the decision gates a measurement update whose magnitude is determined by the Kalman gain in (\ref{eq30}).
A hard substitution scheme, such as that of ININ-LIO~\cite{inin_lio}, assigns unit weight to ${}^B\mathbf{v}^\text{NN}_M$ along the degenerate directions, so both the fluctuation of the neural velocity and every toggle of ${}^B\mathbf{V}_\mathcal{D}$ propagate to the output without attenuation.
In contrast, the Kalman gain $\mathbf{K}$ weights the correction by the relative uncertainty of the state and the network, and the recursive structure of the ESKF accumulates this correction over successive scans.
A single spurious decision therefore produces a bounded correction rather than a discontinuity in the estimate.
Fig.~\ref{fig:degeneracy_neural_gap} illustrates why this distinction matters.
Near $420$\,s on \texttt{ShieldTunnel\_1} and near $200$\,s on \texttt{ShieldTunnel\_5}, the decision toggles repeatedly while the neural speed and the estimated state speed differ visibly.
Under a hard substitution, the estimate would alternate between the two curves at every toggle and thus oscillate, whereas the proposed covariance-weighted update keeps the state continuous.
This interpretation is consistent with the results already reported in the main manuscript: as shown in Fig.~\ref{fig:lio_output_state_update} and Table~\ref{tab:ablation_state_update}, ININ-LIO, which replaces the velocity along the degenerate directions, exhibits substantial velocity oscillations, while our update does not.

\subsection{Calibration of the Predicted Velocity Covariance} \label{appendix:covariance_calibration}
Section~\ref{appendix:rotation_reliability} showed that $\boldsymbol{\Sigma}^\text{NN}_{v, M}$ and the velocity error grow together, which establishes a correlation but does not test whether the predicted magnitude is correct.
We therefore evaluate the confidence-interval coverage, which is sensitive to that magnitude.
Using the projected residual $\bm{e}_\mathcal{D}$ of (\ref{eq27}) and the projected measurement noise $\boldsymbol{\Sigma}_\mathcal{D}$ of (\ref{eq29}), we compute $\gamma^2 = \bm{e}_\mathcal{D}^\top \boldsymbol{\Sigma}_\mathcal{D}^{-1} \bm{e}_\mathcal{D}$, which follows $\chi^2(m_\mathrm{t})$ if the prediction is correct, where $m_\mathrm{t}$ is the number of degenerate translational directions of the scan, as defined in (\ref{eq22}).
Since $\gamma^2$ omits the term $\mathbf{J}_\mathcal{D} \widehat{\mathbf{P}} \mathbf{J}_\mathcal{D}^\top$ of (\ref{eq30}), it assesses the predicted covariance alone rather than the joint consistency of the network and the filter.
Because $m_\mathrm{t}$ varies between one and three, the coverage is assessed through the probability integral transform $u = F_{\chi^2}(\gamma^2; m_\mathrm{t})$, which is uniform under correct calibration, and we also report the mean $\gamma^2/m_\mathrm{t}$, which equals one when the prediction is correct.

With the raw network output ($\sigma_v = 1$), the sequences do not share a single behavior, as shown in Fig.~\ref{fig:covariance_calibration}(a).
\texttt{HIT\_Graffiti\_Wall\_2} lies close to the diagonal ($\gamma^2/m_\mathrm{t} = 1.05$), \texttt{ShieldTunnel\_1} and \texttt{ShieldTunnel\_5} lie above it (\textcolor{underconf}{under-confident region}; $0.40$ and $0.17$), and the remaining sequences lie below it (\textcolor{overconf}{over-confident region}; $5.19$ to $112.86$).
With the implemented $\sigma_v = 0.01$, all of them move into the \textcolor{overconf}{over-confident region} while their relative order is preserved, as shown in Fig.~\ref{fig:covariance_calibration}(b).

The deviations from the diagonal, however, do not translate into a prescription for $\boldsymbol{\Sigma}_\mathcal{D}$.
For \texttt{vehicle\_tunnel\_0}, which falls in the \textcolor{overconf}{over-confident region}, a wider covariance would be required, yet Table~\ref{tab:sigma_sensitivity} shows the ATE rising from $53.05$\,m at $\sigma_v = 0.01$ to $148.17$\,m at $\sigma_v = 1$; for \texttt{ShieldTunnel\_5}, which falls in the \textcolor{underconf}{under-confident region}, a narrower one would be required, yet the ATE rises from $8.73$\,m to $10.75$\,m at $\sigma_v = 0.001$.
The reason is that the coverage evaluates $\boldsymbol{\Sigma}_\mathcal{D}$ in isolation, whereas the weight actually applied to the neural velocity is set by the ratio between $\boldsymbol{\Sigma}_\mathcal{D}$ and $\mathbf{J}_\mathcal{D} \widehat{\mathbf{P}} \mathbf{J}_\mathcal{D}^\top$ in (\ref{eq30}).

In addition, the well-conditioned parts of Fig.~\ref{fig:degeneracy_neural_gap} show a gap between the velocity estimated by the LIO system and the one predicted by the network.
Since the network infers the velocity only from the IMU measurements and the LIO-estimated rotation in (\ref{eq4})--(\ref{eq6}), it has access neither to the velocity estimated by the system nor to its uncertainty, and therefore cannot recognize that its own output deviates from the better-conditioned estimate.
Its predicted covariance therefore describes the network output alone, and it is rescaled by $\sigma_v$ in (\ref{eq29}) before it enters the ESKF.
Because that scaling is heuristic, the neural velocity may still be weighted over-confidently.
As shown in Section~\ref{appendix:sigma_ablation}, the scaling is nonetheless necessary, and determining it in a more principled way remains future work.

\subsection{Detail of Private Datasets} \label{appendix:private_dataset_detail}
For the train dataset, \texttt{handheld\_train} was collected using a Livox Mid-360, where the IMU was oriented in various directions such as sideways, backward, and with tilting motions.

For degeneracy datasets, both \texttt{handheld\_light} and \texttt{handheld\_dark} traversed the tunnel while facing the walls with Livox Avia.
\texttt{handheld\_light} traversed the tunnel in a bright environment with abundant visual features, whereas \texttt{handheld\_dark} was recorded in a darker setting by reducing the camera aperture.
Additionally, \texttt{handheld\_light} retraced its path to return to the starting position, while \texttt{handheld\_dark} crossed to the opposite side of the path and returned to the initial location. For the end-to-end error evaluation, the start and end poses of each sequence were positioned at the same physical location, which was marked with tape.

The reason for using the Livox Mid-360 for training and the Livox Avia in degenerate scenarios is that the Mid-360 provides a $360^\circ$ LiDAR scan, allowing stable odometry estimation via LIO even in environments where the platform faces a wall, whereas the Avia, with its narrow FOV, was used because degeneracy occurs when facing a wall.
Furthermore, for safety reasons, experiments were not conducted in long tunnels where LiDAR degeneracy could occur.

\begin{table}[t]
\caption{System Parameter Settings.}
\centering
\small
\begin{tabular}{>{\centering\arraybackslash}p{2.85cm}|>{\centering\arraybackslash}p{2.3cm}|>{\centering\arraybackslash}p{2.3cm}}\toprule[1pt]
    \textbf{Model} & \textbf{Notation} & \textbf{Value} \\
    \midrule
     \multirow{3}{*}{Neural Network} 
     & $\lambda_C$           & 0.002 \\
     & $\delta_\text{car}$   & 0.5 \\
     & $\delta_\text{Livox}$ & 0.05 \\
     \midrule
     \multirow{5}{*}{Degeneracy} 
     & $\theta_{w}$   & $45^\circ$\\
     & $\theta_{s}$   & $80^\circ$ \\
     & $\tau_{w}$     & 180 \\
     & $\tau_{s}$     & 35 \\
     & $\sigma_{v}$   & 0.01 \\
    \bottomrule[1pt]
\end{tabular}
\label{tab:parameter}
\vspace{-2mm}
\end{table}

\end{document}